\documentclass{article}

\PassOptionsToPackage{numbers,sort&compress}{natbib}
\usepackage[final]{neurips_2021}

\usepackage[utf8]{inputenc} 
\usepackage[T1]{fontenc}    
\usepackage{hyperref}       
\usepackage{url}            
\usepackage{booktabs}       
\usepackage{amsfonts}       
\usepackage{nicefrac}       
\usepackage{microtype}      
\usepackage{xcolor}         
\usepackage{graphicx}
\usepackage{graphbox}
\usepackage{subcaption}

\usepackage{makecell}

\usepackage{graphics}
\usepackage{mathtools}

\usepackage{algorithm}
\usepackage{algorithmic}
\usepackage{enumitem}

\usepackage{amsfonts}       
\usepackage{amsmath}       
\usepackage{amssymb}
\usepackage{amsthm}
\usepackage{bm}

\newcommand{\Fig}[1]{Figure~\ref{#1}}  
\newcommand{\fig}[1]{Fig.~\ref{#1}}    
\newcommand{\tab}[1]{Table~\ref{#1}}

\newcommand{\eqn}[1]{Eq.~\ref{#1}} 
\newcommand{\eqnp}[1]{(Eq.~\ref{#1})} 
\renewcommand{\sec}[1]{Sec.~\ref{#1}} 
\newcommand{\supp}[1]{Suppl.~\ref{#1}}

\usepackage{xspace}
\makeatletter
\DeclareRobustCommand\onedot{\futurelet\@let@token\@onedot}
\def\@onedot{\ifx\@let@token.\else.\null\fi\xspace}
\def\eg{e.g\onedot}
\def\ie{i.e\onedot}

\makeatother

\newcommand{\Real}{\ensuremath{\mathbb R}}        

\let\originalleft\left
\let\originalright\right
\renewcommand{\left}{\mathopen{}\mathclose\bgroup\originalleft}
\renewcommand{\right}{\aftergroup\egroup\originalright}

\newcommand{\g}[1]{
  \ifthenelse{\equal{#1}{(}}
  {\left( }%
    { \ifthenelse{\equal{#1}{)}}
      { \right)}%
    { \ifthenelse{\equal{#1}{[}}
      {\left[}%
        { \ifthenelse{\equal{#1}{]}}
          { \right]}%
        {#1}}
    }
  }
}

\newcounter{mysubfig}%
\newcounter{myfig}%
\newcommand{\startsubfig}{\setcounter{mysubfig}{0}\setcounter{myfig}{\value{figure}}\addtocounter{myfig}{1}\small }%
\newcommand{\subfig}[1]{\refstepcounter{mysubfig}\label{#1}(\alph{mysubfig})\xspace}%

\usepackage{xcolor}

\definecolor{chrisblue}{RGB}{99,106,134}
\definecolor{chrisgreen}{RGB}{145,170,126}
\definecolor{chrisred}{RGB}{155,54,62}

\definecolor{ourorange}{rgb}{0.881,0.611,0.142}
\definecolor{ourviolet}{rgb}{0.528,0.471,0.701}
\definecolor{ourbrown}{rgb}{0.772,0.432,0.102}
\definecolor{ourlightblue}{rgb}{0.364,0.619,0.782}
\definecolor{ourdarkgreen}{rgb}{0.572,0.586,0.}

\usepackage{wrapfig}
\usepackage{hyperref}
\hypersetup{
       colorlinks,
       linkcolor=chrisblue,
       urlcolor=chrisblue,
       citecolor=chrisred}

\newcommand{\netth}{\mathit{f}_\theta}

\newcommand{\method}{GateL0RD\xspace}

\renewcommand{\paragraph}[1]{\par \textbf{#1}}

\title{Sparsely Changing Latent States for Prediction and Planning in Partially Observable Domains}

\author{%
  Christian Gumbsch\\ 
  Autonomous Learning Group\\
  Max Planck Institute for Intelligent Systems\\
  \& Neuro-Cognitive Modeling Group\\
  University of Tübingen\\
  Tübingen, Germany\\
  \texttt{christian.gumbsch@tuebingen.mpg.de} \\
  \And
  Martin V.\ Butz \\
  Neuro-Cognitive Modeling Group\\
  University of Tübingen\\
  Tübingen, Germany\\
  \texttt{martin.butz@uni-tuebingen.de} \\
  \AND
  Georg Martius\\
  Autonomous Learning Group\\
  Max Planck Institute for Intelligent Systems\\
  Tübingen, Germany\\
  \texttt{georg.martius@tuebingen.mpg.de} \\
}

\begin{document}

\maketitle

\begin{abstract}
A common approach to prediction and planning in partially observable domains is to use recurrent neural networks (RNNs), which ideally develop and maintain a latent memory about hidden, task-relevant factors.
We hypothesize that many of these hidden factors in the physical world are constant over time, changing only sparsely.
To study this hypothesis, we propose Gated $L_0$ Regularized Dynamics (\method), a novel recurrent architecture that incorporates the inductive bias to maintain stable, sparsely changing latent states.
The bias is implemented by means of a novel internal gating function and a penalty on the $L_0$ norm of latent state changes.
We demonstrate that \method can compete with or outperform state-of-the-art RNNs in a variety of partially observable prediction and control tasks. 
\method tends to encode the underlying generative factors of the environment, ignores spurious temporal dependencies, and generalizes better, improving sampling efficiency and overall performance in model-based planning and reinforcement learning tasks. 
Moreover, we show that the developing latent states can be easily interpreted, which is a step towards better explainability in RNNs. 
\end{abstract}

\section{Introduction}
When does the meeting start? Where are my car keys? Is the stove turned off?
Humans memorize lots of information over extended periods of time.
In contrast, classical planning methods assume that the state of the environment is fully observable at every time step \cite{Sutton:2018}.
This assumption does not hold for realistic applications, where generative processes are only indirectly observable or entities are occluded. 
Planning in such Partially Observable Markov Decision Processes (POMDP) is a challenging problem, because suitably-structured memory is required for decision making.

Recurrent neural networks (RNNs) are often used to deal with partial observability \citep{hausknecht2017deep, igl2018deep, zhu2018improving}.
They encode past observations by maintaining latent states, which are iteratively updated.
However, continuously updating the latent state causes past information to quickly ``wash out''.
Long-Short Term Memory networks (LSTM, \citep{LSTM1}) and Gated Recurrent Units (GRU, \citep{GRU1}) deal with this problem by using internal gates.
However, they cannot leave their latent states completely unchanged, because small amounts of information continuously leak through the sigmoidal gating functions.
Additionally, inputs typically need to pass through the latent state to affect the output, making it hard to disentangle observable from unobservable information within their latent states.

Our hypothesis is that many generative latent factors in the physical world are constant over extended periods of time.
Thus, there might not be the need to update memory at every time step.
For example, consider dropping an object:
If the drop-off point as well as some latent generative factors, such as gravity and aerodynamic object properties, are known, iteratively predicting the fall can be reasonably accomplished by a non-recurrent process.
Similarly, when an agent picks up a key, it is sufficient to memorize that the key is inside their pocket.
However, latent factors typically do change significantly and systematically at particular points in time.
For example, the aerodynamic properties of an object change drastically when the falling object shatters on the floor, and the location of the key changes systematically when the agent removes it from their pocket.

These observations are related to assumptions used in causality research.
A common assumption is that the generative process of a system is composed of autonomous mechanisms that describe causal relationships between the system's variables \cite{schoelkopfBook, schoelkopfCausality, schoelkopfBengio}.
When considering Markov Decision Processes, it has been proposed that these mechanisms tend to interact sparsely in time and locally in space \cite{Pitis2020, Seitzer2021CID}.
Causal models aim at creating dependencies between variables only when there exists a causal relationship between them, in order to improve generalization  \cite{schoelkopfCausality}.
Updating the latent state of a model in every time step, on the other hand, induces the prior assumption that the generative latent state typically depends on all previous inputs.
Thus, by suitably segmenting the dependencies of the latent variables over time, one can expect improved generalization across spurious temporal dependencies.

Very similar propositions have been made for human cognition.
Humans tend to perceive their stream of sensory information in terms of events \cite{BaldwinKosie:2021,Butz:2021,Kuperberg:2021,radvansky2014event,Zacks:2007}.
Event Segmentation Theory (EST) \cite{Zacks:2007} 
postulates a set of active event models, which encode event-respective aspects over extended periods of time and switch individually at event transitions.
To learn about the transitions and consolidate associated latent event encodings, measurements of surprise and other significant changes in predictive model activities, as well as latent state stability assumptions, have been proposed as suitable inductive event segmentation biases \cite{Butz:2016,Butz:2019,gumbsch2019autonomous, Humaidan:2021,Schapiro:2013,Shin:2021,Zacks:2007}. Explicit relations to causality have been put forward in \cite{Butz:2021a}.

In accordance to EST and our sparsely changing latent factor assumption, we introduce Gated $L_0$ Regularized Dynamics (\method). 
\method applies $L_0$-regularized gates, inducing an inductive learning bias to encode piecewise constant latent state dynamics. 
\method thus becomes able to memorize task-relevant information over long periods of time.
The main contributions of this work can be summarized as follows. 
(i)~We introduce a stochastic, rectified gating function for controlling latent state updates, which we regularize towards sparse updates using the $L_0$ norm. 
(ii)~We demonstrate that our network performs as good or better than state-of-the-art RNNs for prediction or control in various partially-observable problems with piecewise constant dynamics.
(iii)~We also show that the inductive bias leads to better generalization under distributional shifts.
(iv)~Lastly, we show that the latent states can be easily interpreted by humans.

\section{Background}

Let $\netth: \mathcal{X} \times \mathcal{H} \rightarrow \mathcal{Y} \times \mathcal{H}$ be a recurrent neural network (RNN) with learnable parameters $\theta$ mapping inputs\footnote{Notation: bold lowercase letters denote vectors (\eg, $\bm{x}$). Vector dimensions are denoted by superscript (\eg $\bm x = [x^1, x^2, \ldots, x^n] \in \Real^n$).
Time or other additional information is denoted by subscript (\eg, $\bm{x}_t$).}
 $\bm{x}_t \in \mathcal{X}$ and $\bm{h}_{t-1} \in \mathcal{H}$ the latent (hidden) state to the output $\bm{\hat{y}}_t \in \mathcal{Y}$ and updated latent states $\bm h_{t}$.
The training dataset $\mathcal{D}$ consists of sequences of input-output pairs $d = [(\bm{x}_{1}, \bm{y}_{1}), \ldots, (\bm{x}_{T}, \bm{y}_{T})]$ of length $T$.
In this paper, we consider the prediction and control of systems that can be described by a partially observable Markov decision process (POMDP) with state space $\mathcal{S}$, action space $\mathcal{A}$, observations space  $\mathcal{O}$, and deterministic hidden transitions $\mathcal S\times \mathcal A \to \mathcal S$.\footnote{We treat the prediction of time series without any actions as a special case of the POMDP with $\mathcal{A} = \emptyset$.}

\section{$L_0$-regularization of latent state changes}
\label{sec:l0}

We want the RNN $\netth$ to learn to solve a task, while maintaining piecewise constant latent states over time.
The network creates a dynamics of latent states $\bm h_t$ when applied to a sequence: $(\bm{\hat{y}}_{t},\bm h_{t}) = \netth(\bm x_{t}, \bm h_{t-1})$ starting from some $\bm h_0$.
The most suitable measure to determine how much a time-series is piecewise constant is the $L_0$ norm applied to temporal changes.
With the change in latent state as $\bm{\Delta h}_{t} = \bm{h}_{t-1} - \bm{h}_{t}$, we define the $L_0$-loss as
\begin{align}
    \mathcal{L}_{L_0}(\bm{\Delta h}) = \left\Vert \bm{\Delta h} \right\Vert_0 = \sum_{j=1} \mathbb{I}(\Delta h^j \neq 0), \label{eq:L0}
\end{align}
which penalizes the {\bf number of non-zero entries} of the vector of latent state changes $\bm{\Delta h}$.

The regularization loss from \eqn{eq:L0} can be combine in the usual way with the task objective to yield the overall learning objective $\mathcal{L}$ of the network:

\begin{equation}
  \mathcal{L}(\mathcal{D}, \bm{\theta}) = \mathbb{E}_{d\sim \mathcal D} \Big[ \sum_t \mathcal{L}_{\mathrm{task}} (\bm{\hat{y}}_{t}, \bm{y}_{t})  + \lambda \mathcal{L}_{L_0}(\bm{\Delta h}_{t}) \Big]
\label{eq:loss}
\end{equation}
with $(\bm{\hat{y}}_{t},\bm h_{t}) = \netth(\bm x_{t}, \bm h_{t-1})$.
The task-dependent loss $\mathcal{L}_{\mathrm{task}}(\cdot, \cdot)$ can be, for instance, the mean-squared error for regression or cross-entropy loss for classification.
The hyperparameter $\lambda$ controls the trade-off between the task-based loss and the desired latent state regularization.

Unfortunately, we cannot directly minimize this loss using gradient-based techniques, such as stochastic gradient descent (SGD), due to the non-differentiability of the $L_0$-term.
\citet{louizos2018L0} proposed a way to learn $L_0$ regularization of the learnable parameters of a neural network with SGD.
They achieve this by using a set of stochastic gates controlling the parameters' usage.
Each learnable parameter $\theta^j$ that is subject to the $L_0$ loss is substituted by a gated version $\theta'^j = \Theta(s^j)\theta^{j}$ where $\Theta(\cdot)$ is the Heaviside step function ($\Theta(s) = 0$ if $s\le 0$ and $1$ otherwise) and $\bm{s}$ is determined by a distribution $q(\bm{s}|\bm{\nu})$ with learned parameters $\bm{\nu}$. Thus, $\theta'^{j}$ is only non-zero if $s^j>0$.
This allows to rewrite the $L_0$ loss \eqnp{eq:L0} for $\theta'$ as:

\begin{align}
    \mathcal{L}_{L_0}(\bm{\theta'},\bm \nu) &= \left\Vert \bm{\theta'} \right\Vert_0 = \sum_{j}\Theta(s^j) &\text{with } \bm{s} \sim q(\bm{s}; \bm{\nu}), \label{eq:l0loss}
\end{align}
where parameters $\bm \nu$ influence sparsity and are affected by the loss.

To tackle the problem of non-differentiable binary gates, we can use a smooth approximation as a surrogate \citep{maddison2017concrete, louizos2018L0, GumbelSoftmax}.
Alternatively, we can substitute its gradients during the backward pass, for example using the straight-through estimator \citep{StraightThroughBengio}, which treats the step function as a linear function during the backward pass, or approximate its gradients as in the REINFORCE algorithm \citep{REINFORCE}.

To transfer this approach to regularize the latent state dynamics in an RNN, we require an internal gating function $\Lambda(\cdot) \in [0, 1]$, which controls whether the latent state is updated or not. For instance:
\begin{align}
    \bm h_{t} &= \bm h_{t-1} + \Lambda(\bm s)\bm{\Delta \tilde h}_{t-1} & \text{with } \bm{\Delta \tilde h}_{t-1} = \bm{\tilde h}_t - \bm h_{t-1}  \label{eq:latent-update}
\end{align}

where $\bm{\tilde h}$ is the proposed new latent state and $s$
is a stochastic variable depending on the current input and previous latent state and the parameters, \ie $\bm{s}_t \sim q(\bm{s}_t; \bm{x}_t, \bm{h}_{t-1}, \bm{\nu})$. For brevity, we merge the parameters $\bm \nu$ into the overall parameter set, \ie $\bm \nu \subset \bm \theta$.
For computing \eqn{eq:loss} we need to binarize the gate by applying the step function $\Theta(\Lambda(s))$. Thus we can rewrite \eqn{eq:loss} as
\begin{align}
  \mathcal{L}(\mathcal{D}, \bm{\theta}) &=  \mathbb{E}_{d\sim \mathcal D} \Big[ \sum_t \mathcal{L}_\mathrm{task} (\bm{\hat{y}}_{t}, \bm{y}_{t})  + \lambda \sum_{t} \Theta(\Lambda(\bm{s}_t)) \Big]. \label{eq:loss2}
\end{align}

LSTMs and GRUs use deterministic sigmoidal gates for $\Lambda$ in \eqn{eq:latent-update} to determine how to update their latent state.
However, it is not straight forward to apply this approach to them (detailed in \supp{sec:SupplRNNs}).
Thus, we instead introduce a novel RNN, that merges components from GRUs and LSTMs, to implement the proposed $L_0$ regularization of latent state changes while still allowing the network to make powerful computations.
We name our network Gated $L_0$ Regularized Dynamics (\method).

\section{\method}

\begin{figure*}
\begin{subfigure}{0.45\linewidth}
\centering
\includegraphics[width=\linewidth]{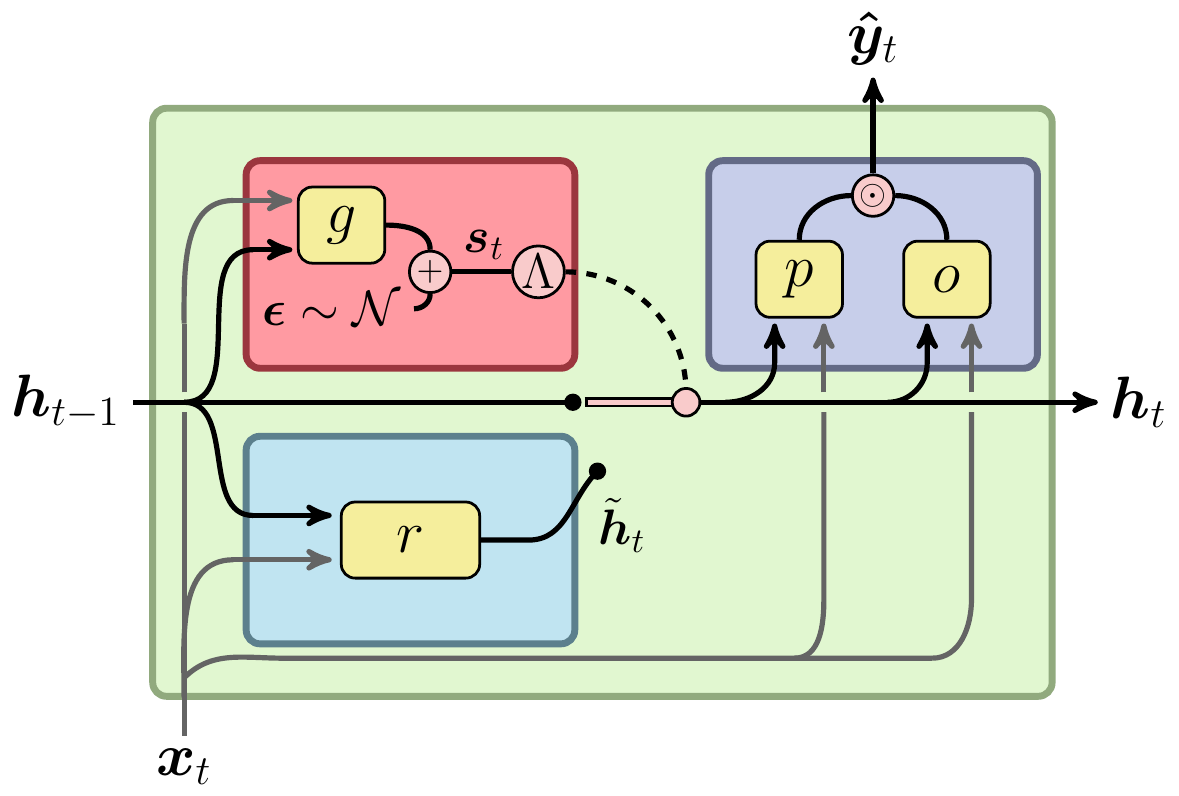}
\caption{Illustration of the core of \method.}\label{fig:architectureCore}
\end{subfigure}
\hfill
\begin{subfigure}{0.25\linewidth}
\begin{subfigure}{\linewidth}
\centering
\includegraphics[width=0.8\linewidth]{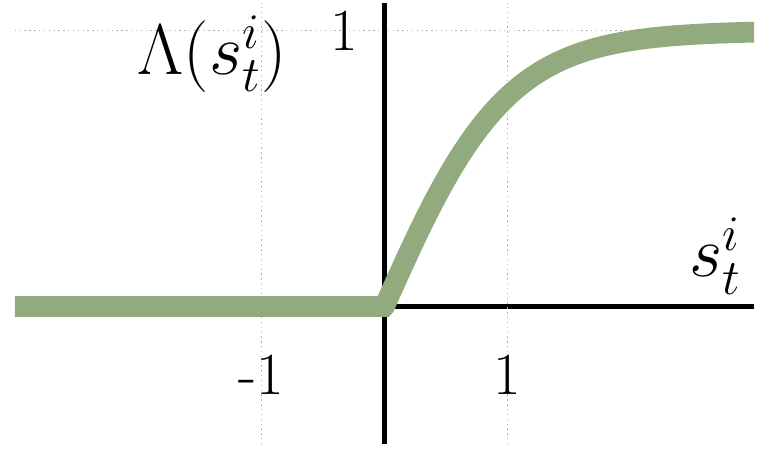}
\vspace*{-0.1cm}
\caption{Gate-activation $\Lambda(s)$}\label{fig:ReTanh}
\vspace*{0.3cm}
\end{subfigure}
\begin{subfigure}{\linewidth}
\centering
\includegraphics[width=0.8\linewidth]{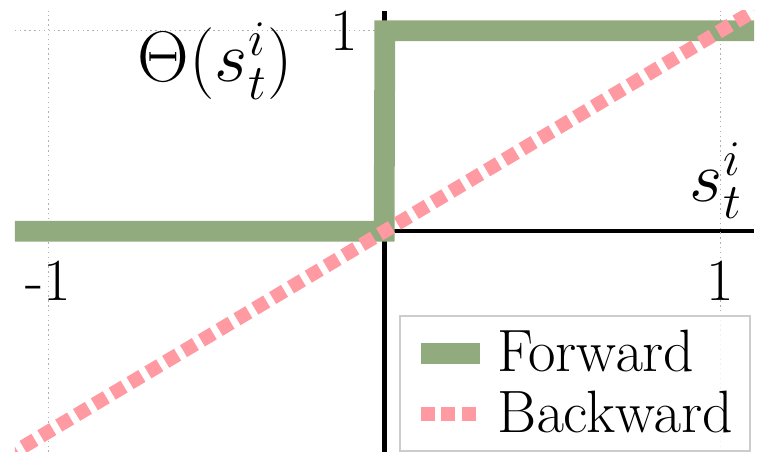}
\vspace*{-0.1cm}
\caption{$\Theta$ with its substitution}\label{fig:Theta}
\end{subfigure}
\end{subfigure}
\hfill
\begin{subfigure}{0.24\linewidth}
\centering
\includegraphics[width=0.87\linewidth]{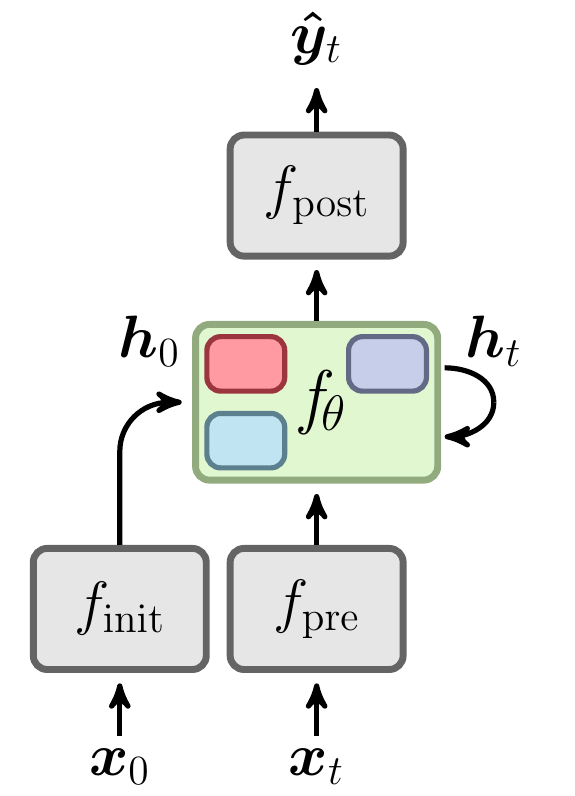}
\caption{Overall \method}\label{fig:overall}
\end{subfigure}\vspace{-.5em}
    \caption{Architecture overview. (a)~\method with its three subnetworks. The \emph{gating function} controls the latent state update (red), the \emph{recommendation function} computes a new latent state (blue) and the \emph{output function} computes the output (purple).
    (b)~Gate-activation function $\Lambda$ (ReTanh).
    (c)~Heaviside step function $\Theta$ and its gradient estimator.
    (d)~Overall architecture. } \label{fig:architecture}
\end{figure*}

The core of \method implements the general mapping $(\bm{\hat{y}}_{t},\bm h_{t}) = \netth(\bm x_{t}, \bm h_{t-1})$ using three functions, or subnetworks:
(1) a \emph{recommendation network} $r$, which proposes a new candidate latent state, (2) a \emph{gating network} $g$, which determines how the latent state is updated, and (3) an \emph{output function}, which computes the output based on the updated latent state and the input.
The network is systematically illustrated in \fig{fig:architectureCore}.

The overall processing is described by the following equations:
\begin{align}
    \bm{s_t} &\sim \mathcal{N}(g(\bm x_t, \bm h_{t-1}), \bm \Sigma)&\text{(sample gate input)}\label{eq:s}\\
    \Lambda(\bm{s}) &:= \max \left(0, \tanh(\bm{s})\right)&\text{(new gating function)}\label{eq:Lambda}\\
    \bm{h}_t &= \bm h_{t-1} + \Lambda(\bm{s_t}) \odot \g(r\g(\bm{x}_t,\bm{h}_{t-1}\g) - \bm h_{t-1}\g) & \text{(update or keep  latent state)} \label{eq:new_hidden}\\
    \bm{\hat y}_t &= p(\bm{x}_t, \bm{h}_{t}) \odot o(\bm{x}_t, \bm{h}_{t}),&\text{(compute output)}\label{eq:output}
\end{align}
where $\odot$ denotes element-wise multiplication (Hadamard product).

We start with the control of the latent state in \eqn{eq:new_hidden}.
Following \eqn{eq:latent-update}, a new latent value is proposed by the \emph{recommendation function} $r(\bm{x}_t, \bm{h}_{t-1})$ and the update is ``gated'' by $\Lambda(\bm s)$. Importantly, if $\Lambda(\bm s)= \bm{0}$ no change to the latent state occurs.
Note that the update in \eqn{eq:new_hidden} is in principle equivalent to the latent state update in GRUs \citep{GRU1}, for which it is typically written as $\bm{h}_t = \Lambda(\bm s) \odot r\g(\bm{x}_t,\bm{h}_{t-1}\g) + (1 - \Lambda(\bm s)) \odot \bm{h}_{t-1}$ with $\Lambda(\bm s)$ a deterministic sigmoidal gate.

Because we aim for piecewise constant latent states, the gating function $\Lambda$ defined in \eqn{eq:Lambda} needs to be able to output exactly zero.
A potential choice would be the Heaviside function, \ie either copy the new latent state or keep the old one.
This, however, does not allow any multiplicative computation.
So a natural choice is to combine the standard sigmoid gate of RNNs with the step-function: $\Lambda(\bm{s}) = \max \g(0, \tanh(\bm{s})\g)$ which we call ReTanh (rectified tanh)\footnote{Note that $\tanh(s) = 2\cdot \text{sigmoid}(2s) - 1$.}.
\Fig{fig:ReTanh} shows the activation function $\Lambda$ depending on its input.
The gate is closed ($\Lambda(s^i) = 0$) for all inputs $s^i \leq 0$.
A closed gate results in a latent state that remains constant in dimension $i$, \ie, $h^i_t = h^i_{t-1}$.
On the other hand, for $s^i > 0$ the latent state is interpolated between the proposed new value and the old one.

The next puzzle piece is the input to the gate. Motivated from the $L_0$ regularization in \eqn{eq:L0} we use a stochastic input. However, in our RNN setting, it should depend on the current situation. Thus, we use a Gaussian distribution for $q$ with the mean determined by the \emph{gating network} $g(\bm x_t,\bm h_{t-1})$ as defined in \eqn{eq:s}.
We chose a fixed diagonal covariance matrix $\bm{\Sigma}$, which we set to $\Sigma^{i, i} = 0.1$.
To train our network using backpropagation, we implement the sampling using the \emph{reparametrization trick} \citep{VAE}.
We introduce a noise variable $\epsilon$ and compute the gate activation as
\begin{align}
    \bm{s}_t &= g(\bm x_t,\bm h_{t-1}) + \bm{\epsilon} & \text{with ~} \bm{\epsilon} \sim \mathcal{N}(\bm{0}, \bm{\Sigma}) \label{EqReparametrization}.
\end{align}
During testing we set $\bm{\epsilon} = \bm{0}$ 
to achieve maximally accurate predictions.

Finally the output $\bm{\hat y}$ is computed from the inputs and the new latent state $\bm{h}_t$ in \eqn{eq:output}.
Inspired by LSTMs \citep{LSTM1}, the output is determined by a multiplication of a normal branch ($p(\bm{x}_t, \bm{h}_{t})$) and a sigmoidal gating branch ($o(\bm{x}_t, \bm{h}_{t})$).
 We thus enable both additive as well as multiplicative effects of $\bm{x}_t$ and $\bm{h}_t$ on the output, enhancing the expressive power of the piecewise constant latent states.

In our implementation, all subnetworks are MLPs.
$r, p$ use a $\tanh$ output activation; $o$ uses a sigmoid; $g$ has a linear output.
$p, o$ are one-layer networks. By default, $r, g$ are also one-layer networks. 
However, when comparing against deep (stacked) RNNs, we increase the number of layers of $r$ and $g$ to up to three (cf.\ \supp{sec:suppExpDetails}).

We use the loss defined in \eqn{eq:loss2}. \method is fully differentiable except for the Heaviside step function $\Theta$ in \eqn{eq:loss2}.
A simple approach to deal with discrete variables is to approximate the gradients by a differentiable estimator \citep{StraightThroughBengio, GumbelSoftmax, maddison2017concrete}.
We employ the straight-through estimator \citep{StraightThroughBengio}, which substitutes the gradients of the step function $\Theta$ by the derivative of the linear function (see \fig{fig:Theta}).

We use \method as a memory module of a more general architecture illustrated in \fig{fig:overall}.
The network input is preprocessed by a feed-forward network $f_\mathrm{pre}(\bm{x}_t)$. 
Similarly, its output is postprocessed by an MLP $f_\mathrm{post}(\bm{\hat y}_t)$ (\ie a readout layer) before computing the loss.
The latent state $\bm{h}_0$ of \method could be initialized by $\bm{0}$.
However, improvements can be achieved if the latent state is instead initialized by a context network $f_\mathrm{init}$, a shallow MLP that sets $\bm{h}_0$ based on the first input \citep{hiddenInit, ba2014multiple}.

In the Supplementary Material we ablate various components of \method, such as the gate activation function $\Lambda$ (\supp{sec:SupplGateFunction}), the gate stochasticity (\supp{sec:SupplGateStochasticity}), the context network $f_\mathrm{init}$ (\supp{sec:SupplFInit}), the multiplicative output branch $o$ (\supp{sec:SupplOutputGate}), and compare against $L_1$/$L_2$-variants (\supp{sec:SuppL1L2}).

\section{Related Work}

\paragraph{Structural regularization of latent updates:}
Pioneering work on regularizing latent updates was done by \citet{schmidhuber1992learning} who proposed the Neural History Compressor, a hierarchy of RNNs that autoregressively predict their next inputs. 
Thereby, the higher level RNN only becomes active and  updates its latent states, if the lower level RNN fails to predict the next input. 
To structure latent state updates, the Clockwork RNN \citep{clockworkRNN} partitions the hidden neurons of an RNN into separate modules, where each module operates at its own predefined frequency.
Along similar lines, Phased LSTMs \citep{PhasedLSTM} use gates that open periodically.
The update frequency in Clockwork RNNs and Phased LSTMs does not depend on the world state, but only on a predefined time scale.

\paragraph{Loss-based regularization of latent updates:}
For latent state regularization, \citet{RegHiddenStates} have proposed using an auxiliary loss term that punishes the change in $L_2$-norms of the latent state, which results in piecewise constant norms but not dynamics of the hidden states.

\paragraph{Binarized update gates:} Closely related to our ReTanh, Skip RNNs \citep{SkipRNN} use a binary gate to determine latent state update decisions.
Similarly, Gumbel-Gate LSTMs \citep{BinaryLSTM} replace sigmoid input and forget gates with stochastic, binary gates, approximated by a Gumbel-Softmax estimator \citep{GumbelSoftmax}.
Selective-Activation RNNs (SA-RNNs) \cite{hartvigsen2020learning} modify a GRU by masking the latent state with deterministic, binary gate and also incentivize sparsity. 
However, for GRUs the network output corresponds to the networks' latent state, thus, a piecewise constant latent state will result in piecewise constant outputs.
All of these models were designed for classification or language processing tasks -- none were applied for prediction or control in a POMDP setup, which we consider here.

\paragraph{Attention-based latent state updates:} Sparse latent state updates can also be achieved using attention \cite{graves2014neural, bahdanau2015neural, VaswaniEtAl2017:transformers}.
Neural Turing Machines  \cite{graves2014neural} use an attention mechanism to update an external memory block.
Thereby, the attention mechanism can focus and only modify a particular locations within the memory. 
Recurrent Independent Mechanisms (RIMs) \cite{RIMs} use a set of recurrent cells that only sparsely interact with the environment and one another through competition and a bottleneck of attention.
Recent extensions explore the update of the cells and the attention parameters at different time scales \cite{MetaRIMs}.
For RIMs the sparsity of the latent state changes is predefined via a hyperparameter that sets the number of active cells. In contrast, our $L_0$ loss implements a soft constraint.

\paragraph{Transformers:} 
Transformers \cite{VaswaniEtAl2017:transformers} omit memory altogether, processing a complete sequence for every output at once
 using key-based attention.
While this avoids problems arising from maintaining a latent state, their self-attention mechanism comes with high computational costs.
Transformers have shown breakthrough success in natural language processing.
However, it remains challenging to train them for planning or reinforcement learning applications in partially-observable domains \cite{parisotto2020stabilizing}.

\section{Experiments} 
\label{sec:Experiments}
Our experiments offer answers to the following questions: 
(a) Does \method generalize better to out-of-distribution inputs in partially observable domains than other commonly used RNNs?
(b) Is \method suitable for control problems that require (long-term) memorization of information?
(c) Are the developing latent states in \method easily interpretable by humans? 
Accordingly, we demonstrate both \method's ability to generalize from a 1-step prediction regime to autoregressive $N$-step prediction (\sec{sec:resBilliard}) and its prediction robustness when facing action rollouts from different policies (\sec{sec:resRRC}).
We then reveal precise memorization abilities (\sec{sec:resShepherd}) and show that \method is more sample efficient in various decision-making problems requiring memory (\sec{sec:resMiniGrid}). 
Finally, we examine exemplary latent state codes demonstrating their explainability (\sec{sec:resExplainability}).

In our experiments we compare \method to LSTMs \citep{LSTM1}, GRUs \citep{GRU1}, and Elman RNNs \citep{ElmanRNN}.
We use the architecture shown in \fig{fig:overall} for all networks, only replacing the core $\netth$.
We examine the RNNs both as a model for \emph{model-predictive control} (MPC)  as well as a memory module in a \emph{reinforcement learning} (RL) setup.
When used for prediction, the networks received the input $\bm{x}_t =(\bm{o}_t, \bm{a}_t)$ with observations $\bm{o}_t \in \mathcal{O}$ and actions $\bm{a}_t \in \mathcal{A}$  at time $t$ and were trained to predict the change in observation, \ie $\bm{y}_t = \Delta \bm{o}_{t+1}$ (detailed in \supp{sec:SuppTrainingDetails}).
During testing the next observational inputs were generated autoregressively as $\hat{\bm{o}}_{t+1} = \bm{o}_t + \hat{\bm{y}}_{t}$.
In the RL setting, the networks received as an input $\bm{x}_t = \bm{o}_t$ the observation $\bm{o}_t \in \mathcal{O}$ and were trained as an actor-critic architecture to produce both policy and value estimations (detailed in \supp{sec:SupplRL}).
The networks were trained using Adam \citep{kingma2014adam}, with learning rates and layer numbers determined via grid search for each network type individually (cf.\ \supp{sec:suppExpDetails}).

\begin{figure*}
\hfill
\begin{subfigure}{0.31\linewidth}
\centering
\includegraphics[width=\linewidth]{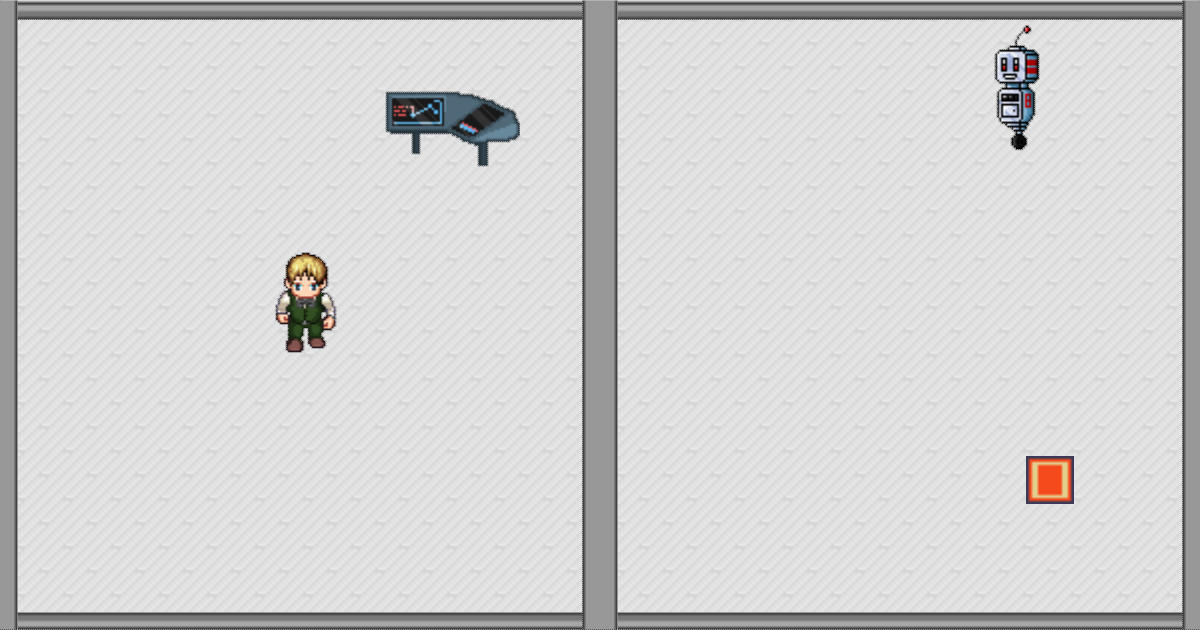}
\caption{Robot Remote Control\label{fig:RRC}}
\end{subfigure}
\hfill
\begin{subfigure}{0.31\linewidth}
\centering
\includegraphics[width=\linewidth]{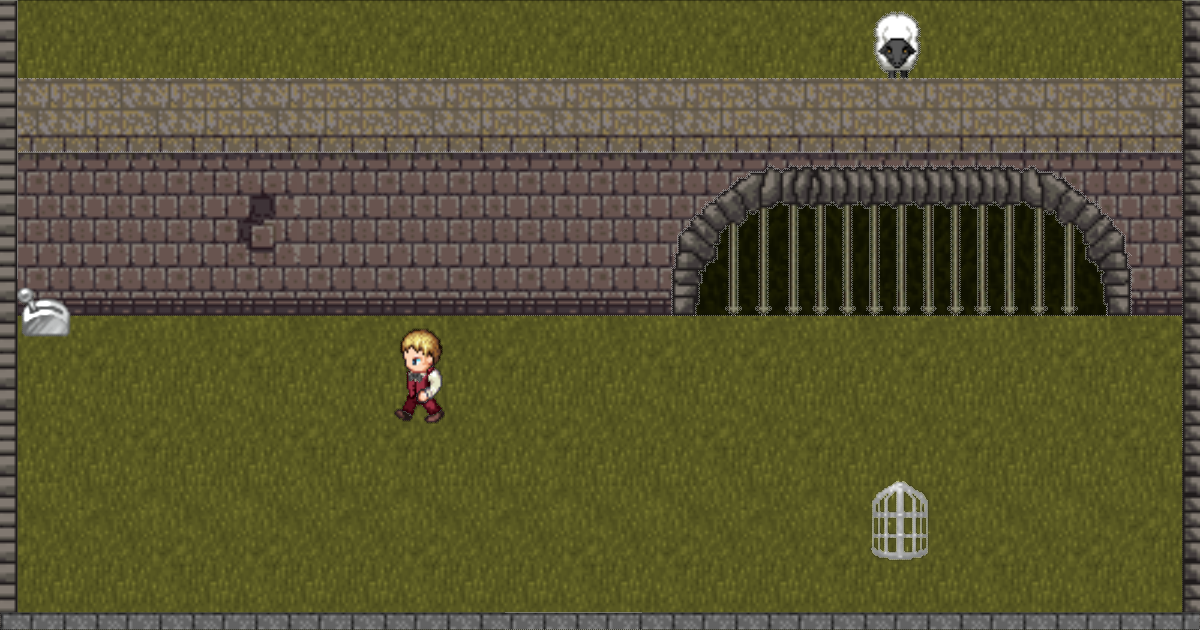}
\caption{Shepherd \label{fig:SMG}}
\end{subfigure}
\hfill
\begin{subfigure}{0.18\linewidth}
\centering
\includegraphics[width=0.9\linewidth]{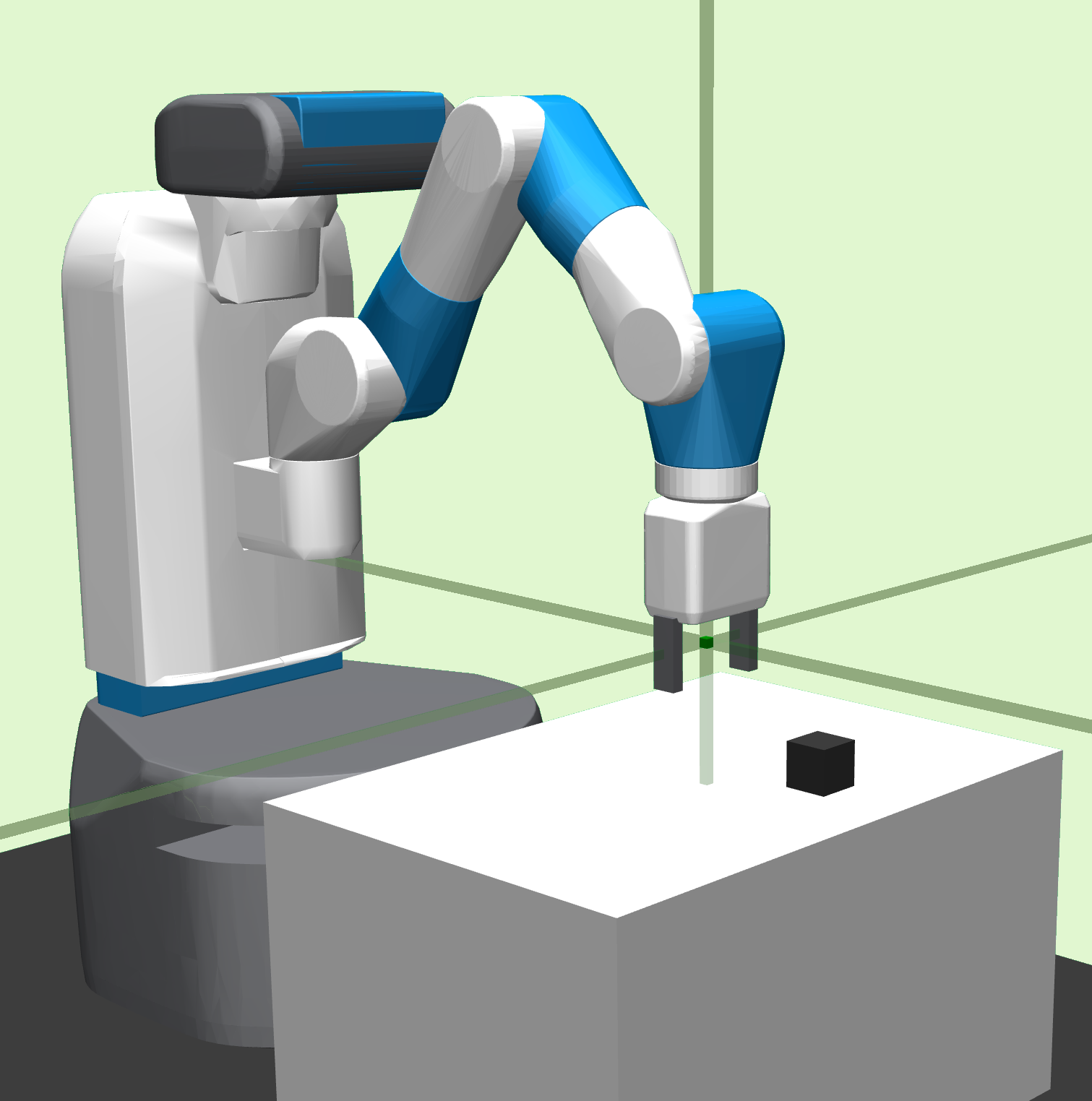}
\caption{Pick\&Place \label{fig:FPP}}
\end{subfigure}
\hfill
\begin{subfigure}{0.17\linewidth}
\centering
\includegraphics[width=\linewidth]{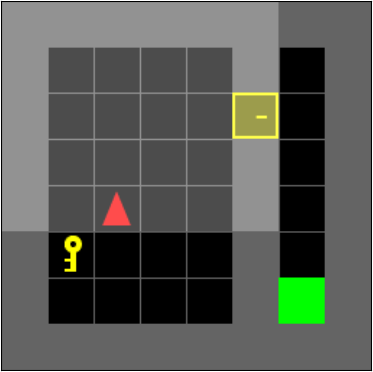}
\caption{MiniGrid \label{fig:MiniGrid}}
\end{subfigure}
\hfill \vspace{-.5em}
    \caption{Simulations used to test \method. (a) and (b) are continuous 2D-control tasks: (a)~requires triggering the control of a robot by getting a remote control; (b)~needs memorization of the sheep's position to capture it later. 
    (c)~is the Fetch Pick\&Place environment \citep{brockman2016openai} modified to become partially observable and (d) shows a problem (DoorKey-8x8) of the Mini-Gridworld suite \citep{MiniGrid}.
    }
    \label{fig:scenarios}
\end{figure*}

We evaluate \method in a variety of partially observable scenarios. 
In the \textbf{Billiard Ball} scenario a single ball, simulated in a realistic physics simulator, is shot on a pool table with low friction from a random position in a random direction with randomly selected velocity. 
The time series contain only the positions of the ball. This is the only considered scenario without actions. 

\textbf{Robot Remote Control} is a continuous control problem where an agent moves according to the two-dimensional actions $\bm{a}_t$ (\fig{fig:RRC}).
Once the agent reaches a fixed position (terminal), a robot in another room is also controlled by the actions.
The observable state $\bm{o}_t$ is composed of the agent's position and the robot's position.
Thus, whether the robot is controlled or not is not observable directly.
When planning, the goal is to move the robot to a particular goal position (orange square).

\textbf{Shepherd} is a challenging continuous control problem that requires long-term memorization (\fig{fig:SMG}).
Here, the agent's actions $\bm{a}_t$ are the two movement directions and a grasp action controlling whether to pick up or drop the cage.
The sheep starts at the top of the scene moving downwards with a fixed randomly generated velocity.
The sheep is then occluded by the wall, which masks its position from the observation.
If the agent reaches the lever, the gate inside the wall opens, and the sheep appears again at the same horizontal position at the open gate.
The goal is to get the sheep to enter the previously placed cage.
The challenge is to memorize the sheep's horizontal position exactly over a potentially long time to place the cage properly and to then activate the lever during mental simulation. 
The seven-dimensional observation $\bm{o}_t$ is composed of the height of the occluder and the positions of all entities.

\textbf{Fetch Pick\&Place} (OpenAI Gym v1, \citep{brockman2016openai}) is a benchmark RL task where a robotic manipulator has to move a randomly placed box (\fig{fig:FPP}). 
In our modified setting\footnote{
We omit all velocities and the rotation of the object to make the scenario partially observable.}\!\!, the observable state $\bm{o}_t$ is composed of the gripper- and object position and the relative positions of object and fingers with respect to the gripper.
The four-dimensional actions $\bm{a}_t$ control the gripper position and the opening of the fingers.

\textbf{MiniGrid} \cite{MiniGrid} is a gridworld suite with a variety of partially observable RL problems.
At every time $t$, the agent (red triangle in  \fig{fig:MiniGrid}) receives an image-like, restricted, ego-centric view (grey area) as its observation $\bm{o}_t$ ($7 \times 7 \times 3$-dimensional).
It can either move forward, turn left, turn right, or interact with objects via its one-hot-encoded actions $\bm{a}_t$.
The problems vary largely in their difficulty, typically contain only sparse rewards, and often involve memorization, e.g., remembering that the agent picked up a key.
\supp{sec:SupplMiniGrid} details all examined MiniGrid environments.

\subsection{Learning autoregressive predictions} \label{sec:resBilliard}
First,  we consider the problem of autoregressive $N$-step prediction in the \textbf{Billiard Ball} scenario.
Here, during testing the networks receive the first two ball positions as input and predict a sequence of 50 ball positions.
We first train the RNNs using \emph{teacher forcing},
whereby the real inputs are fed to the networks.
\Fig{fig:resTeacherForcing} shows the prediction error for autoregressive predictions. 
Only \method with latent state regularization ($\lambda=0.001$) is able to achieve reasonable predictions in this setup.
The other RNNs seem to learn to continuously update their estimates of the ball's velocity based on the real inputs. 
Because \method punishes continuous latent state updates, learning leads to updates of the estimated velocity only when required, \ie upon collisions, improving its prediction robustness.

\begin{figure*}
  \startsubfig{}
  \begin{tabular}{@{}l@{\ \ }l@{\ \ }l@{\ \ }l@{}}
    \subfig{fig:resTeacherForcing} & \subfig{fig:resSS} &\subfig{fig:predErrorLambdas}& \subfig{fig:gateRateLambdas}\\[-1em]
    \includegraphics[width=0.24\linewidth]{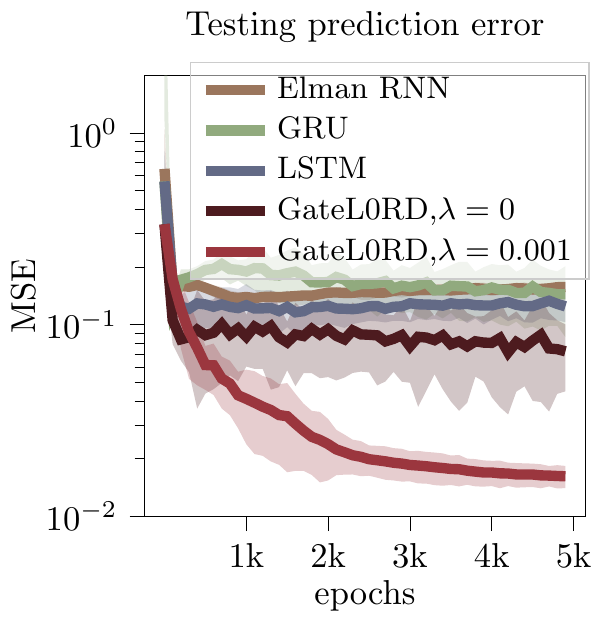} &
    \includegraphics[width=0.24\linewidth]{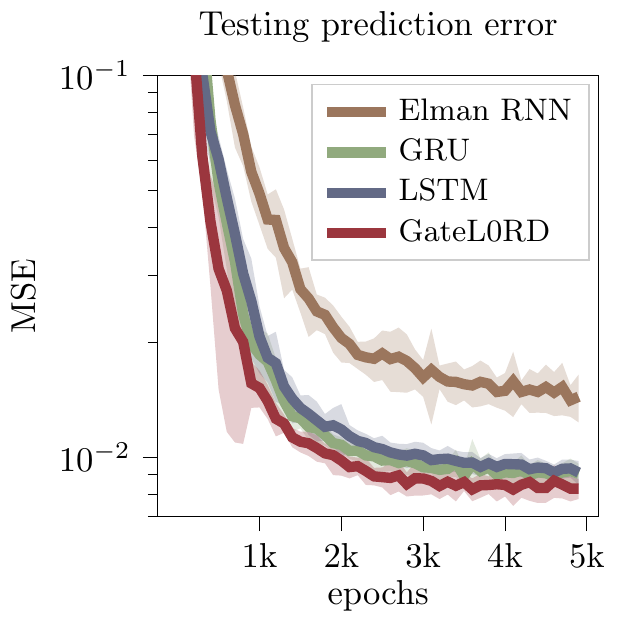} &
    \includegraphics[width=0.24\linewidth]{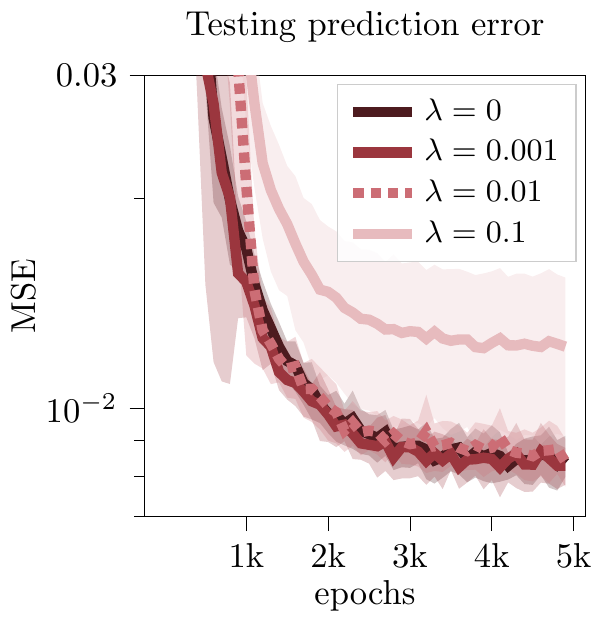} &
    \includegraphics[width=0.24\linewidth]{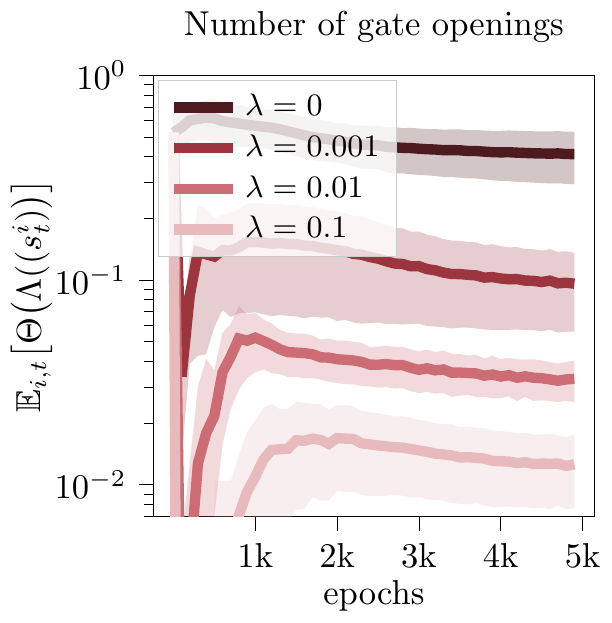}
  \end{tabular}\vspace{-1em}
  \caption{Billiard Ball results: prediction errors when trained using teacher forcing (a), or using scheduled sampling (b).
  \method's prediction error (c) and mean number of gate openings (latent state updates) (d) for different values of $\lambda$. Shaded areas show $\pm$ one standard deviation.   \label{fig:Billiard}}
\end{figure*}

The problems of RNNs learning autoregressive prediction are well known \cite{scheduledSampling, ProfessorForcing}. 
A simple countermeasure is \emph{scheduled sampling} \cite{scheduledSampling}, where each input is stochastically determined to be either the last network's output or the real input.
The probability of using the network output increases over time.
While the prediction accuracy of all RNNs improves when trained using scheduled sampling, \method ($\lambda=0.001$) still achieves the lowest mean prediction error (see \fig{fig:resSS}).

How does the regularization affect \method? 
\Fig{fig:predErrorLambdas} shows the prediction error for \method for different settings of $\lambda$.
While a small regularization ($\lambda=0.001$) leads to the highest accuracy in this scenario, similar predictions are obtained for different strengths ($\lambda \in [0, 0.01]$).
Overly strong regularization ($\lambda=0.1$) degrades  performance.
\Fig{fig:gateRateLambdas} shows the average gate openings per sequence. 
As indented, $\lambda$ directly affects how often \method's latent state is updated: a higher value results in fewer gate openings and, thus, fewer latent state changes.
Note that even for $\lambda=0$ \method learns to use fewer gates over time. We describe this effect in more detail in \supp{sec:SuppBBLatent}.

\subsection{Generalization across policies} \label{sec:resRRC}

Particularly when priorities change or an agent switches behavior, different spurious temporal correlations can occur in the resulting sensorimotor timeseries data.
Consequently, models are needed that generalize across those correlations.
We use the networks trained as predictive models for the \textbf{Robot Remote Control} scenario to investigate this aspect. 

In Robot Remote Control the training data is generated by performing rollouts with 50 time steps of a policy that produces random but linearly magnitude-increasing actions.
The actions' magnitude in the training data is positively correlated with time, which is a spurious correlation that does not alter the underlying transition function of the environment in any way.
We train the networks to predict the sequence of observations given the initial observation and a sequence of actions.
Thereby, we test the networks using data generated by the same policy (\emph{test set}) and generated by a policy that samples uniformly random actions (\emph{generalization set}).
Additionally, we use the trained RNNs for model-predictive control (MPC) using iCEM \cite{PinneriEtAl2020:iCEM}, a random shooting method that iteratively optimizes its actions to move the robot to the given goal position.

\begin{figure*}
    \startsubfig{}
  \begin{tabular}{@{}l@{\ \ }l@{\ \ }l@{\ \ }l@{}}
    \subfig{fig:resRRCTesting} & \subfig{fig:resRRCGeneralization}&\subfig{fig:resRRCPlan}& \subfig{fig:RRCLatentStates}\\[-1em]
   \includegraphics[width=.2\linewidth]{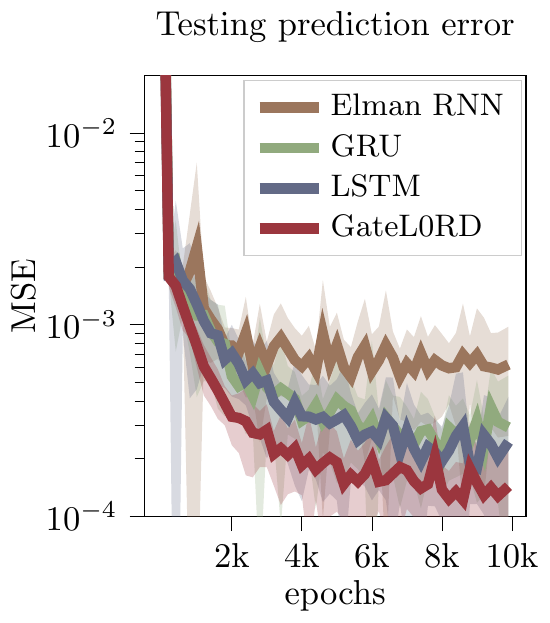}&
    \includegraphics[width=.2\linewidth]{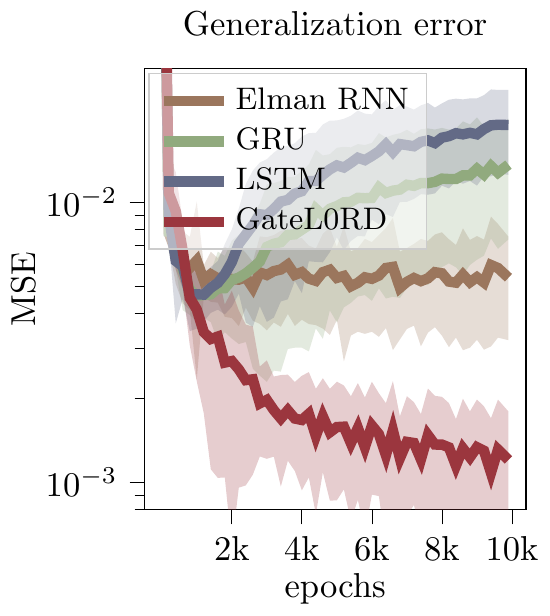}&
    \includegraphics[width=.2\linewidth]{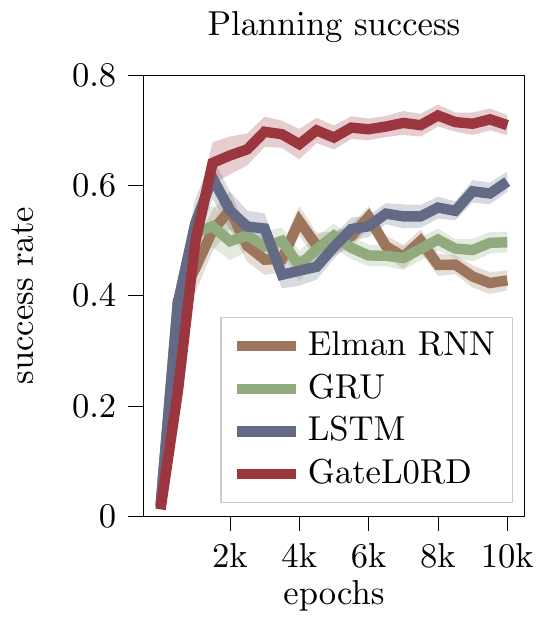}&
    \includegraphics[width=.36\linewidth]{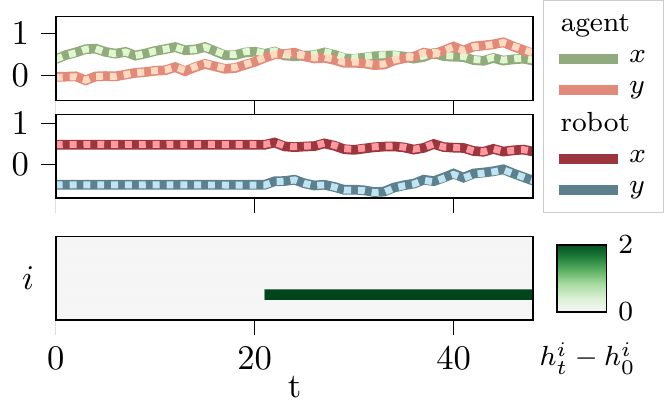}
  \end{tabular}\vspace{-1em}
  \caption{Robot Remote Control results: prediction error on the test set (a) and on the generalization set (b). Success rate for MPC (c). Shaded areas show standard deviation (a \& b) or standard error (c). Exemplary generalization sequence (d) showing the agent's positions (top), the robot's positions (middle) with \method's predictions shown as dots, and \method's latent states (bottom).}
\end{figure*}

As shown in \fig{fig:resRRCTesting}, \method ($\lambda = 0.001$) outperforms all other RNNs on the test set.
When tested on the generalization data, the prediction errors of the GRU and LSTM networks even increase over the course of training. Only \method is able to maintain a low prediction error. 
\Fig{fig:resRRCPlan} shows the MPC performance.
\method yields the highest success rate.

Note that the lack of generalization is not primarily caused by the choice of hyperparameters:
even when the learning rate of the other RNNs was optimized for the generalization set, GateL0RD still outperformed them (additional experiment in \supp{sec:RRCGenLRsExperiment}).
Instead, \method's better performance is likely because it mostly encodes unobservable information within its latent state $\bm h_t$.
This is shown exemplarily in \fig{fig:RRCLatentStates} (bottom row) and analyzed further in \supp{sec:SupplLatentStatesRRC}.
The latent state remains constant and only one dimension changes once the agent controls the robot's position (middle row) through its actions.
Because the other RNNs also encode observable information, \eg actions, within their latent state, they are more negatively affected by distributional shifts and spurious dependencies.

\method's improved generalization across temporal dependencies also holds for more complicated environments.
In an additional experiment in \supp{sec:SuppResFPPGen} we show similar effects for the \textbf{Fetch Pick\&Place} environment when trained on reach-grasp-and-transport sequences and tested to generalize across timings of the grasp.

\subsection{Long-term memorization} \label{sec:resShepherd}

\begin{figure*}[b]
  \startsubfig{}
  \begin{tabular}{@{}l@{\ \ }l@{\ \ }l@{}}
    \subfig{fig:SMGPredError} & \subfig{fig:sheepError}&\subfig{fig:SMGPlan}\\[-1em]
    \includegraphics[width=0.32\linewidth]{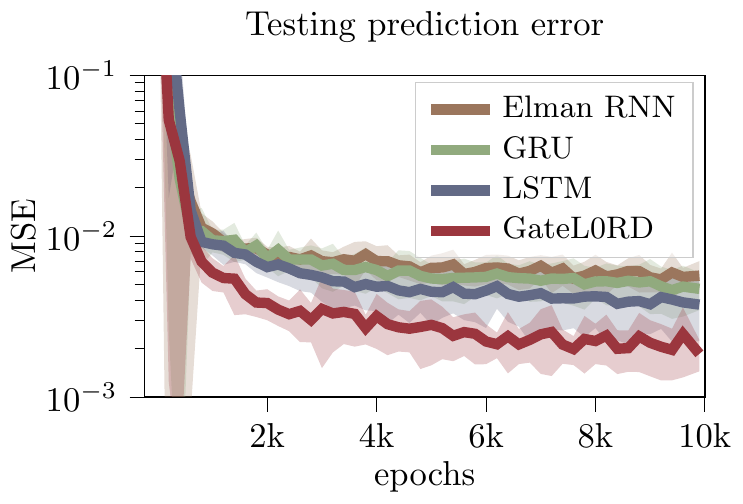}&
    \includegraphics[width=0.32\linewidth]{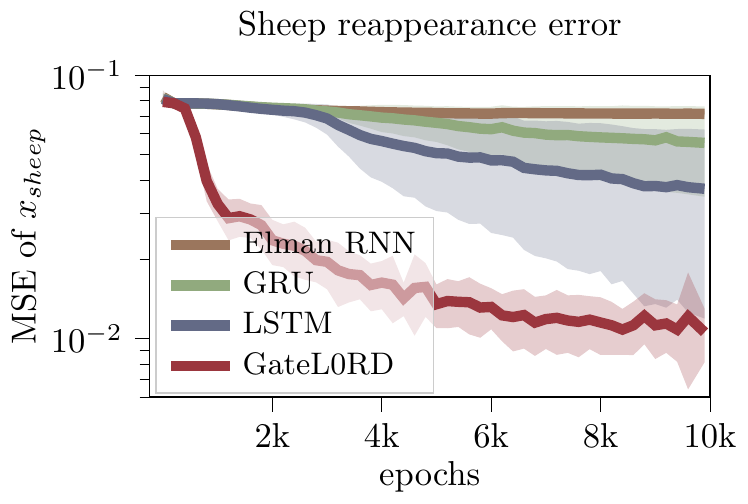}&
    \includegraphics[width=0.32\linewidth]{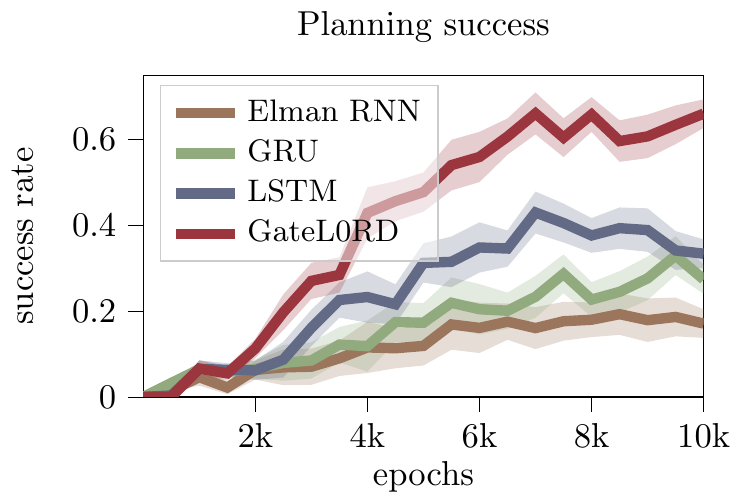}
  \end{tabular}\vspace{-1em}
  \caption{Shepherd results: prediction error for 100-step predictions (a) and 1-step prediction errors of the sheep's $x-$position at the time step of reappearance (b). Success rate for capturing the sheep using MPC (c). Shaded areas show standard deviation (a-b) or standard error (c).}
\end{figure*}

We hypothesized that \method's latent state update strategy fosters the exact memorization of unobservable information, which we examine in the \textbf{Shepherd} task.
We test the RNNs' when predicting sequences of 100 observations given the first two observations and a sequence of actions.
Again, we use the trained models for MPC using iCEM \cite{PinneriEtAl2020:iCEM}, aiming at catching the sheep by first placing a cage and then pulling a lever.
This is particularly challenging to plan because the sheep's  horizontal position needs to be memorized before it is occluded for quite some time ($>30$ steps) in order to accurately predict and thus place the cage at the sheep's future position.

\Fig{fig:SMGPredError} shows the prediction errors during training.
\method ($\lambda = 0.0001$) continuously achieves a lower prediction error than the other networks.
Apparently, it is able to accurately memorize the sheep's future position while occluded.
To investigate the memorization we consider the situation occurring during planning: the sequence of (past) observations is fed into the network and the prediction error of the sheep's horizontal position at the time of reappearance is evaluated (\fig{fig:sheepError}).
Only \method reliably learns to predict where the sheep will appear when the lever is activated.
GRU and Elman RNNs do not noticeably improve in predicting the sheep's position.
LSTMs take much longer to improve their predictions and do not reliably reach \method's level of accuracy.
This is also reflected in the success rate when the networks are used for MPC (\fig{fig:SMGPlan}).
Only GateL0RD manages to solve this challenging task with a mean success rate over 50\%.

\subsection{Sample efficiency in reinforcement learning}
\label{sec:resMiniGrid}
Now that we have outlined some of \method's strengths in isolation, we want to analyze whether \method can improve existing RL-frameworks when it is used as a memory module for POMDPs.
To do so, we consider various problems that require memory in the \textbf{MiniGrid} suite \cite{MiniGrid}.
Previous work \cite{BabyAI, RIMs, MetaRIMs} used Proximal Policy Optimization (PPO) \cite{PPO} to solve the MiniGrid problems.
We took an existing architecture based on \cite{BabyAI} (denoted as \emph{vanilla}, detailed in \supp{sec:SupplRL}) and replaced the internal LSTM module with \method ($\lambda = 0.01$).
Note, that we left the other hyperparameters unmodified.

\begin{figure*}
    \startsubfig{}
  \begin{tabular}{@{}l@{\ \ }l@{\ \ }l@{\ \ }l@{\ \ }l@{\ \ }l@{}}
    \subfig{} & \subfig{}&\subfig{ }& \subfig{ }&\subfig{ }& \subfig{ }\\[-1em]
    \includegraphics[width=.16\linewidth]{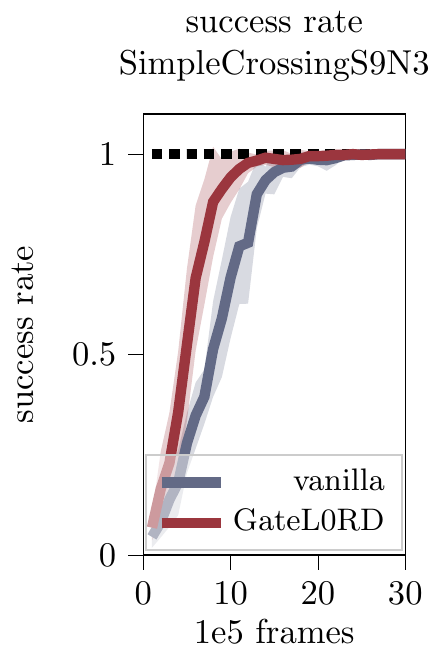}&
    \includegraphics[width=.16\linewidth]{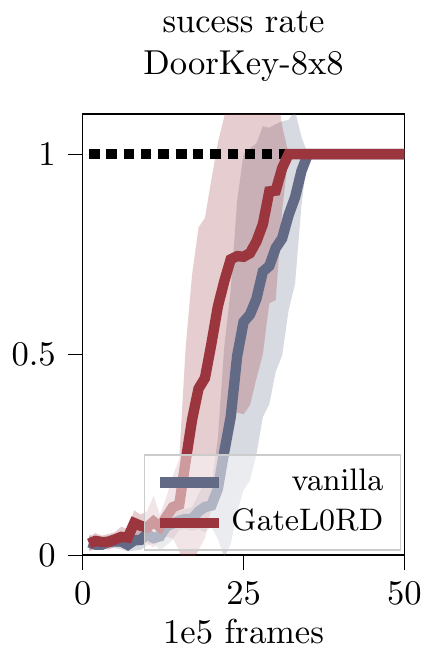}&
    \includegraphics[width=.16\linewidth]{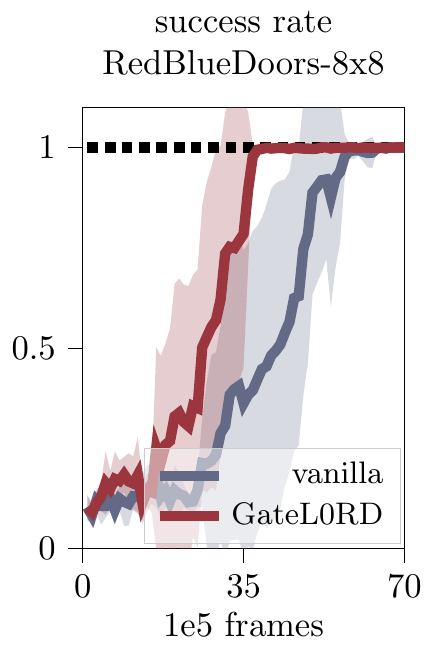}&
    \includegraphics[width=.16\linewidth]{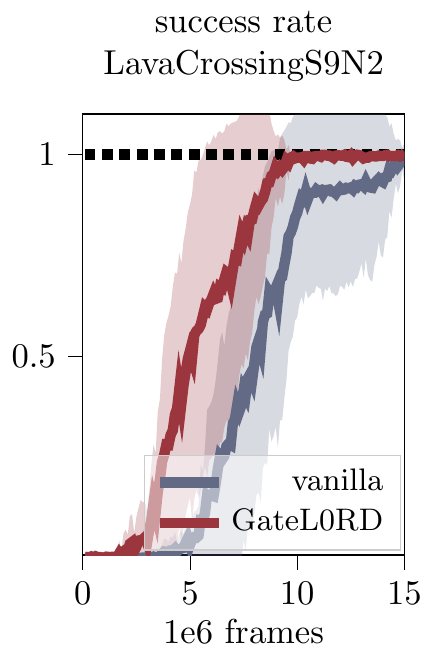}&
    \includegraphics[width=.16\linewidth]{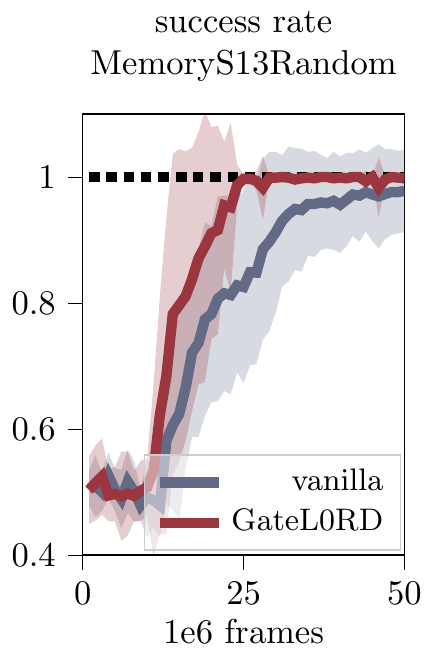}&
    \includegraphics[width=.16\linewidth]{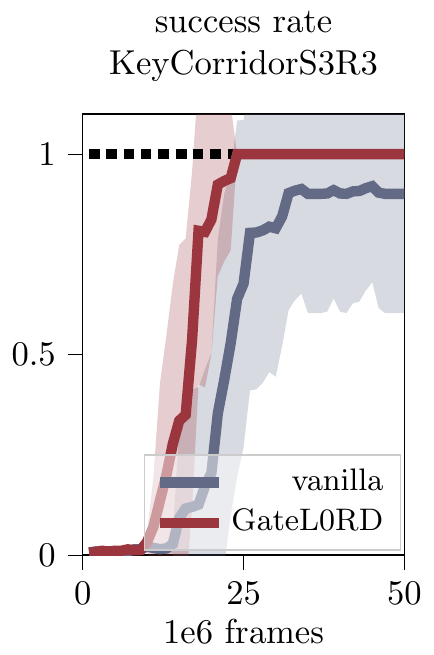}
  \end{tabular}\vspace{-1em}
  \caption{MiniGrid results: success rate in solving various tasks when GateL0RD replaces an LSTM (vanilla) in a PPO architecture. Shaded areas depict the standard deviation. \label{fig:MiniGridSuccess}}
\end{figure*}

As shown in \fig{fig:MiniGridSuccess} the architecture containing \method achieves the same success rate or higher than the vanilla baseline in all considered tasks. 
Additionally, \method is more sample efficient, \ie, it is able to reach a high success rate (\fig{fig:MiniGridSuccess}) or high reward level faster (\supp{sec:SuppMiniGridRewards}).
The difference in sample efficiency tends to be more pronounced for problems that require more training time.
It seems that the inductive bias of sparsely changing latent states enables \method to quicker learn to encode task-relevant information, such as the pick-up of a key, within its latent states.
Additional experiments in \supp{sec:SuppMiniGridZeroShot} show that this can also translates to improved zero-shot policy transfer, when the system is tested on a larger environment than it was trained on.

\subsection{Explainability of the latent states} \label{sec:resExplainability}

Lastly, we analyze the latent representations of \method, starting with \textbf{Billiard Ball}.
\Fig{fig:BBtraj} shows one exemplary ball trajectory in white and the prediction in red. Inputs for which at least one gate opened are outlined in black. \Fig{fig:BBlatent} shows the corresponding latent states $\bm h_t$ relative to the initial latent state $\bm h_0$.
\method updates two dimensions of its latent states around the points of collisions to account for the changes in $x$- and $y$-velocity of the ball. 
For $\lambda=0.01$ we find on average only two latent state dimensions change per sequence (see Suppl \supp{sec:SuppBBLatent}), which hints at a tendency to encode $x$- and $y$-velocity using separate latent dimensions.
In contrast, the exemplary latent states of the GRU and LSTM networks shown in \fig{fig:BBlatent} are not as easily interpretable.

\begin{figure*}[]
  \startsubfig{}
  \begin{minipage}[t]{.34\linewidth}\centering
    \subfig{fig:BBtraj} Billiard Ball trajectory\\
    \vspace*{0.3cm}
    \includegraphics[width=\linewidth]{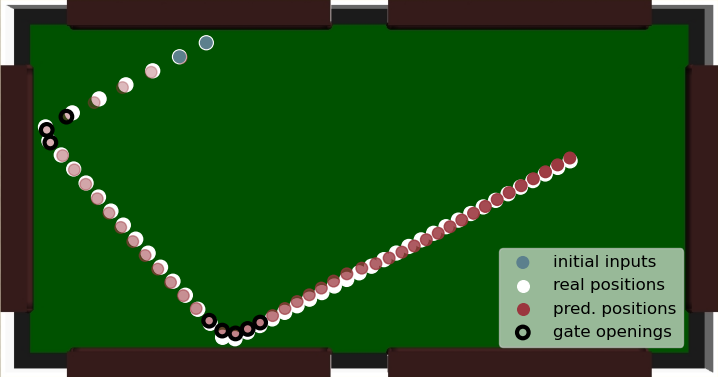}
  \end{minipage}\hfill
  \begin{minipage}[t]{.34\linewidth}\centering
    \subfig{fig:BBlatent}: Billiard Ball latent states\\
    \includegraphics[width=\linewidth]{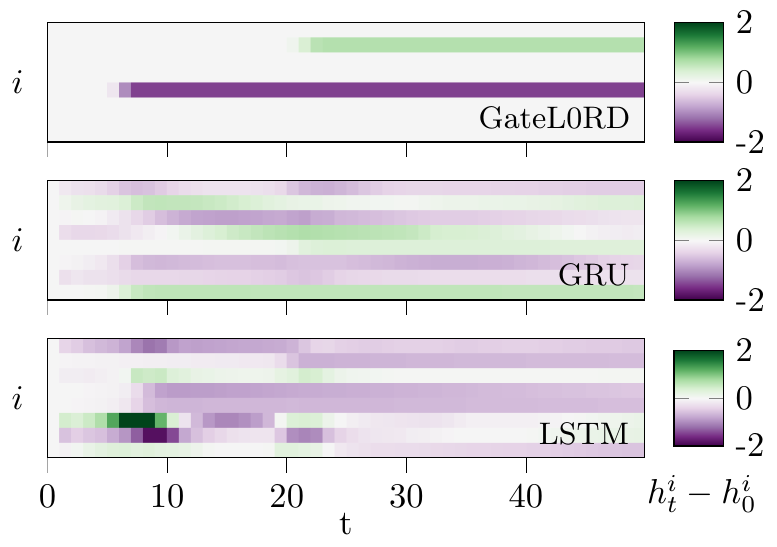}
  \end{minipage}\hfill
  \begin{minipage}[t]{.3\linewidth}\centering
    \subfig{fig:FPPLatent}: Fetch Pick\&Place sequence\\
    \includegraphics[width=\linewidth]{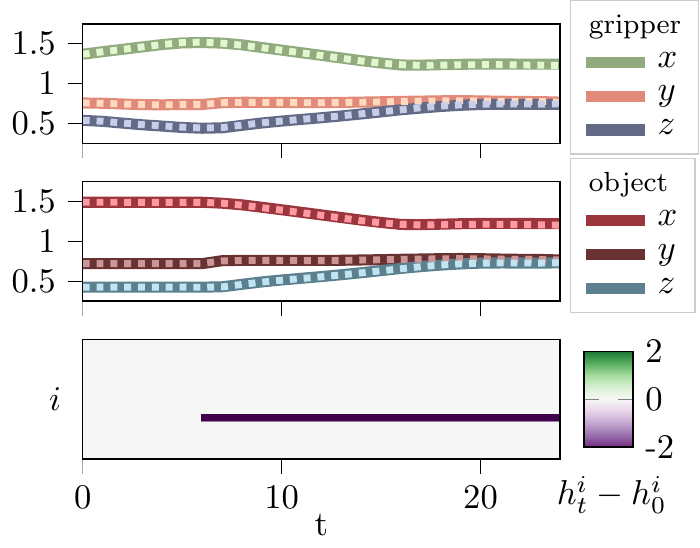}
  \end{minipage}

  \caption{Example sequences and latent states: (a) Billiard Ball trajectory for \method ($\lambda=0.01$) with real positions (white), provided inputs (blue), and predicted positions (red, saturation increasing with time). The inputs for which at least one gate opened are outlined in black. (b) The latent states for the trajectory  for \method, GRU, and LSTM (cell states). (c) Fetch Pick\&Place sequence with real (solid) and predicted (dotted) positions of gripper (top) and object (middle) and \method's latent states (bottom). Latent states are shown relative to initialization, \ie $\bm h_t - \bm h_0$.}
\end{figure*}

For \textbf{Robot Remote Control}, \method ($\lambda=0.001$) updates only its latent state once it controls the robot (exemplary shown in \fig{fig:RRCLatentStates}).
Thus, the latent state clearly encodes control over the robot. 
We use the \textbf{Fetch Pick\&Place} scenario as a higher-dimensional problem to investigate latent state explainability when training on grasping sequences (detailed in \supp{sec:SuppFPP}).
Here, \method  updates the latent state typically when the object is grasped (exemplary shown in \fig{fig:FPPLatent}).
This hints at an encoding of `object transportation' using one dimension.
Other RNNs do not achieve such a clear representation, neither in Robot Remote Control nor in Fetch Pick\&Place (see \supp{sec:SupplLatentStatesRRC} and \ref{sec:SuppResFPPGen}).

\section{Discussion} \label{sec:discussion}
We have introduced a novel RNN architecture (\method), which implements an inductive bias to develop sparsely changing latent states.
The bias is realized by a gating mechanism, which minimizes the $L_0$ norm of latent updates.
In several empirical evaluations, we quantified and analyzed the performance of \method on various prediction and control tasks, which naturally contain piecewise constant, unobservable states. 
The results support our hypothesis that networks with piecewise constant latent states can generalize better to distributional shifts of the inputs, ignore spurious time dependencies, and enable precise memorization.
This translates into improved performance for both model-predictive control (MPC) and reinforcement learning (RL). 
Moreover, we demonstrated that the latent space becomes interpretable, which is important for explainability reasons.

Our approach introduces an additional hyperparameter, which controls the trade-off between the task at hand and latent space constancy.
When chosen in favor of explainability, it can reduce the in-distribution performance while improving its generalization abilities.
When the underlying system has continuously changing latent states, our regularization is counterproductive. 
As demonstrated by an additional experiment in \supp{sec:resFPP}, the unregularized network performs well in such cases.

Our sparsity-biased gating mechanism segments sequences into chunks of constant latent activation.
These segments tend to encode unobservable, behavior-relevant states of the environment, such as if an object is currently `under control'.
Hierarchical planning and control methods require suitable, temporally-extended encodings, such as options \cite{Sutton1999:Options-framework, barto2003recent}.
Thus, a promising direction for future work is to exploit the discrete hidden dynamics of \method for hierarchical, event-predictive planning.

\begin{ack}
The authors thank the International Max Planck Research School for Intelligent Systems (IMPRS-IS) for supporting Christian Gumbsch. Georg Martius and  Martin Butz are members of the Machine Learning Cluster of Excellence, EXC number 2064/1--project number 390727645.
We acknowledge the support from
the German Federal Ministry of Education and Research through the Tübingen AI Center (FKZ: 01IS18039B).
This research was funded by the German Research Foundation (DFG) within Priority-Program ``The Active Self'' SPP 2134--project BU 1335/11-1.
The authors thank Maximilian Seitzer for the helpful feedback and  Sebastian Blaes for the help in applying iCEM.
\end{ack}

\nocite{gradientNormClipping}

\bibliography{neurips_2021}

\newpage

\appendix

\begin{center}
    \Large\bf  Supplementary Material for:\\
Sparsely Changing Latent States for Prediction and Planning\\ in Partially Observable Domains 
\end{center}
\section{Relation to other RNNs}

In \sec{sec:l0} we set out to create an RNN $\netth$ that maintains piecewise constant latent states over time.
This led us to the conclusion, that a simple approach to implement this is by employing an internal gating function $\Lambda$ that controls the latent state update (\eg as in \eqn{eq:latent-update}).
The gating function $\Lambda$ can be binarized using the Heaviside step function and a sparse gating can be incentivized using the loss function outlined in \eqn{eq:loss2}.

GRUs \cite{GRU1} and LSTMs \cite{LSTM1} both use internal gates with the sigmoid activation function $\sigma$ to control the update of their latent state $\bm{h}_t$. GRUs update  $\bm{h}_t$ with
\begin{align}
    \bm h_{t} &= (1- \sigma(\bm s_t))\bm h_{t-1} + \sigma(\bm s_t)\bm{\tilde h}_t,
\end{align}
where $\bm s_t$ is a linear projection of input $\bm x_t$ and previous latent state $\bm h_{t-1}$ and $\bm{ \tilde h}_t$ is a proposed new latent state, also determined based on the input $\bm x_t$ and previous latent state $\bm h_{t-1}$.

LSTMs use two gates, \ie a forget and an input gate, with the sigmoid activation function $\sigma$ to determine whether to update their latent (cell) state $\bm h_t$ with
\begin{align}
    \bm h_{t} &= \sigma(\bm s_{1, t})\bm{h}_{t-1} + \sigma(\bm s_{2, t})\bm{\tilde h}_t,
\end{align}
where $\bm s_{1, t}$ and $\bm s_{2, t}$ are linear projections and $\bm{\tilde h}_t$ a non-linear function of the input and previous hidden state (RNN cell output).

Nonetheless, it is not straight forward to apply our approach, outlined in \sec{sec:l0}, to GRUs and LSTMs.
Our loss (see \eqn{eq:loss2}) punishes non-zero gate activation. 
The sigmoid activation function $\sigma$ only achieves an output of zero if its input converges to negative infinity, thus, never truly achieving zero output.
Thus, their gating function would need to be modified or replaced, \eg by our ReTanh gate $\Lambda$. 

However, even when replacing their gate activation function, the performance of LSTMs and GRUs are negatively affected by piecewise constant latent states.
For both networks, input information essentially needs to pass through the latent state to affect the network output.
For GRUs the network output corresponds to the latent state $\bm{h}_t$. 
Thus, a GRU with constant latent states will produce constant outputs.
In LSTMs the network output is computed by multiplying the latent (cell) state with an input-dependent output gate.
Thus, in LSTMs a constant latent state will result in a constant output that is scaled depending on the network input.

\method attempts to overcome the outlined downsides of using LSTMs and GRUs with our proposed latent state regularization. 
Like GRUs, \method uses a single update gate to avoid unnecessary parameters. 
Additionally, \method separates the latent state from the network output, as done in LSTMs, which have both a cell state and a hidden state.
Besides that, \method uses more powerful functions for computing the network output such that input and latent state both have an additive as well as multiplicative effects on the network output. 
Note that \method still has approximately the same number of parameters as a GRU.

\label{sec:SupplRNNs}

\section{Experimental Details}
\label{sec:suppExpDetails}

\subsection{Predictive Models: General Training Principles and Hyperparameter Search}
\label{sec:SuppTrainingDetails}

In the following, we will outline the general training principles that we used in all experiments, when the RNNs were trained as \emph{predictive models}.
Training details for the reinforcement learning experiments are found in \supp{sec:SupplRL}.
\supp{sec:SuppBB} - \ref{sec:SuppFPP} provide further details specific to each simulation independent of the hyperparameter search (\eg dataset size, batch size, etc.).

In our experiments, we train each network to predict the change in observations instead of the next observation (\ie residual connections) to avoid the trivial solution of achieving a high prediction accuracy by simply outputting the input observation. However, since the change in observation can be quite small (typically $\Delta\bm{o}_t  < 0.1$) we use a constant $c$ to scale the network output when used as autoregressive input, \ie $\hat{\bm o}_{t+1} = \bm{o}_{t} + c \cdot \bm{\hat{y}}_t$.
We set $c = 0.1$ in all our experiments, which corresponds to scaling $\Delta \bm{o}_t$ by a factor of 10.
For the task-based loss, \ie $\mathcal{L}_{\text{task}}$ in \eqn{eq:loss}, we use the mean squared error between predicted observations $\hat{\bm o}_{t}$ and real observations  $\bm o_t$.

We train the networks using Adam \cite{kingma2014adam} with the hyperparameters $\beta_1 = 0.9$, $\beta_2 = 0.999$, and $\epsilon = 0.0001$.  The learning rate $\alpha$ was determined via a grid search with $\alpha \in \{0.005,$ $0.001,$ $0.0005,$ $0.0001, 0.00005\}$ for each scenario. For this grid search, we examined two random seeds for each parameter configuration and chose the setting resulting in the lowest mean squared prediction error on a validation set after full training. The best learning rates for all experiments are listed in \tab{table:lr}.

\begin{table*}
    \centering
    \caption{Learning rate choices}
    \label{table:lr}
    \begin{tabular}{lrrrr}
        \toprule
        \textbf{Experiment} & \textbf{GateL0RD} & \textbf{LSTM} & \textbf{GRU} & \textbf{Elman RNN}\\
        \midrule
        \textbf{Billiard Ball} teacher forcing (\sec{sec:resBilliard}) & $0.001$ & $0.001$ & $0.001$ & $0.00005$ \\
        \textbf{Billiard Ball} scheduled sampling  (\sec{sec:resBilliard}) & $0.0005$ & $0.0005$ & $0.0005$ & $0.0005$ \\
        \textbf{Robot Remote Control (RRC)}  (\sec{sec:resRRC}) & $0.005$ & $0.005$ & $0.005$ & $0.005$ \\
        \textbf{RRC} improved generalization (\supp{sec:RRCGenLRsExperiment}) & $0.005$ & $0.001$ & $0.001$ & $0.001$ \\
        \textbf{Shepherd}  (\sec{sec:resShepherd}) & $0.001$ & $0.001$ & $0.001$ & $0.001$ \\
        \textbf{Fetch Pick\&Place} filtered data (\supp{sec:SuppResFPPGen}) & $0.005$ & $0.005$ & $0.005$ & $0.005$ \\
        \textbf{Fetch Pick\&Place} full data (\supp{sec:resFPP}) & $0.001$ & $0.001$ & $0.001$ & $0.001$ \\
        \bottomrule
        \end{tabular}
\end{table*}

Besides determining the learning rate, we also use grid search to determine the number of RNN layers for all scenarios with simulated physics, \ie Billiard Ball and Fetch Pick\&Place.
For LSTMs, GRUs, and Elman RNNs we compare the 1-layered RNNs to a stacked version in which up to three RNN cells ($\netth$ in \fig{fig:overall}) are composed.
For \method we instead considered 1- to 3-layered $r$ and $g$-networks (see \fig{fig:architectureCore}), since we found that this typically results in a stronger increase in performance with fewer parameters compared to stacking \method cells.
In Billiard Ball (\sec{sec:resBilliard}) and Fetch Pick\&Place (full data, \supp{sec:resFPP}), all networks achieve a slightly better mean prediction accuracy with the 3-layered versions, which is why we use the 3-layered versions to compare the prediction accuracy. 
However, for GRUs and LSTMs the 3-layered versions have three times the number of latent state dimensions, which negatively affects the interpretability of the latent states.
Thus, to make a fair comparison in terms of explainability, we additionally ran experiments with 1-layered LSTMs and GRUs to visualize the latent states (\eg in \fig{fig:BBlatent}).
For Fetch Pick\&Place with pre-selected reach-grasp-lift sequences (\supp{sec:SuppResFPPGen}) there was no noticeable improvement when increasing the number of layers, thus, we used one-layered versions of the networks.

RNNs can suffer from the exploding gradient problem when predicting long sequences.
An effective technique to deal with this is \emph{gradient norm clipping} \cite{gradientNormClipping}.
Here, the norm of a backpropagated gradient is clipped when it exceeds a threshold.
We apply gradient norm clipping in all our experiments with a clipping threshold of 0.1.

In \sec{sec:resBilliard} we showed that training the models using teacher forcing can be problematic. 
Thus, in all of our other experiments, we train the networks using scheduled sampling \cite{scheduledSampling}, a curriculum learning strategy that smoothly changes the training regime from teacher forcing to autoregressive predictions.
When applying scheduled sampling, a probability $p_i$ is used to stochastically determine whether the real input is fed into the network (teacher forcing) or whether to use the previous network output.
This sampling probability $p_i$ decreases over training time $i$.
Based on \citet{scheduledSampling}, we use an exponentially decreasing probability $p_i$ with
\begin{equation}
    p_i = \max(k^i, p_\mathrm{min})
\end{equation}
where $i$ is the epoch number, $k<1$ a constant, and $p_\mathrm{min}$ the minimum sampling probability. We set $k = 0.998$ in all experiments. The minimum sampling probability $p_\mathrm{min}$ is chosen individually for each scenario.

All experiments using predictive models were run with 20 different random seeds for each setting.

\subsection{Billiard Ball}
\label{sec:SuppBB}

In the Billiard Ball scenario, a ball is shot on a pool table with low friction.
We generated sequences of 50 time steps by shooting the ball from a random starting position in a random direction with a randomly selected velocity. 
The sequences were generated using the Open Dynamics Engine (ODE)\footnote{ODE, available at \url{http://www.ode.org/}, is licensed under the GNU Lesser General Public License version 2.1 as published by the Free Software Foundation.}---an open-source physics simulator for simulating rigid-body dynamics.
The sequences contain only the observations $\bm{o}_t \in [-1, 1]^2$, which are composed of the positions of the ball, and no actions ($\mathcal{A} = \emptyset$). 

The networks were trained on a training set of 12.8k sequences and tested on a testing set of 3.2k sequences.
Hyperparameters were determined based on a validation set of 3.2k sequences. 
All datasets were balanced to include different velocities and to guarantee that in at least $15\%$ of the sequences the ball drops into a pocket.
We trained the networks using minibatches of size 128 for 5k epochs.
We applied scheduled sampling \cite{scheduledSampling} by exponentially annealing the sampling probability $p_i$ to 0.

We used an 8-dimensional latent state $\bm{h}_t$ for all RNNs.
The latent state $\bm{h}_0$ was initialized based on the first two inputs using a 3-layered MLP $f_\mathrm{init}$  (neurons per layer: $64 \rightarrow 32 \rightarrow 16$).
All RNNs used a 3-layered MLP $f_\mathrm{pre}$ (neurons per layer: $64 \rightarrow 32 \rightarrow 16$) for preprocessing the inputs and a single linear mapping as a readout layer $f_\mathrm{post}$.

\subsection{Robot Remote Control}
\label{sec:SuppRRC}

In the Robot Remote Control scenario, an agent continuously moves through a room based on its two-dimensional actions $\bm{a}_t \in [-1, 1]^2$. 
After the agent reaches a computer, it also controls the position of a robot in another room through its actions.
The goal during planning is to move the robot to a goal area.
The observation $\bm{o}_t \in [-1, 1]^4$ is composed of the position of the agent and the position of the robot.
The robot and agent start from randomly sampled positions while the computer and goal area are always at the same fixed positions.
The robot is controlled as soon as the distance between agent and computer is below a certain interaction threshold (0.1).

We generated datasets composed of 50 time step rollouts using two synthetic policies.
The dataset $\mathcal{D}_\mathrm{time}$, containing spurious temporal dependencies, was generated by sampling uniformly distributed random actions that were scaled by a factor that linearly increases with time from 0.0001 to 1.0. 
The (generalization) dataset $\mathcal{D}_\mathrm{random}$ was generated by sampling uniformly distributed random actions without further modifications.
Both datasets were balanced in terms of robot control events, such that in half of the sequences the robot was controlled by the agent.
The datasets were split into equally sized training, validation, and testing sets (6.4k sequences each).
The validation sets were used to determine hyperparameters. 
The networks were trained for 5k epochs using minibatches of size 128.
We trained the networks using scheduled sampling \cite{scheduledSampling} by exponentially annealing the sampling probability $p_i$ to a minimum value of $p_\mathrm{min} = 0.02$.

In this scenario, the latent states $\bm{h}_t$ of all RNNs were 8-dimensional and were initialized based on the first input using a 2-layered MLP $f_\mathrm{init}$  (neurons per layer: $16 \rightarrow 8$).
All RNNs used a 3-layered preprocessing $f_\mathrm{pre}$ (neurons per layer: $32 \rightarrow 16 \rightarrow 8$) and a linear mapping $f_\mathrm{post}$ from the RNN cell output to the overall output.

During planning, the goal was to move the robot to the goal area (distance < 0.15) within 50 time steps.
For model-based planning, we used iCEM \cite{PinneriEtAl2020:iCEM}.
We left the default hyperparameters as outlined in \citet{PinneriEtAl2020:iCEM}, but used the same planning horizon of 50 time steps as during training and simulated 256 trajectories per optimization step. Additionally, we used colored noise with $\beta=3$. 
The cost was defined as the distance between robot and goal area.
We found that iCEM, which was previously used with the ground truth simulator as a model \cite{PinneriEtAl2020:iCEM}, was relatively sensitive towards model errors, resulting in the agent often slightly missing the computer or stepping over it without activating the robot.
To avoid floor effects based on the planning method, we simplified the task during planning by increasing the radius to interact with the computer by $50\%$.

\subsection{Shepherd}
\label{sec:SuppSMG}

\begin{figure}
    \centering
    \includegraphics[width=0.7\linewidth]{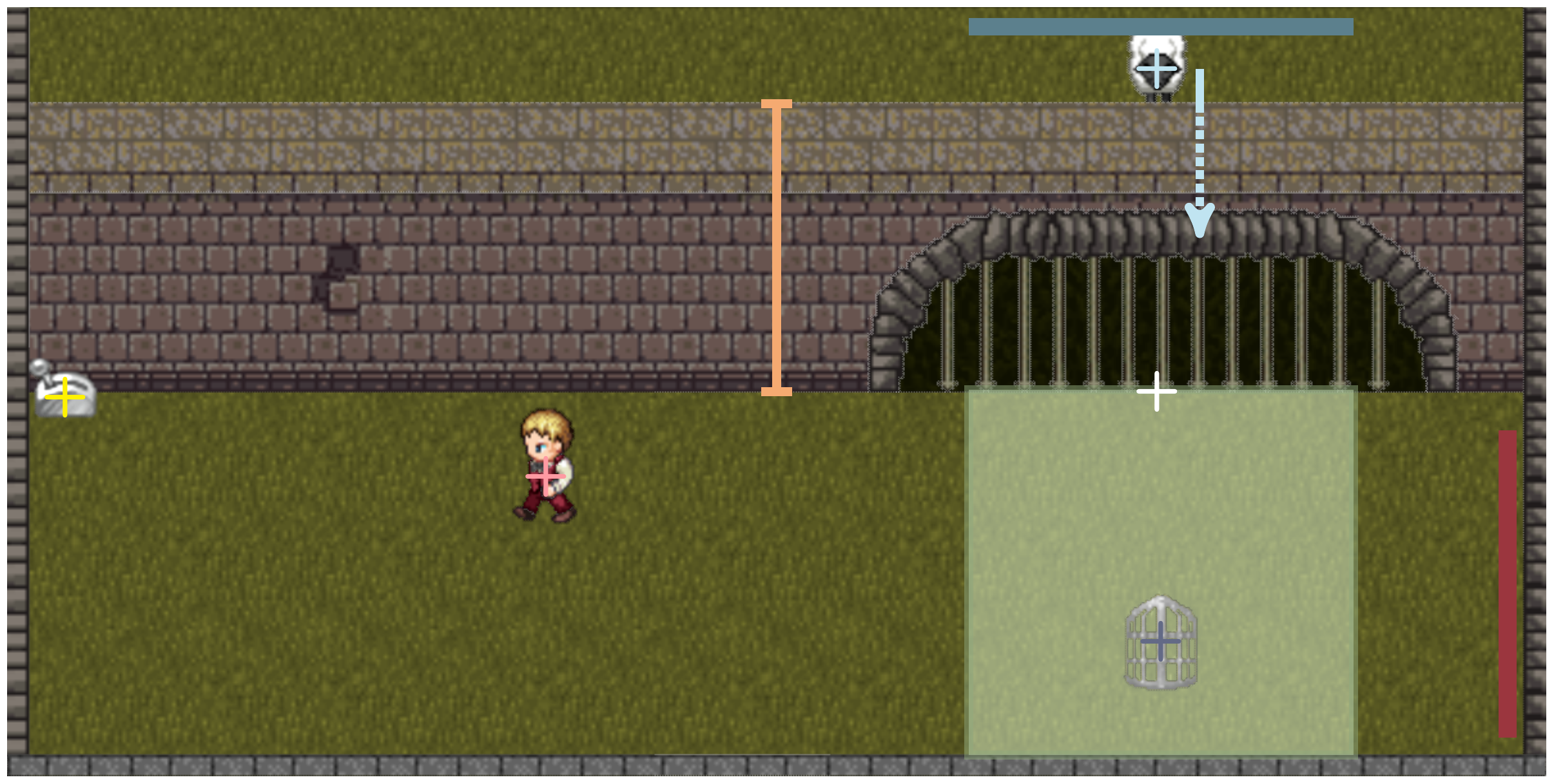}
    \caption{Shepherd scenario. Relevant positions are marked by a plus (+), with the lever in yellow, the agent in pink, the sheep in cyan, its reapearrance position in white, and the cage in purple. The orange bar visualized the wall height. The blue line and red line illustrate the sheep's and agent's starting position, respectively. The green area illustrates the cost function used for model-based planning. See text for more details.}
    \label{fig:shepherdScenario}
\end{figure}

In the Shepherd scenario, illustrated in \fig{fig:shepherdScenario}, an agent's goal is to catch a sheep using a portable cage.
The agent's actions $\bm{a}_t  \in [-1, 1]^3$ control the agent's two-dimensional movement and whether the cage is grasped and carried ($a^3_t > 0$) if it is in proximity. 
When the cage is carried, it moves with the agent. 
In every sequence, a sheep starts at the upper side of the scene (blue line  in \fig{fig:shepherdScenario}).
The sheep moves downwards with a randomly selected velocity, \ie only  changing its $y$-position (cyan arrow  in \fig{fig:shepherdScenario}).
Thereby, the horizontal $x$-position of the sheep remains the same. 
Once the sheep reaches a wall, its position is occluded from the observation. 
The height of the wall (orange bar in \fig{fig:shepherdScenario}) varies between simulations.
The agent can make the sheep reappear again by activating a lever at a fixed position (yellow + in \fig{fig:shepherdScenario}).
The lever is activated once the distance of the agent to the lever is below a certain interaction threshold.
As a result, a gate in the wall opens, causing the sheep to appear at the same horizontal position as before but at a lower vertical position (white + in \fig{fig:shepherdScenario}).
After its reappearance, the sheep moves downwards with the same  velocity as before. 
It stops moving if it reaches the cage (distance below a certain threshold) or if it reaches the lower border of the scene.
Observation $\bm{o}_t \in [-1, 1]^7$ contains the agent's position (pink + in \fig{fig:shepherdScenario}), the sheep's position (cyan +), the cage's position (purple +), and the height of the wall (orange bar). 
When the sheep is occluded, its position is masked  by replacing it with a fixed value outside the normal range of coordinates.

We generated a dataset of 100 time step sequences by using randomly sampled actions. 
In $75\%$ of the sequences up- and left-movements were sampled more frequently to get the agent to activate the lever. 
The dataset was split into training data (12.8k sequences), testing data (12.8k sequences), and validation data (6.4k sequences).
To balance the datasets across possible events, we ensured that in each dataset during $75\%$ of the sequences the lever was activated and in $25\%$ of the sequences the sheep was caught in the cage.
We trained the networks using minibatches of size 128 for 10k epochs.
We used scheduled sampling \cite{scheduledSampling} as a training regime and exponentially decreased the sampling probability $p_i$ to a minimum value of $p_\mathrm{min} = 0.05$.

All RNNs used 8-dimensional latent states $\bm{h}_t$. 
The latent state $\bm{h}_0$  was initialized based on the first two inputs using a 3-layered MLP $f_\mathrm{init}$  (neurons per layer: $64 \rightarrow 32 \rightarrow 16$).
All RNNs used a 3-layered preprocessing $f_\mathrm{pre}$ (neurons per layer: $64 \rightarrow 32 \rightarrow 16$) and a linear mapping $f_\mathrm{post}$ as a readout layer.

During planning, the agent started on the right side of the environment (red line in \fig{fig:shepherdScenario}) holding the cage. The agent had 60 time steps to place the cage, move to the lever to open the gate, and let the sheep enter the previously placed cage.
We chose a very short time of 60 time steps for this task to eliminate time-consuming solutions that avoid predicting the occluded sheep's future position, \eg by catching the slowly moving sheep after its reappearance by going back and replacing the cage.
For model-based planning, we used iCEM \cite{PinneriEtAl2020:iCEM} with the same parameters as in \supp{sec:SuppRRC} but predicting for a longer planning horizon of 100 time steps as during training. 
The cost was defined as the distance between the sheep and the cage, which was clipped to a large constant value when the sheep was above the gate (\ie outside of the green area in \fig{fig:shepherdScenario}).
As in \supp{sec:SuppRRC}, we increased the interaction radius of the lever and the cage during planning by $50 \%$. 

\subsection{Fetch Pick\&Place}
\label{sec:SuppFPP}

Fetch Pick\&Place is a benchmark reinforcement learning environment of OpenAI Gym\footnote{OpenAI Gym is released under MIT license.}  \cite{brockman2016openai}.
In Fetch Pick\&Place a 7 DoF robotic arm with a two-fingered gripper is position-controlled through its four-dimensional action.
The state of the scenario $\bm{s}_t \in \mathbb{R}^{25}$ is composed of the positions of the endeffector and the object, the relative position between endeffector and object, the distance of the fingers to the center of the gripper, the rotation of the object, and the positional and rotational velocities of the endeffector, the object, and the fingers.
To make the scenario partially observable, we omitted positional and rotational velocities as well as the rotation of the object in the observation $\bm{o}_t \in \mathbb{R}^{11}$.
The four-dimensional actions $\bm{a}_t \in [-1, 1]^4$ control the three-dimensional position of the endeffector and the closing or opening of the fingers.
Internally, the position control of the endeffector is realized by a PID-controller that runs at a higher frequency.

We generated our data consisting of sequences using APEX \cite{pinneri2021:strong-policies}, a policy-guided model predictive control method, which was trained to move the object to a random goal position. 
APEX was deployed using the ground truth simulator as the internal model and hyperparameters as detailed in \citet{pinneri2021:strong-policies}. 

APEX finds various, surprisingly creative ways to move the object to the goal position, including pushing, sliding or flicking the object.
For the experiments on policy generalization (\sec{sec:SuppResFPPGen}), we only considered sequences in which the object was grasped and lifted.
Thus, we excluded all sequences in which the object moved while not being inside the gripper. 
For training and testing we  considered 3.84k sequences wit a length of 25 time steps, in which the hand graps the object after at $t=5$. 
A grasp was only considered if the relative $x-$ and $y-$ distance to the gripper was less than 0.0005 and the relative $z-$distance was below 0.15.
Additionally, the object must not have changed its position before $t=5$ to exclude sequence in which the object was pushed before.
We randomly split this dataset into a training (3.2k) and testing set (640).
For the generalization set we used 3.2k randomly selected sequences in which the grasp occurs at a later time $t$ with $t \in [6, 10]$.

In an additional experiment outlined in \supp{sec:resFPP} we train the networks on all kinds of sequences.
For that, we randomly split the collected dataset without further filtering into training (12.8k sequences), validation (6.4k sequences) and testing (6.4k sequences) sets.
Here we considered sequences that are 50 time steps long.

In both experiments we trained the networks using minibatches of size 128 for 5k epochs using scheduled sampling \cite{scheduledSampling}, where we exponentially decreased the sampling probability $p_i$ to a minimum value of $p_\mathrm{min} = 0.05$.
The latent state $\bm{h}_t$ of all RNNs was 16-dimensional. 
The first latent state $\bm{h}_0$  was initialized based on the initial input $(\bm{o}_1, \bm{a}_1)$ using a 2-layered MLP $f_\mathrm{init}$  (neurons per layer: $32 \rightarrow 16$).
All RNNs used a 3-layered preprocessing $f_\mathrm{pre}$ (neurons per layer: $64 \rightarrow 32 \rightarrow 16$) and a linear mapping two-layered MLP $f_\mathrm{post}$ to the network output.
In the experiment using simpler, filtered data (\supp{sec:SuppResFPPGen}) we used one-layered RNN cells.
For the diverse data set (\supp{sec:resFPP}) we use stacked RNN cells (3 layers) and \method with 3-layered $g-$ and $r-$functions.

\subsection{Reinforcement Learning: General Training Principles and Hyperparameter Search}

\label{sec:SupplRL}

\begin{wrapfigure}{r}{.25\textwidth}
    \vspace{-0.5\baselineskip}
    \centerline{%
      \setlength{\fboxrule}{0pt}
      \setlength{\fboxsep}{0pt}%
      \fbox{\includegraphics[width=0.9\linewidth]{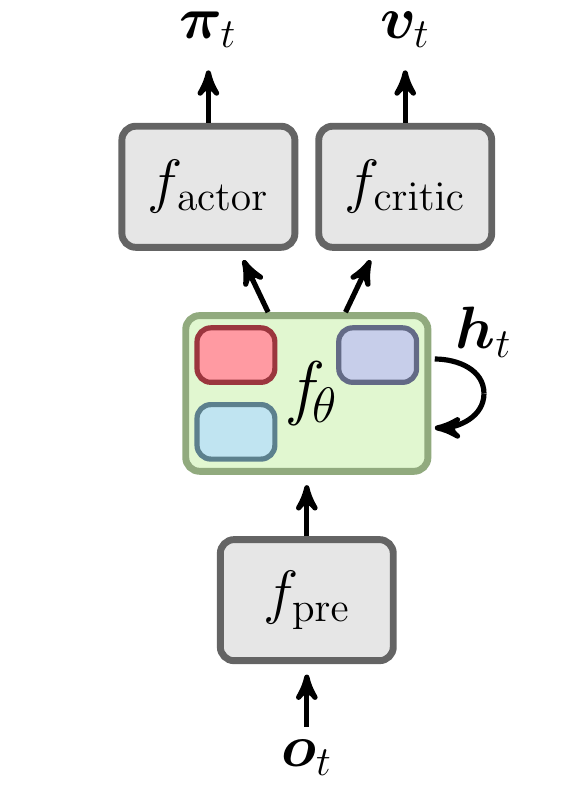}}
    }%
   \caption{Reinforcement learning architecture \label{fig:rlArchitecture}}
\end{wrapfigure}

For our reinforcement learning experiments in the Mini-Gridworld \cite{MiniGrid}, we used an actor-critic architecture as previously done by Chevalier-Boisvert et al. \cite{BabyAI}.\footnote{We used an implementation by one of the authors available at \url{https://github.com/lcswillems/rl-starter-files}. The code is licensed under MIT license.} 
The architecture is a modified version of our general architecture (\fig{fig:overall}),  shown in \fig{fig:rlArchitecture}.
The image-like input is preprocessed by a three-layered convolutional neural network $f_\text{pre}$ with $2 \times 2$ convolution kernels and with max-pooling after the first layer.
The 64-dimensional image embdeding is processed by an LSTM with 64-dimensional latent state.
The LSTM output is processed by two separate MLPs, akin to using two $f_\text{post}$ in \fig{fig:overall}, that take the role of the actor and the critic.
The actor MLP $f_\text{actor}$ outputs the policy $\bm{\pi}_t$, which determines the next action $\bm a_{t}$.
The critic MLP  $f_\text{critic}$ outputs a value estimate $\bm{v}_{t}$.
Both MLPs use two layers with 64 neurons on the intermediate layer.
In our experiments with \method, we only replace the LSTM cell and leave $f_{\text{pre}}$, $f_\text{actor}$, and  $f_\text{critic}$ unmodified.

As done by Chevalier-Boisvert et al. \cite{BabyAI}, we train the system using Proximal Policy Optimization (PPO) \cite{PPO} with parallel data processing.
We performed 4 epochs of PPO with a batch size of 256.
We took the PPO hyperparameters from \cite{BabyAI}, setting $\gamma=0.99$ and the generalized advantage estimation to $0.99$.

We train the system using Adam \cite{kingma2014adam}  with  $\beta_1 = 0.9$, $\beta_2 = 0.999$, and $\epsilon = 0.0001$.
To determine the learning rate $\alpha$ we ran a grid search on the vanilla system (LSTM) with $\alpha \in \{0.005, 0.001, 0.0005, 0.0001\}$ for two random seeds and compared the mean rewards after training.
In five of the six environments $\alpha = 0.001$ achieved the best results.
Thus, for consistency we ran the MiniGrid experiments with a learning rate of $\alpha=0.001$.
For the one environment (KeyCorridorS3R2) in which a smaller learning rate ($\alpha = 0.0005$) produced better results, we additionally evaluated the system with the optimized learning rate and report the results in \supp{sec:SuppMiniGridRewards}.
As before, we apply gradient norm clipping \cite{gradientNormClipping} with a clipping threshold of $0.1$.
The loss was backpropagated for 32 time steps.

When using \method, we simply replaced the LSTM cell and left all hyperparameters unmodified.
The PPO loss \cite{PPO} was used as $\mathcal{L}_\text{task}$ in \eqn{eq:loss} and we set $\lambda=0.01$ in all experiments.
All reinforcement learning experiments were run with 10 random seeds per configuration.

\subsection{MiniGrid}
\label{sec:SupplMiniGrid}

\begin{figure*}
\begin{subfigure}{0.24\linewidth}
\centering
\includegraphics[width=\linewidth]{figs/MiniGrid/DoorKey8x8.png}
\caption{DoorKey-8x8 \label{fig:DoorKey8x8}}
\end{subfigure}
\hfill
\begin{subfigure}{0.48\linewidth}
\centering
\includegraphics[width=\linewidth]{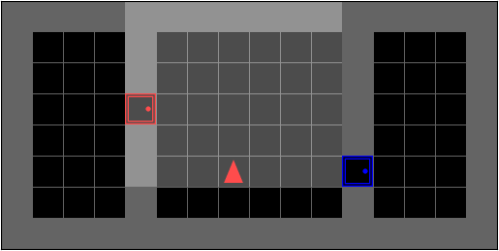}
\caption{RedBlueDoor-8x8 \label{fig:RedBlueDoor8x8}}
\end{subfigure}
\hfill
\begin{subfigure}{0.24\linewidth}
\centering
\includegraphics[width=\linewidth]{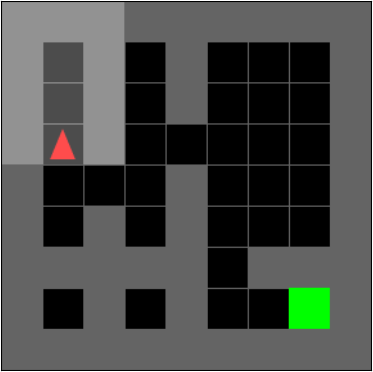}
\caption{SimpleCrossingS9N3\label{fig:SimpleCrossing}}
\end{subfigure}
\begin{subfigure}{0.24\linewidth}
\centering
\includegraphics[width=\linewidth]{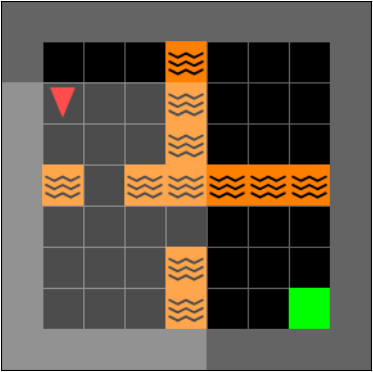}
\caption{LavaCrossingS9N2\label{fig:LavaCrossing}}
\end{subfigure}
\hfill
\begin{subfigure}{0.24\linewidth}
\centering
\includegraphics[ width=\linewidth]{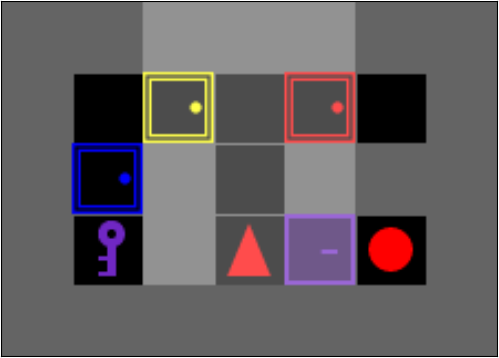}
\caption{KeyCorridorS3R2 \label{fig:KeyCorridor}}
\end{subfigure}
\hfill
\begin{subfigure}{0.24\linewidth}
\centering
\includegraphics[width=\linewidth]{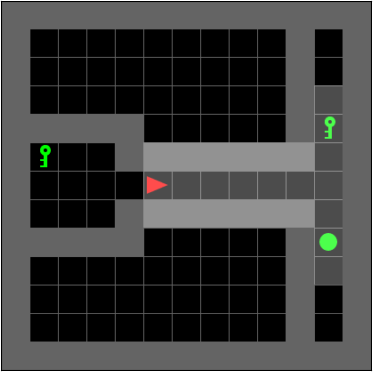}
\caption{MemoryS13-Random \label{fig:MemoryS13}}
\end{subfigure}
\begin{subfigure}{0.48\linewidth}
\centering
\includegraphics[width=\linewidth]{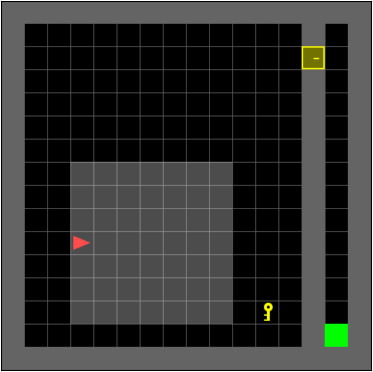}
\caption{DoorKey-16x16 \label{fig:DoorKey16x16}}
\end{subfigure}
\hfill
\begin{subfigure}{0.48\linewidth}
\centering
\includegraphics[width=\linewidth]{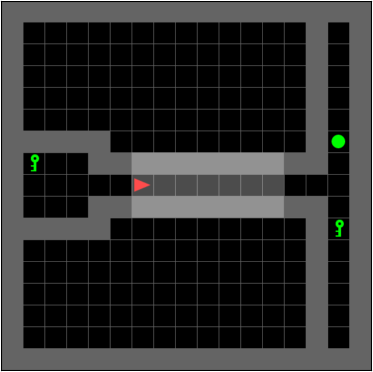}
\caption{MemoryS17-Random \label{fig:MemoryS17}}
\end{subfigure}
\vspace{-.5em}
    \caption{All MiniGrid envrionments used in this work. \label{fig:MiniGridEnvs}}
\end{figure*}

MiniGrid \cite{MiniGrid} is a library of partially-observable benchmark reinforcement learning problems.\footnote{MiniGrid is available at \url{https://github.com/maximecb/gym-minigrid}. MiniGrid is licensed under Apache License 2.0.}
All MiniGrid environments consist of a $N \times M$ tiles.
Each tile can be empty or contain one entity such as keys, doors, or walls.
The agent receives an image-like, egocentric view of the $7 \times 7$ tiles in front of the agent.
For each tile the agent receives a 3-dimensional signal, describing what type of object is in this tile, the color of the object, and its state (e.g. open, closed, or locked doors).
The agent can't see through walls or closed doors.
In every time step the agent can perform one of the following actions: move forward, turn left, turn right, pick-up an object, drop-off an object, or interact with an object (e.g. open doors).
In all environments a sparse reward of 1 is given once the task is fulfilled. In some environments the time to fulfill a task is used to discount the rewards.
\Fig{fig:MiniGridEnvs} shows all the problems we consider.

In \textbf{DoorKey-8x8} (\fig{fig:DoorKey8x8}) the agent needs to move to the green square behind a locked yellow door. The agent needs to learn to pick up a yellow key to open the door. The environment is $8 \times 8$ tiles big but the size of the two rooms varies per simulation. \textbf{DoorKey-16x16} (\fig{fig:DoorKey16x16}) is the same problem but in a larger $16 \times 16$ environment. We use the larger version to test zero-shot generalization, by training the system on the smaller environment and testing it on the larger one (see \sec{sec:SuppMiniGridZeroShot}).

In \textbf{RedBlueDoors-8x8} (\fig{fig:RedBlueDoor8x8}) the agent is randomly placed in a room ($8 \times 8$ tiles) with a red and a blue door. The agent has to first open the red door and afterwards open the blue door. Opening the blue door first result in ending the simulation  without any reward. 

In \textbf{SimpleCrossingS9N3} (\fig{fig:SimpleCrossing})  the agent needs to navigate through a maze to a green square in the bottom left corner. 
The maze is randomly constructed by three walls that run horizontally or vertically through the room. 
Each wall has a single gap. 
\textbf{LavaCrossingS9N2} (\fig{fig:LavaCrossing}) poses the same problem, however, the walls of the maze are replaced by two lava rivers.
Lava rivers do not occlude the view but entering lava terminates the episode without rewards.
Because of the early terminations and sparse rewards, this environment is much more challenging to learn than the maze with walls.

In \textbf{KeyCorridorS3R2} (\fig{fig:KeyCorridor}) the agent needs to pick up a ball. The ball is locked behind a door and the key is hidden in some other room. 
Thus, the agent needs to learn to explore the rooms, by opening differently colored doors, to find the key.
The agent can only pick up the ball if the agent is not holding the key, so after unlocking the door leading to the ball, the agent needs to drop the key.

\textbf{MemoryS13Random} (\fig{fig:MemoryS13}) is a memory task. Here the agent needs to memorize a green object (key or ball) in one room, move through a corridor, and then either go left or right to the matching object. The environment is $13 \times 13$ tiles big.
The length of the corridor is randomly generated per run. In \textbf{MemoryS17Random} (\fig{fig:MemoryS17}) the same problem needs to solved, but the environment is bigger ( $17 \times 17$ tiles). We use this version to test zero-shot generalization, by training the system on the smaller environment and testing it on the larger one (see \sec{sec:SuppMiniGridZeroShot}).

\subsection{Code and Computation}
\label{sec:ComputeDetails}
The code to run our experiments can be found at \url{https://github.com/martius-lab/GateL0RD}
All experiments were run on an internal CPU cluster. Robot Remote Control experiment using \method take between 3-4 hours run time.  Billiard Ball and Fetch Pick\&Place experiments, which use larger datasets, take around 6-9 hours run time for \method. Shepherd simulations, which we train for twice the number of epochs, take approximately 18-22 hours of run time. 
The MiniGrid experiments vary largely in their training time and took between 2 and 30 hours to train.
The baseline RNNs are roughly a factor of 0.8 faster than \method. 
This is mainly due to their optimized implementation in PyTorch.

\section{Ablation studies}
In this section, we investigate the importance of each of the components of our proposed architecture. 
\subsection{Ablation 1: Ablation of the type of gate function}
\label{sec:SupplGateFunction}

\begin{wrapfigure}{r}{.3\textwidth}
    \vspace{-0.5\baselineskip}
    \centerline{%
      \setlength{\fboxrule}{0pt}
      \setlength{\fboxsep}{0pt}%
      \fbox{\includegraphics[width=0.9\linewidth]{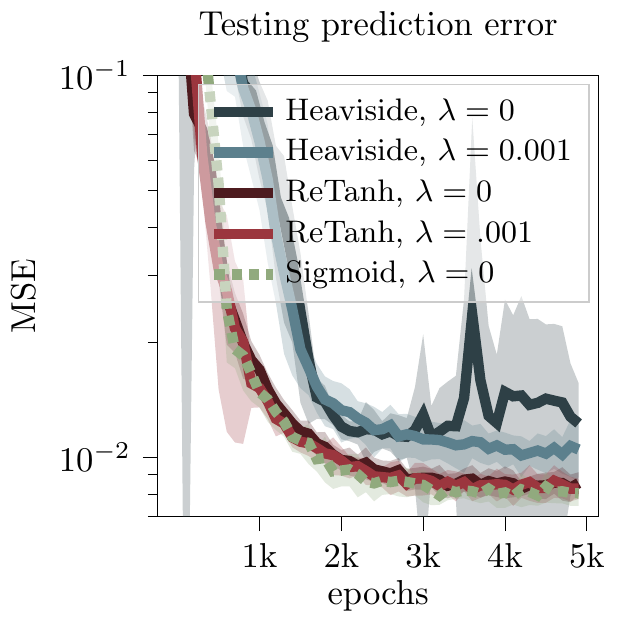}}
    }%
   \caption{Billiard Ball testing error for \method with different gate functions. \label{fig:ablation1}}
\end{wrapfigure}

We use the Billiard Ball scenario, trained using scheduled sampling \cite{scheduledSampling} as in \sec{sec:resBilliard}, to analyze the effect of different gate activation functions.
In one ablated setting, we replace our ReTanh activation $\Lambda$ in \eqn{eq:new_hidden} with a sigmoid activation function $\sigma$.
Additionally, we test using the Heaviside step function $\Theta$ as gate activation function in \eqn{eq:new_hidden}.
When using the Heaviside step function, we estimate the gradients using the straight-through estimator \cite{StraightThroughBengio}, which treats the step function as a linear function during the backward pass (illustrated in \fig{fig:Theta}). 
We test the Heaviside gates both with our $L_0$ loss ($\lambda = 0.001$) and without latent state regularization ($\lambda = 0$). 
Because a gate output of 0 is practically not achieved for the sigmoid function, we test the sigmoidal gates without latent state regularization ($\lambda=0$).

\Fig{fig:ablation1} shows the autoregressive prediction errors of the ablated versions of \method. 
The ablations with Heaviside gates perform worse than \method with the non-binary gates.
When using the Heaviside gate without any regularization, the mean prediction error even increases over training time.
\method with a sigmoid gate and our ReTanh gate reach the same level of prediction accuracy.

We believe that the worse performance of the Heaviside gate is due to the network profiting from multiplicative computations when computing the next latent state.
For the Heaviside gate, interpolations of old and new latent states are not possible.
Here, the latent state is either completely replaced or left unmodified.
We conclude that our novel ReTanh gate is as suitable for gating as the classically used sigmoid gate.
Additionally, it has the practical advantage of achieving an output of exactly 0, thus allowing the gate activation to be regularized as we do it with our $L_0$ loss.

\subsection{Ablation 2: Effect of gate stochasticity}
\label{sec:SupplGateStochasticity}

\begin{wrapfigure}{r}{.5\textwidth}
    \startsubfig{}
  \begin{minipage}[t]{.49\linewidth}\centering
    \subfig{fig:ablationGateNoiseErr} \\
    \includegraphics[width=\linewidth]{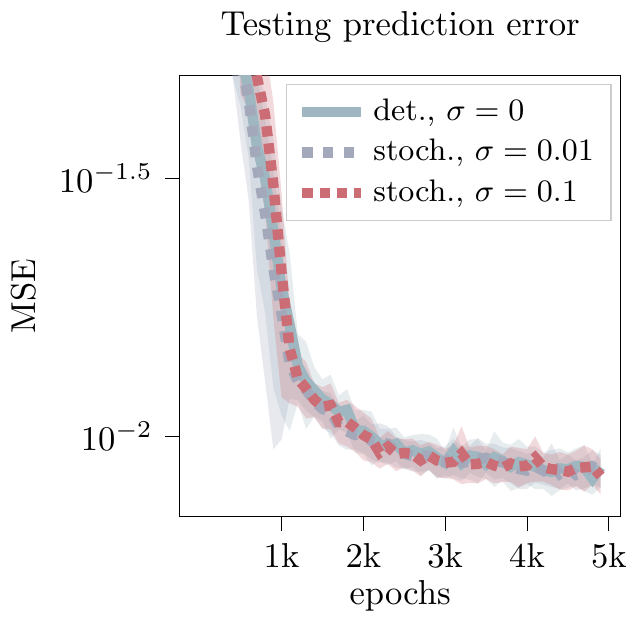}
  \end{minipage}\hfill
  \begin{minipage}[t]{.49\linewidth}\centering
    \subfig{fig:ablationGateNoiseGateRate} \\
    \includegraphics[width=\linewidth]{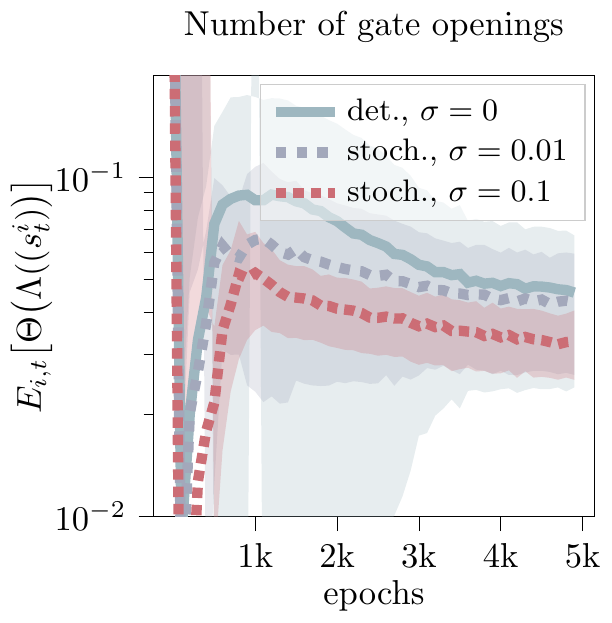}
  \end{minipage}\hfill
  \caption{Billiard Ball task comparing effect of gate noise on \method ($\lambda = 0.01$): prediction error (a) and mean number of ate openings(b). Shaded areas denote standard deviation.}
\end{wrapfigure}

To ablate the effect of the gate noise we compare \method with different strengths of the gate noise.
For deterministic gates we set $\epsilon = 0$ in \eqn{EqReparametrization}. 
Additionally we compare two values for the noise variance $\sigma$ of the diagonal covariance matrix $\bm{\Sigma}$ in \eqn{EqReparametrization}.
We test the effects of gate stochasticity for a fixed value of gate regularization $\lambda = 0.01$ in the Billiard Ball task.

\Fig{fig:ablationGateNoiseErr} shows the prediction errors comparing deterministic gates to stochastic gates with different gate noise.
There is no noticeable difference in prediction accuracy between the different settings.
Thus, reasonable values of noise on the gate input during training does not noticeably affect the prediction error during testing. \Fig{fig:ablationGateNoiseGateRate} shows the average latent state changes per sequence, computed as $\mathbb{E}_{i, t}\big[\Theta\big(\Lambda(s^i _t)\big)\big]$, for all settings.
Here, a larger value of gate noise results in fewer gate openings and thus, in fewer changes in the latent state.

We conclude that using stochastic gates together with our $L_0$ loss has a regularizing effect: \method trained with stochastic gates seems to achieve the same level of prediction accuracy as when trained with deterministic gates but changes its latent states more sparsely.

\newpage

\subsection{Ablation 3: Ablation of the latent state initialization network $f_\mathrm{init}$}
\label{sec:SupplFInit}

\begin{wrapfigure}{r}{.57\textwidth}
    \vspace{-0.5\baselineskip}
    \centerline{%
      \setlength{\fboxrule}{0pt}
      \setlength{\fboxsep}{0pt}%
      \fbox{\includegraphics[width=0.75\linewidth]{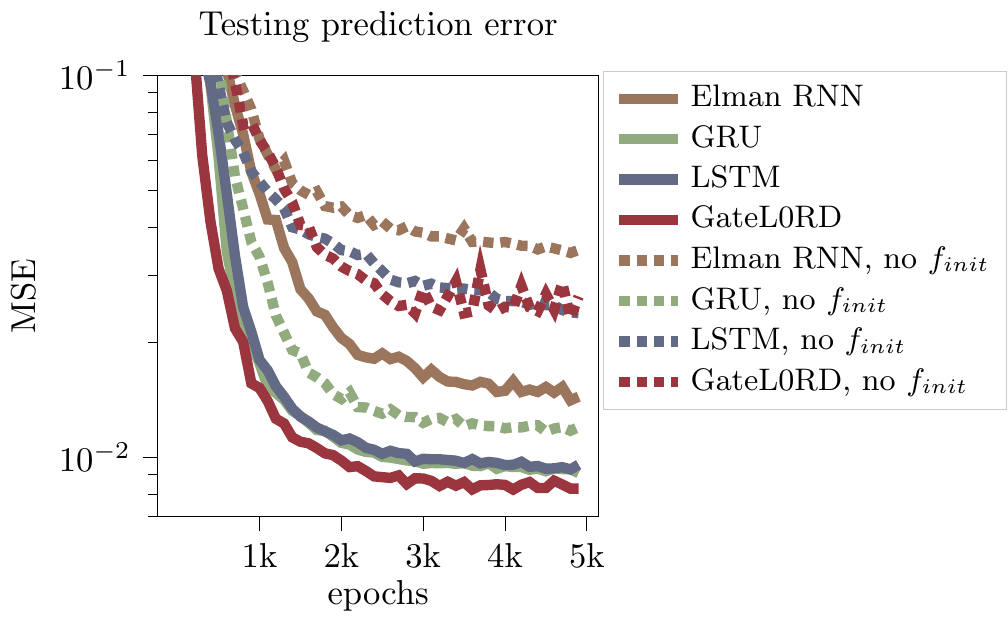}}
    }%
   \caption{Billiard Ball: effect of latent state initialization with or without $f_\text{init}$ on prediction errors.}
   \label{fig:ablation3}
\end{wrapfigure}

Next we ablate the effect of the context network $f_\text{init}$, which sets the latent state based on a few initial inputs (see \fig{fig:overall}). 
We compare all RNNs against variants without $f_\text{init}$ in the Billiard Ball scenario.
When omitting $f_\text{init}$, we initialize the latent state with $\bm{h}_0 = \bm{0}$.

\Fig{fig:ablation3} shows the prediction errors for all RNNs when using the context network $f_\text{init}$ (solid lines) and when initializing the latent state with zeros (dotted lines). 
The prediction accuracy decreases for all network types when trained without the context network.
However, how much their performance drops varies across the different RNN types.
GRUs seem to be much less affected by using them without $f_\text{init}$ than LSTMs and \method ($\lambda = 0.001$).

\subsection{Ablation 4: Ablation of the output function}
\label{sec:SupplOutputGate}
\begin{wrapfigure}{r}{.3\textwidth}
    \vspace{-0.5\baselineskip}
    \centerline{%
      \setlength{\fboxrule}{0pt}
      \setlength{\fboxsep}{0pt}%
      \fbox{\includegraphics[width=0.9\linewidth]{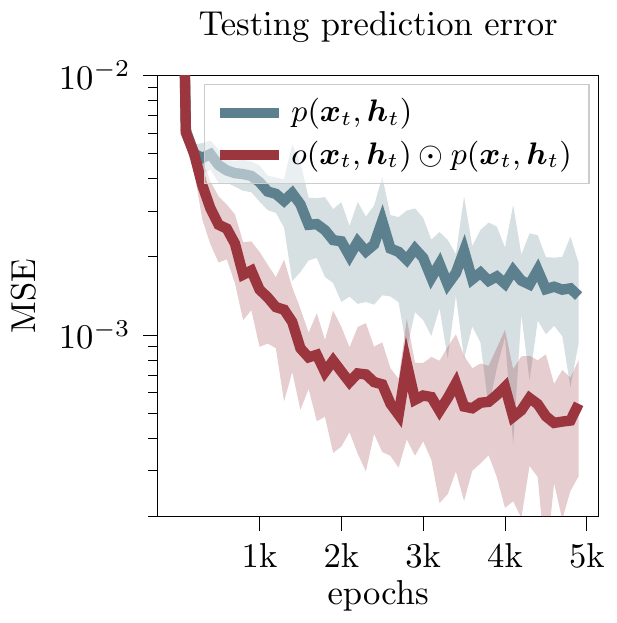}}
    }%
   \caption{Robot Remote Control: \method with an output function containing a multiplication ($p \odot o$) or not ($p$)}
   \label{fig:ablation4}
\end{wrapfigure}

After updating its latent state $\bm h_t$, \method uses two one-layered MLPs $p$ and $o$ to compute the network output as $p(\bm x_t , \bm h_t) \odot o(\bm x_t, \bm h_t)$ (see \eqn{eq:output}).
With this output function we want to enable both additive as well as multiplicative effects of the latent state $\bm h_t$ and input $\bm x_t$ on the network output. Is this justified or would a simple MLP as output function suffice? 

We analyze the effect of our output function in the Robot Remote Control Scenario ($\lambda=0.001$, trained on random action rollouts $\mathcal{D}_\mathrm{rand}$).
Here we compare \method using our standard output function ($p(\bm x_t , \bm h_t) \odot o(\bm x_t, \bm h_t)$) to an ablated version using just a one-layered MLP with $\tanh$ activation ($p(\bm x_t, \bm h_t)$).

\Fig{fig:ablation4} shows the resulting prediction errors of \method using its normal output function compared to the case without a multiplicative gate ($p(\bm x_t, \bm h_t)$). 
Clearly \method achieves a much better prediction when using a multiplicative output gate instead of a simple MLP.
Thus, a multiplicative branch for computing the network output seems to improve the prediction accuracy.
This may also explain the worse prediction accuracy of Elman RNNs in most tasks since they lack the multiplicative gates that can be found in all other investigated RNNs.

\subsection{Ablation 5: Comparison against $L_1$/$L_2$-versions}
\label{sec:SuppL1L2}
\begin{wrapfigure}{r}{.5\textwidth}
    \startsubfig{}
  \begin{minipage}[t]{.49\linewidth}\centering
    \subfig{fig:ablationL1L2TestErr} \\
    \includegraphics[width=\linewidth]{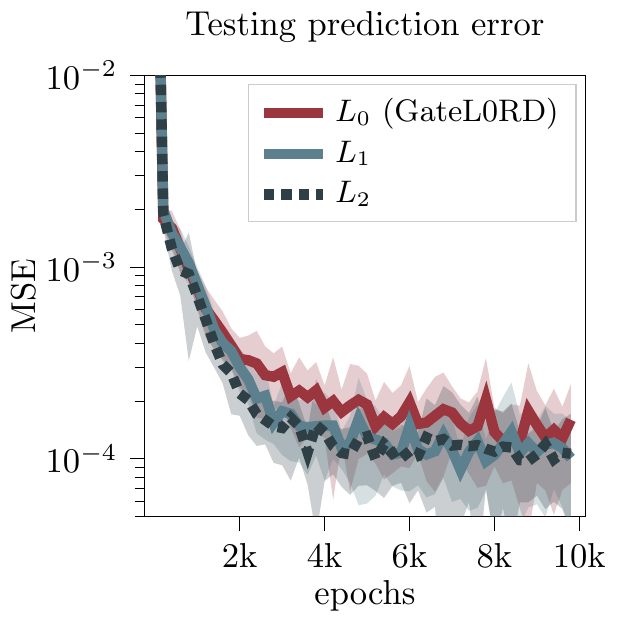}
  \end{minipage}\hfill
  \begin{minipage}[t]{.49\linewidth}\centering
    \subfig{fig:ablationL1L2GenErr} \\
    \includegraphics[width=\linewidth]{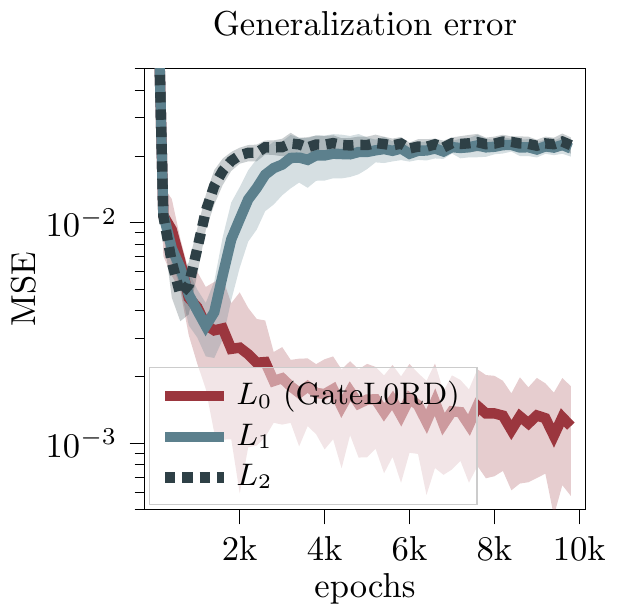}
  \end{minipage}\hfill
  \caption{Robot Remote Control: prediction error on the test set (a) and on the generalization set (b) for \method, $L_1$-, and $L_2$-variants.}
\end{wrapfigure}

Our hypothesis is that sparsely changing latent states allows better generalization across spurious temporal dependencies in the training data.
\method enforces such a sparsity of latent updates via an $L_0$-regularization of the changes in latent state. 
This is implemented using the novel ReTanh gate, instead of the commonly used sigmoid gates, and an auxiliary $L_0$ loss term that is made differentiable using the straight through estimator.
Is this necessary or would a simple sigmoid gate in conjuction with an $L_1$ or $L_2$ loss also improve generalization?

To analyze this, we compare \method against ablated versions that use a sigmoid gate and penalize the $L_1$ or $L_2$ norm of the gate activations.
We compare the version in the Robot Remote Control setting as in \sec{sec:resRRC}.
Thus, we train the networks on random action rollouts with linearly increasing action magnitude and test it either on data generated by the same process (testing) or on uniformly sampled random actions (generalization).
We chose a suitable regularization hyperparameter $\lambda = 0.001$ for all variants.

\Fig{fig:ablationL1L2TestErr} shows the prediction errors during testing for all variants. 
The $L_1$- and $L_2$-ablations achieve a very low prediction error on the test set, even exceeding \method's prediction in terms of accuracy.
However, when tested on the generalization set, shown in \fig{fig:resRRCGeneralization}, their prediction error increases drastically.

We conclude that the $L_1$/$L_2$-variants behave similar to GRUs and LSTMs (compare \fig{fig:resRRCGeneralization} and \fig{fig:ablationL1L2GenErr}).
They achieve a low testing error but fail to generalize to data generated by a different policy. 
This suggests that they also strongly overfit to spurious temporal dependencies, unlike our $L_0$-version.

However, it is noteworthy that on the test set the $L_2$-variant manages to achieve the lowest mean prediction error of all investigated RNNs. 
Krueger and Memisevic \cite{RegHiddenStates} previously suggested to penalize the $L_2$ norm of latent state changes in RNNs to prevent exploding or vanishing activations.
Our results suggest that applying $L_2$-regularization on the latent state changes seems to be a promising approach to increase the in-distribution performance of RNNs.

\section{Additional experiments and analysis}

\subsection{Billiard Ball: Analyzing the latent states and gate usage}
\label{sec:SuppBBLatent}

\begin{figure*}
  \startsubfig{}
  \begin{minipage}[t]{.43\linewidth}\centering
    \subfig{fig:BBtraj2} example trajectory\\
    \includegraphics[width=\linewidth]{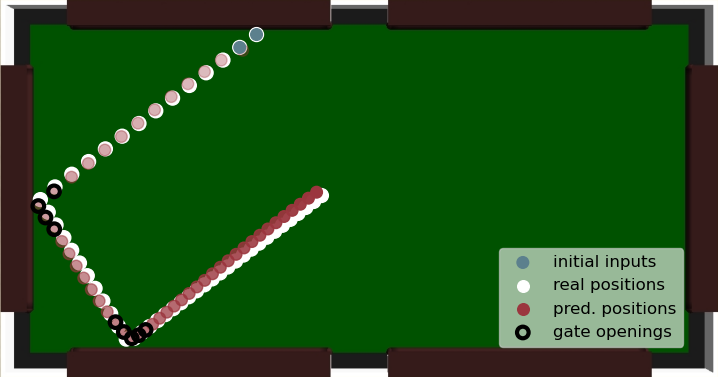}
  \end{minipage}\hfill
  \begin{minipage}[t]{.56\linewidth}\centering
    \subfig{fig:BBlatent2}: RNN latent states ($h_t^i - h_0^i$)\\
    \begin{tabular}{@{}p{.17\linewidth}@{\ \ }r@{}}
      \method & \includegraphics[width=.8\linewidth,align=c]{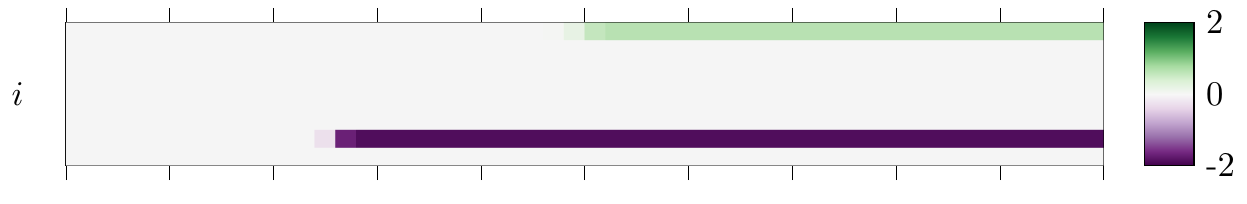}\\
      GRU & \includegraphics[width=.8\linewidth,align=c]{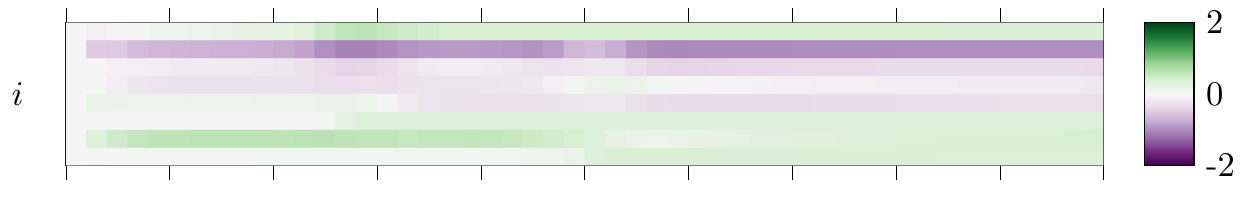}\\
      LSTM & \includegraphics[width=.8\linewidth,align=c]{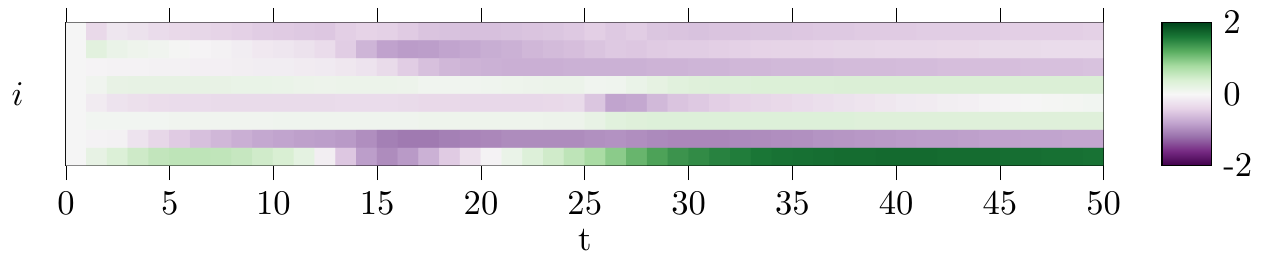}
    \end{tabular}
  \end{minipage}
  \vspace*{0.5cm}
  
  \begin{minipage}[t]{.43\linewidth}\centering
    \subfig{fig:BBtraj3} example trajectory\\
    \includegraphics[width=\linewidth]{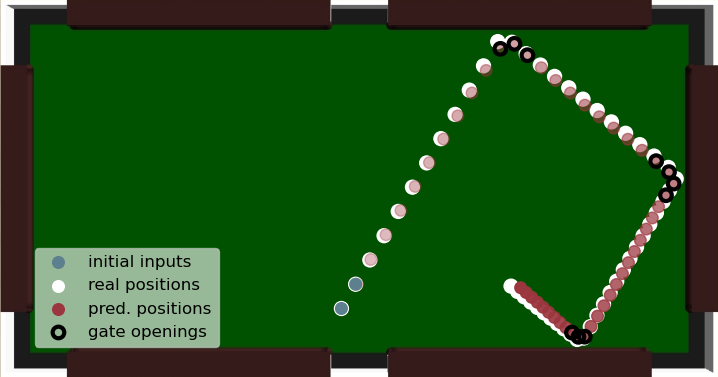}
  \end{minipage}\hfill
  \begin{minipage}[t]{.56\linewidth}\centering
    \subfig{fig:BBlatent3}: \method latent states ($h_t^i - h_0^i$)\\
    \begin{tabular}{@{}p{.17\linewidth}@{\ \ }r@{}}
      $\lambda = 0$ & \includegraphics[width=.8\linewidth,align=c]{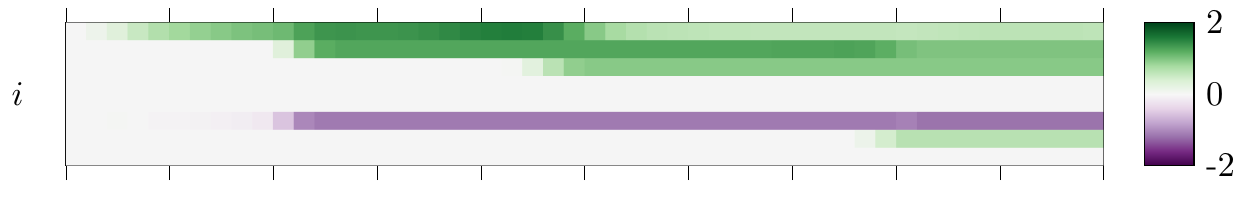}\\
      $\lambda = .001$ & \includegraphics[width=.8\linewidth,align=c]{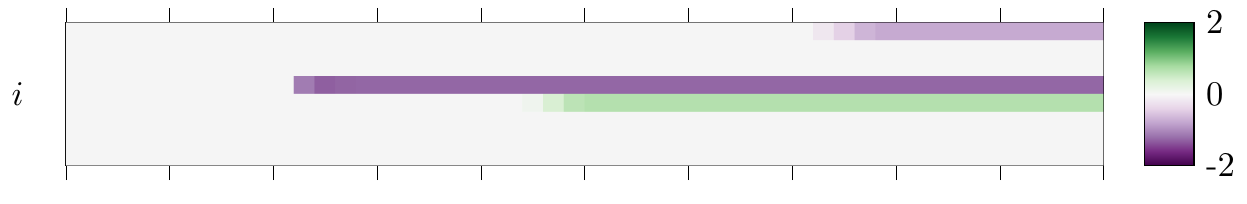}\\
      $\lambda = 0.01$ & \includegraphics[width=.8\linewidth,align=c]{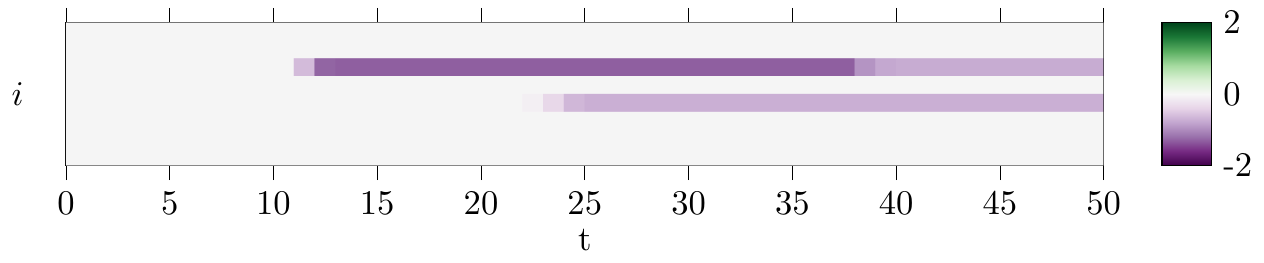}
    \end{tabular}\vspace{-.5em}
  \end{minipage}
  \caption{Billiard Ball example trajectory. (a) \& (c): Real positions are shown in white, the provided inputs in blue,  \method  ($\lambda=0.01$) position predictions in red (saturation increasing with time). The inputs for which at least one gate opened are outlined in black. (b): The latent states $\bm h_t$ for different RNNs for the sequence shown in (a). (d): The latent states $\bm h_t$ for \method with different values of $\lambda$ for the sequence shown in (c). Latent states are shown relative to the initial latent state $\bm h_0$. \label{fig:BBLatentExample}}
\end{figure*}

In this section, we provide further exemplary latent states for RNNs when applied to the Billiard Ball scenario.
\Fig{fig:BBLatentExample} shows two exemplary ball trajectory and the corresponding latent states. \method is able to make accurate autoregressive predictions (see red dots in \fig{fig:BBtraj2} and \fig{fig:BBtraj3}) and tends to open its gates around wall collisions (black circles). 
 \Fig{fig:BBlatent2} shows the latent states of \method ($\lambda = 0.01$) compared to the latent states of a GRU and a LSTM for the trajectory shown in \fig{fig:BBtraj2}.
\method's changes in latent states are easily interpretable: \method seems to encode $x-$ and $y-$ velocity in two dimensions of its latent state and changes the latent state at these particular dimensions when the ball velocity changes upon collision. The LSTM and GRU also tend to change their latent states more around points of collision but also change many latent state dimensions throughout the trajectory, making them much harder to interpret. 

\Fig{fig:BBlatent3} shows the latent states of \method for the same sequence, shown in \fig{fig:BBtraj3}, using different values of the sparsity regularization hyperparameter $\lambda$. 
As before, \method with $\lambda=0.01$ uses two dimensions of its latent state to encode the ball velocity and updates these two dimensions upon collisions.
In this example, \method with $\lambda = 0.001$ uses three dimensions to encode the ball's velocity. 
With every collision a different latent state dimension is updated, instead of using the same dimension for changes in $y-$velocity, as done by \method with $\lambda=0.01$.
In this example, \method with $\lambda = 0$ uses five dimensions to encode $x-$ and $y-$ velocities. At points of collision, multiple latent dimensions change.

\begin{wrapfigure}{r}{.3\textwidth}
    \vspace{-0.5\baselineskip}
    \centerline{%
      \setlength{\fboxrule}{0pt}
      \setlength{\fboxsep}{0pt}%
      \fbox{\includegraphics[width=0.9\linewidth]{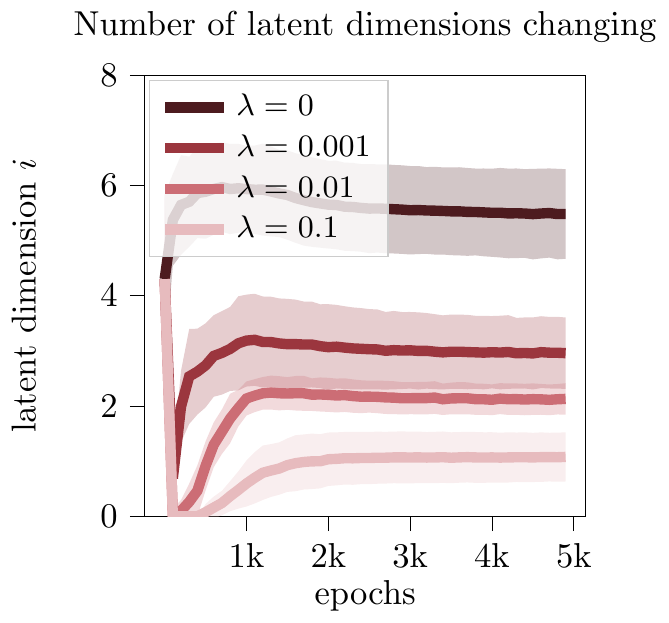}}
    }%
   \caption{Billiard Ball: latent state dimension usage for different values of $\lambda$.}
   \label{fig:dimsUsed}
\end{wrapfigure}

To further illustrate how the regularization hyperparameter $\lambda$ affects the latent state changes, we plot the number of latent state dimensions that change on average while predicting a Billiard Ball sequence in \fig{fig:dimsUsed}.
As expected, a stronger regularization through $\lambda$ results in fewer dimensions of the latent state changing.
Without regularization ($\lambda = 0$) \method changes on average less than 6 dimensions of the 8-dimensional latent state.
For $\lambda = 0.01$, \method quickly converges to on average using two latent states. 
For $\lambda = 0.1$, fewer latent state dimensions change on average.

As shown in \fig{fig:gateRateLambdas}, even without regularization ($\lambda=0$) \method continuously decreases the mean number of gate openings. After 5k epochs, GateL0RD on average opens a gate less than $50\%$ of the time. 
Similarly, it does not use all dimensions of its latent state, as shown in \fig{fig:dimsUsed}.
This effect emerges from the interplay of stochastic gradient descent and the ReTanh having gradients of 0 for inputs $s^i_t \leq 0$. 
Over training time, gates will randomly close and kept shut if they do not contribute to decreasing the loss. 
This effect is closely related to the "dying ReLU problem” when using ReLU activation functions \cite{DyingRelu}. 
While dying ReLUs are considered a problem, in our case this is advantageous whenever the gate regularization is beneficial. 
We believe that this results in \method, even without regularization, being more robust to out-of-distribution shifts than GRUs and LSTMs. 
For example, \method with $\lambda=0$ achieves a smaller mean autoregressive prediction error when trained using teacher forcing (\fig{fig:resTeacherForcing}), compared to the baseline RNNs.

\subsection{Robot Remote Control \& Shepherd: Loss and scheduled sampling}

\begin{figure*}[b]
  \startsubfig{}
  \begin{tabular}{@{}l@{\ \ }l@{\ \ }l@{}}
    \subfig{fig:RRC_SS_prob} & \subfig{fig:RRC_loss}&\subfig{fig:SMG_loss}\\[-1em]
    \includegraphics[width=0.32\linewidth]{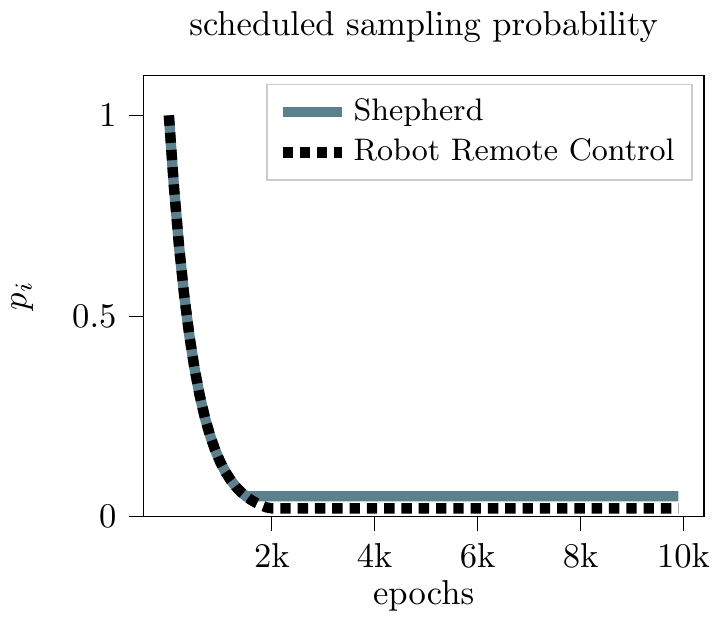}&
    \includegraphics[width=0.32\linewidth]{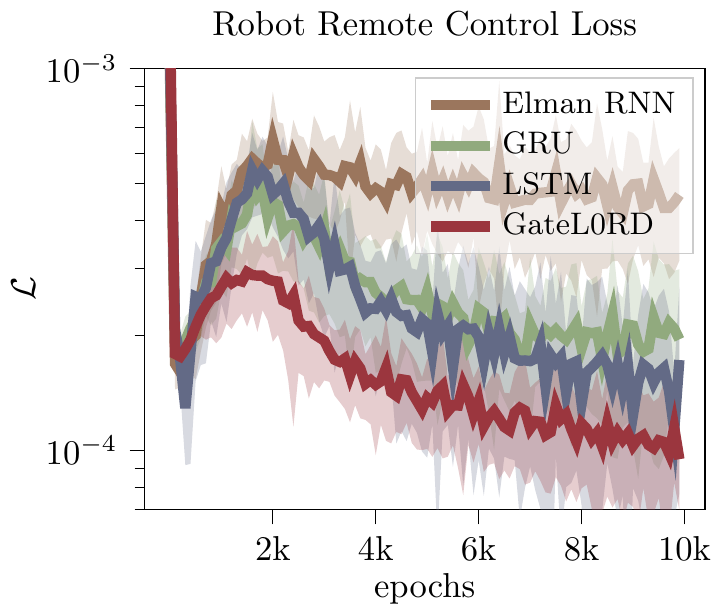}&
    \includegraphics[width=0.32\linewidth]{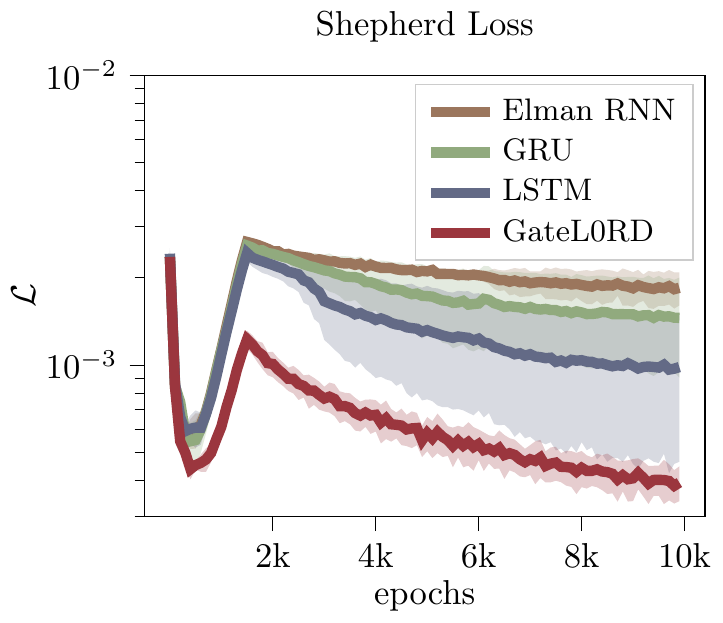}
  \end{tabular}\vspace{-1em}
  \caption{Scheduled sampling and loss curves: Probability $p_i$ of using the real input instead of the predicted input (a). Loss curves in  Robot Remote Control (b) and Shepherd (c). Shaded areas show standard deviation.}
\end{figure*}

In \fig{fig:RRC_loss} we provide the loss curves for the Robot Remote Control scenario and \fig{fig:SMG_loss} shows the loss curves for the Shepherd task.
For both tasks the loss decreases during the first couple of epochs, increases again until roughly 2k epochs, and continuously decreases afterwards.
This development is caused by using scheduled sampling \cite{scheduledSampling} as a training regime (detailed in \supp{sec:SuppTrainingDetails}).
The probability $p_i$ of applying teacher forcing exponentially decreases over the first 2k epochs, as shown in \fig{fig:RRC_SS_prob} for the Robot Remote Control task.
Thus, over the first 2k epochs the problems change from 1-step prediction problems to $N$-step prediction problems.
This drastically increases the difficulty within the first 2k epochs. 
However, this transition helps to learn autoregressive predictions \cite{scheduledSampling} as also demonstrated by our Billiard Ball experiments (\sec{sec:resBilliard}).

\subsection{Robot Remote Control: Improving RNN generalization}
\label{sec:RRCGenLRsExperiment}

\begin{wrapfigure}{r}{.3\textwidth}
    \vspace{-0.5\baselineskip}
    \centerline{%
      \setlength{\fboxrule}{0pt}
      \setlength{\fboxsep}{0pt}%
      \fbox{\includegraphics[width=0.9\linewidth]{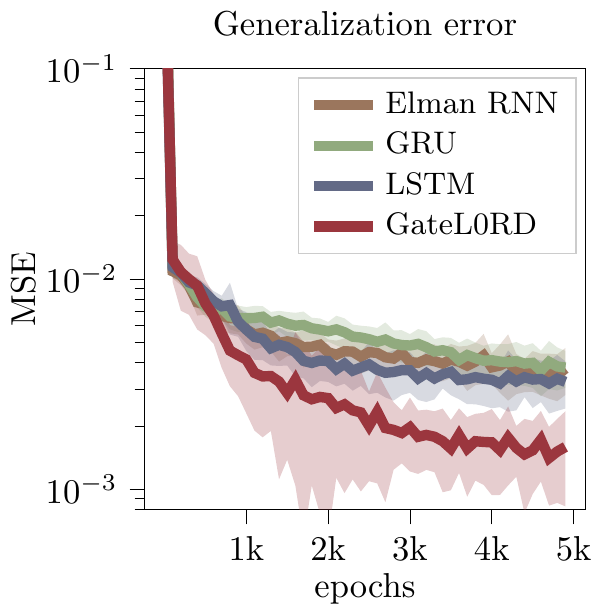}}
    }%
   \caption{Robot remote control generalization prediction error with optimized learning rates.}\label{fig:GenLRPlots}
\end{wrapfigure}

In \sec{sec:resRRC} we showed that LSTMs and GRUs trained for the Robot Remote Control environment using data in which action magnitude was positively correlated with time ($\mathcal{D}_\mathrm{time}$), failed to properly generalize to testing data without this correlation ($\mathcal{D}_\mathrm{rand}$).
\method showed less performance degeneration when tested on the generalization dataset.
We hypothesized, that \method's superior generalization performance was based on its tendency to only encode unobservable information within the latent states, making it less prone to overfit to observable spurious temporal dependencies within the training data.
However, an alternative explanation would be that the overfitting of LSTMs and GRUs was caused by their learning rate.
To investigate if the other RNNs' generalization abilities can be improved to the level of \method by choosing a different learning rate, we ran a grid search over three learning rate values ($\{0.005, 0.001, 0.0005\}$) for LSTMs, GRUs, Elman RNNs with two random initializations.
We selected the learning rate that lead to the lowest mean squared prediction error for the 50-timestep  predictions on the validation dataset of $\mathcal{D}_\mathrm{rand}$ after 5k epochs.
Seeing that a learning rate of $0.001$ yielded the best validation error for all RNNs, we reran the experiment with this learning rate (10 random seeds).

Figure \ref{fig:GenLRPlots} shows the resulting prediction error when testing the RNNs on the generalization test set of $\mathcal{D}_\mathrm{rand}$.
While the prediction error of GRUs and LSTMs on the generalization test set improved compared to our previous experiment, \method still achieved a lower prediction error on the generalization data than the other RNNs.
Note that \method was not further optimized in this experiment.
Thus, we conclude that \method's superior generalization performance in this setting is not caused by the learning rate.

\subsection{Robot Remote Control: Training on uniformly sampled random action rollouts}
\label{sec:RRCRandom}

\begin{wrapfigure}{r}{.3\textwidth}
    \vspace{-0.5\baselineskip}
    \centerline{%
      \setlength{\fboxrule}{0pt}
      \setlength{\fboxsep}{0pt}%
      \fbox{\includegraphics[width=0.9\linewidth]{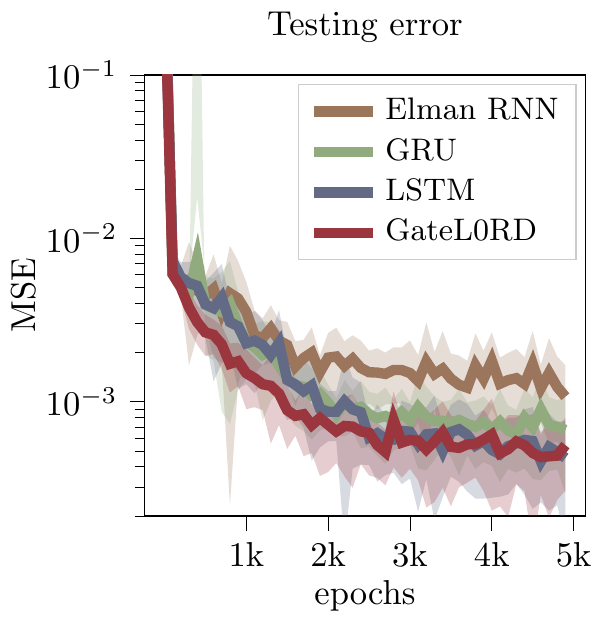}}
    }%
   \caption{Robot remote control prediction error when trained and tested on $\mathcal{D}_\mathrm{rand}$.}\label{fig:RandomActionsTestError}
\end{wrapfigure}

We previously showed for the Robot Remote Control environment that \method generalized better than the other RNNs to the data generated from random action rollouts ($\mathcal{D}_\mathrm{rand}$) when trained on a dataset that contained spurious temporal correlations ($\mathcal{D}_\mathrm{time}$) even for different learning rates.
Besides \method better capabilities in generalization, another explanation could be that \method is simply better at predicting sequences from the particular dataset $\mathcal{D}_\mathrm{rand}$.
To rule out this alternative explanation this, we trained the RNNs on a training set, generated from uniformly sampled random action rollouts ($\mathcal{D}_\mathrm{rand}$), and tested the network on data generated by the same process. 

\Fig{fig:RandomActionsTestError} shows the testing prediction error for predicting sequences based on the first observation and a sequence of actions.
After 5k epochs of training, LSTMs and GRUs achieve a similar prediction accuracy as \method ($\lambda=0.001$).
Thus, \method superior prediction accuracy on $\mathcal{D}_\mathrm{rand}$ in previous experiments can indeed be attributed to its better generalization capabilities.

\subsection{Robot Remote Control: Learned latent states}

\label{sec:SupplLatentStatesRRC}

\begin{figure*}
  \startsubfig{}
  \begin{minipage}[t]{.49\linewidth}\centering
    \subfig{fig:RRCLatentControl} example sequence with robot control\\
    \includegraphics[width=\linewidth]{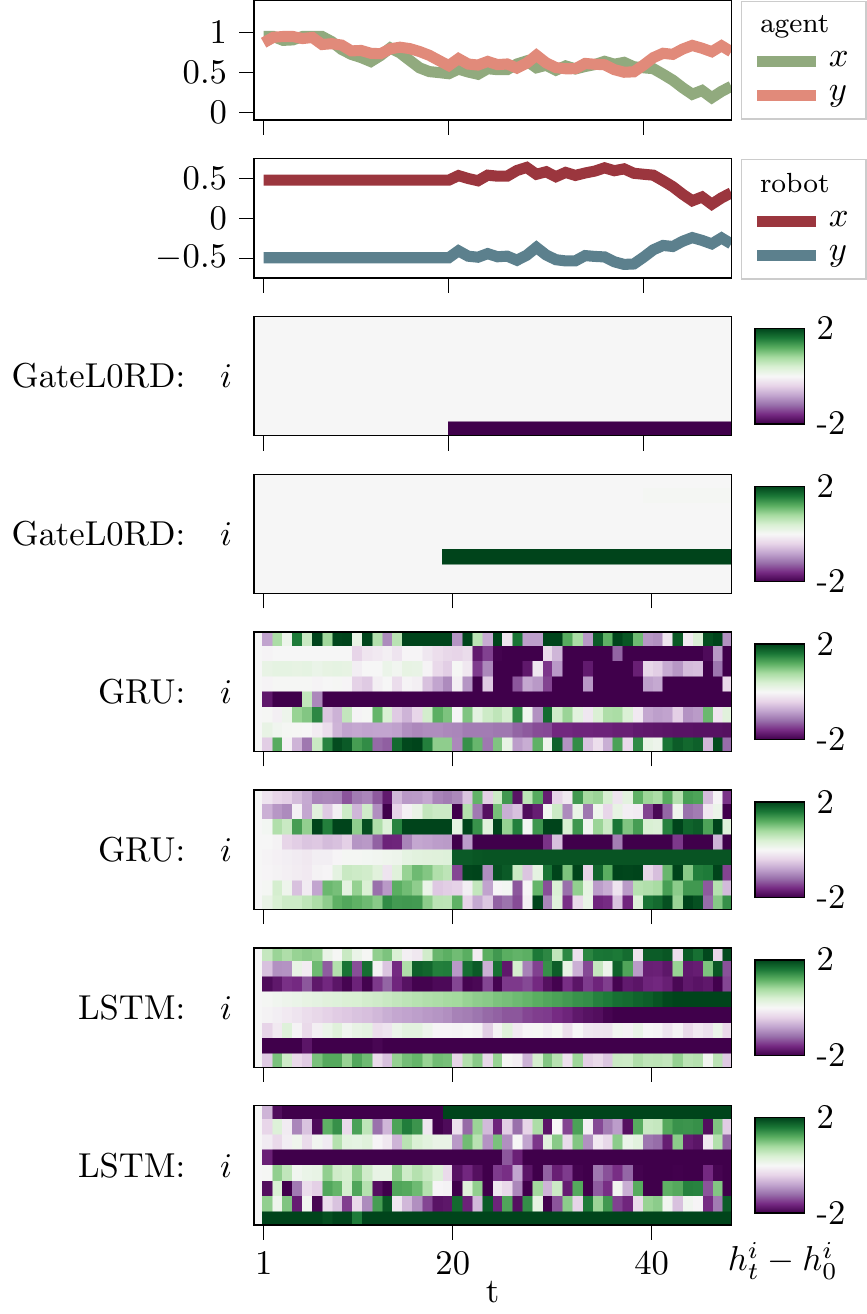}
  \end{minipage}\hfill
  \begin{minipage}[t]{.49\linewidth}\centering
    \subfig{fig:RRCLatentNoControl} example sequence without robot control\\
    \includegraphics[width=\linewidth]{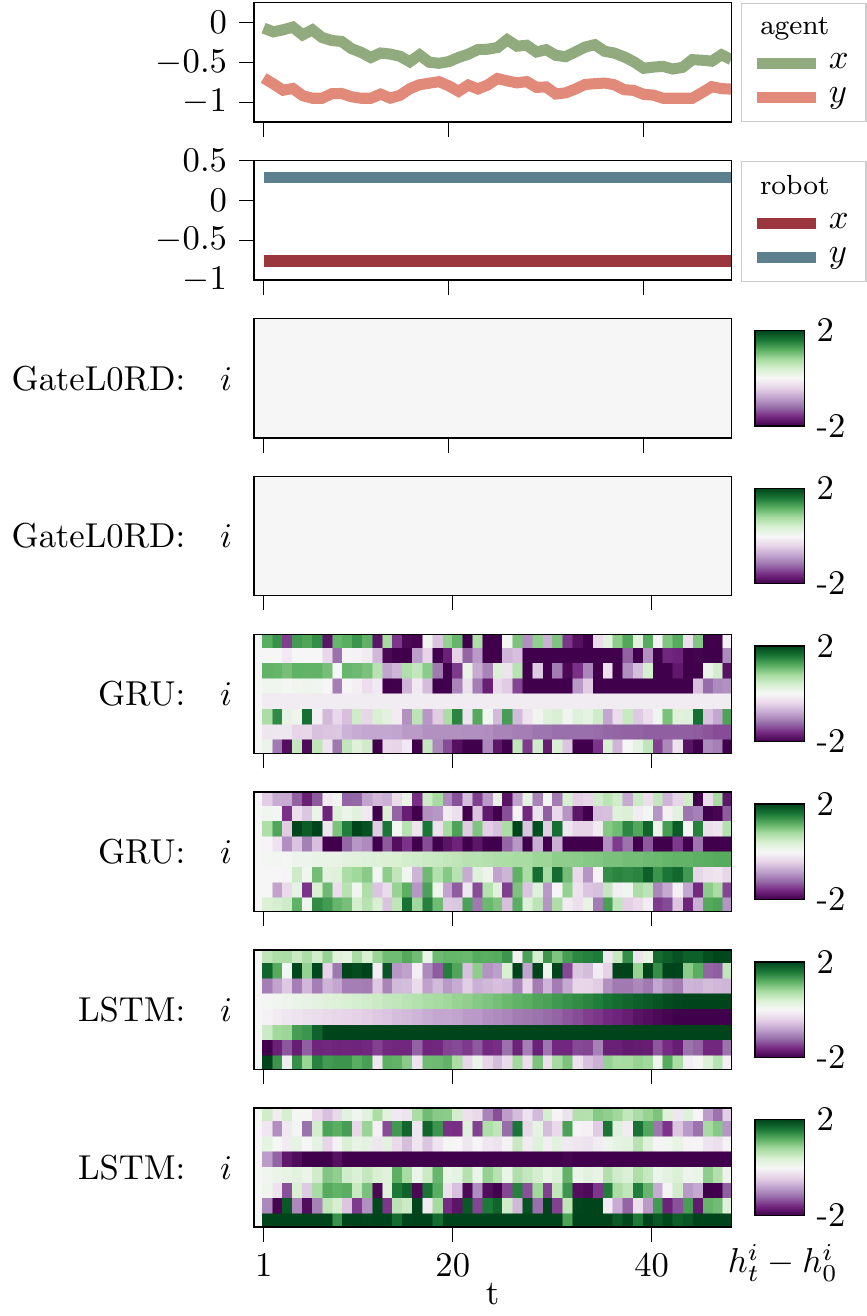}
  \end{minipage}\hfill
  \caption{Latent state for two exemplary Robot Remote Control sequences in which the robot was either controlled (a) or not (b). The latent states $\bm h_t$ are shown relative to their initialization $\bm h_0$. We provide the latent states for two GateL0RDs, GRUs, and LSTMs (different random seeds). One row shows the same random seed.}
\end{figure*}

In this section, we provide further exemplary latent states of the RNNs trained in the Robot Remote Control scenario as described in \sec{sec:resRRC}. 
\Fig{fig:RRCLatentControl} shows one exemplary sequence in which the robot was controlled by the agent and the corresponding latent states for two instantiations of \method, GRU, and LSTM with different random seeds.
\method seems to use one dimension of its latent state to encode when the agent controls the robot with its actions.
For GRUs and LSTMs the latent states also seem to strongly change around the point where the agent gains control over the robot, however, their latent states are not as clearly interpretable.
\Fig{fig:RRCLatentNoControl} shows one exemplary sequence, in which the robot was not controlled.
Here, \method does not modify its latent states, whereas LSTMs and GRUs continuously change their latent states over the course of the sequence.

Note that when the robot is not controlled, as in \fig{fig:RRCLatentNoControl}, Robot Remote Control is fully observable.
Thus, it seems that \method able to learn to distinguish observable from unobservable information and attempts to only update its latent state when the unobservable information changes.
To quantitavely evaluate this claim, we feed in all generalization sequences and classify the gate usage and unobservable events of task.
The inputs of the sequences were classified based on whether control of the robot was triggered at this time step (control) or not (no control). 
Additionally we analyzed for each input whether one of \method's gates opened (gate open) or not (gate closed).
The mean gate openings for the two events are shown in \tab{table:gating} with $\pm$ denoting standard deviation.
\method seems to mostly open its gates when the robot is controlled and tends to keeps its gate shut at other time steps.
Thus, \method indeed seems to mostly update its latent state when the unobservable state of the environment changes.

\begin{table*}[b]
    \centering
    \caption{Gating in Robot Remote Control}
    \label{table:gating}
    \begin{tabular}{lll}
        \toprule
          & \textbf{gate open} & \textbf{gate closed}\\
        \midrule
        \textbf{control}& $0.978 \pm 0.016$ (hits) & $0.022 \pm 0.016$ (misses)\\
        \textbf{no control} & $ 0.089 \pm 0.037$ (false alarms) & $0.911 \pm 0.037$ (correct rejections) \\
        \bottomrule
        \end{tabular}
\end{table*}

\subsection{Robot Remote Control: Clockwork RNNs}

In the tasks we considered, the latent states need to change at irregular times and are depending on the state of the environment. 
Thus, we hypothesize that RNNs operating on predefined time scales, such as Clockwork RNNs \cite{clockworkRNN}, are not well suited for these tasks.
We evaluate this in the Robot Remote Control task using Clockwork RNNs (CRNN, 3 clock modules, clock rates $T_1 = 1, T_2 = 4, T_3 = 8$). 
The learning rate ($\alpha = 0.001$) was determined via a grid search with $\alpha \in \{0.005,$ $0.001,$ $0.0005,$ $0.0001, 0.00005\}$. 
Unlike the other RNNs, the CRNN did not fully converge after 10k epochs, thus, instead, we trained it for 20k epochs.

In \tab{table:crnn} we list the mean prediction error after full training (20 random seeds, $\pm$ denotes standard deviation) on the test and generalization set, compared to the other RNNs.
CRNNs behave similarly to the other RNN baselines in that they achieve a reasonable test prediction error. 
However, they overfit even more drastically to the temporal correlations of the actions in the training set, resulting in a high prediction error on the generalization set.

\begin{table*}
    \centering
    \caption{Robot Remote Control: final prediction errors}
    \label{table:crnn}
    \begin{tabular}{lll}
        \toprule
          & \textbf{testing} & \textbf{generalization}\\
        \midrule
        \textbf{CRNN}& $0.00056  \pm 0.00015$ & $0.0331 \pm 0.0128$ \\
        \textbf{Elman RNN}& $0.00064 \pm 0.00036$ & $0.0062 \pm 0.0040$ \\
		\textbf{GRU}& $0.00030 \pm 0.00036$ & $0.0118 \pm 0.0058$ \\
		\textbf{LSTM}& $0.00018 \pm 0.00013$ & $0.0173 \pm 0.0058$ \\
		\textbf{\method}& $0.00015 \pm 0.00010$ & $0.0011 \pm 0.0007$ \\
        \bottomrule
        \end{tabular}
\end{table*}

\subsection{Fetch Pick\&Place: Generalization across grasp timings}
\label{sec:SuppResFPPGen}

In \sec{sec:resRRC} we showed using the Robot Remote Control scenario that \method is better at generalizing across spurious temporal dependencies in the training data than other RNNs.
In a follow-up experiment we want to investigate if similar effects can be found in a more complex environment, using more natural training data.
For that we use the \textbf{Fetch Pick\&Place} environment and train the networks to predict reach-grasp-and-lift sequences.
The training sequences were generated by a policy-guided model-predictive control method \cite{pinneri2021:strong-policies}.
Importantly, we train the network only on sequences where the gripper first touches the object exactly at time $t = 5$.
We test the networks on predicting sequences where gripper-object contact occurs as during training (testing set) or on sequences where the object is grasped later (generalization set).

\begin{wrapfigure}{r}{.5\textwidth}
    \startsubfig{}
  \begin{minipage}[t]{.49\linewidth}\centering
    \subfig{fig:FPPGraspTest} \\
    \includegraphics[width=\linewidth]{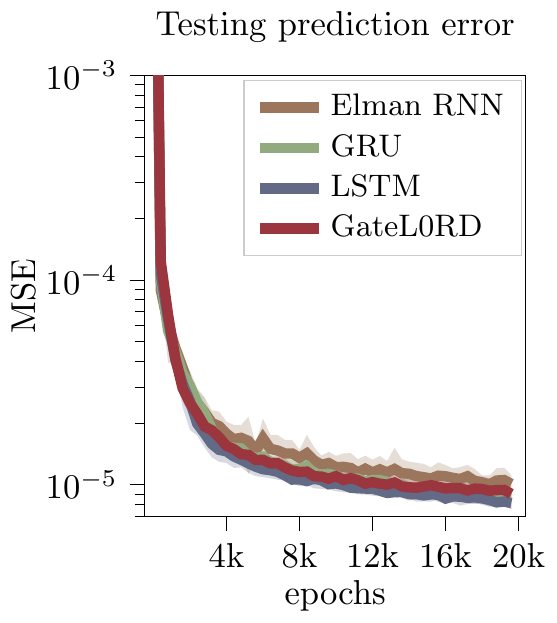}
  \end{minipage}\hfill
  \begin{minipage}[t]{.49\linewidth}\centering
    \subfig{fig:FPPGraspGen} \\
    \includegraphics[width=\linewidth]{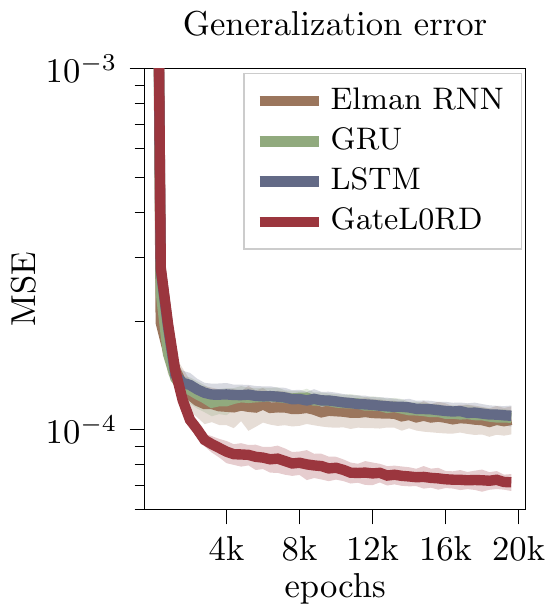}
  \end{minipage}\hfill
  \caption{Fetch Pick\&Place results: prediction error on test set (a) and generalization set (b). Shaded areas denote standard deviation.}
\end{wrapfigure}

\Fig{fig:FPPGraspTest} shows the mean prediction errors during testing. All networks achieve a very low prediction error.
The prediction accuracy is similar for all RNNs, but LSTMs achieve a slightly lower prediction error than \method ($\lambda=0.0001$).
When the networks were tested on sequences with different grasp timings, they produce much higher prediction errors as shown in \fig{fig:FPPGraspGen}.
\method prediction accuracy does not drop as strongly as the accuracy of the other networks.
Thus, as in the Robot Remote Control experiments, \method more robustly generalizes across spurious temporal correlations.

\begin{figure*}
  \startsubfig{}
  \begin{minipage}[t]{.49\linewidth}\centering
    \subfig{fig:FPPLatentLambdas1} Fetch Pick\&Place example sequence 1\\
    \includegraphics[width=\linewidth]{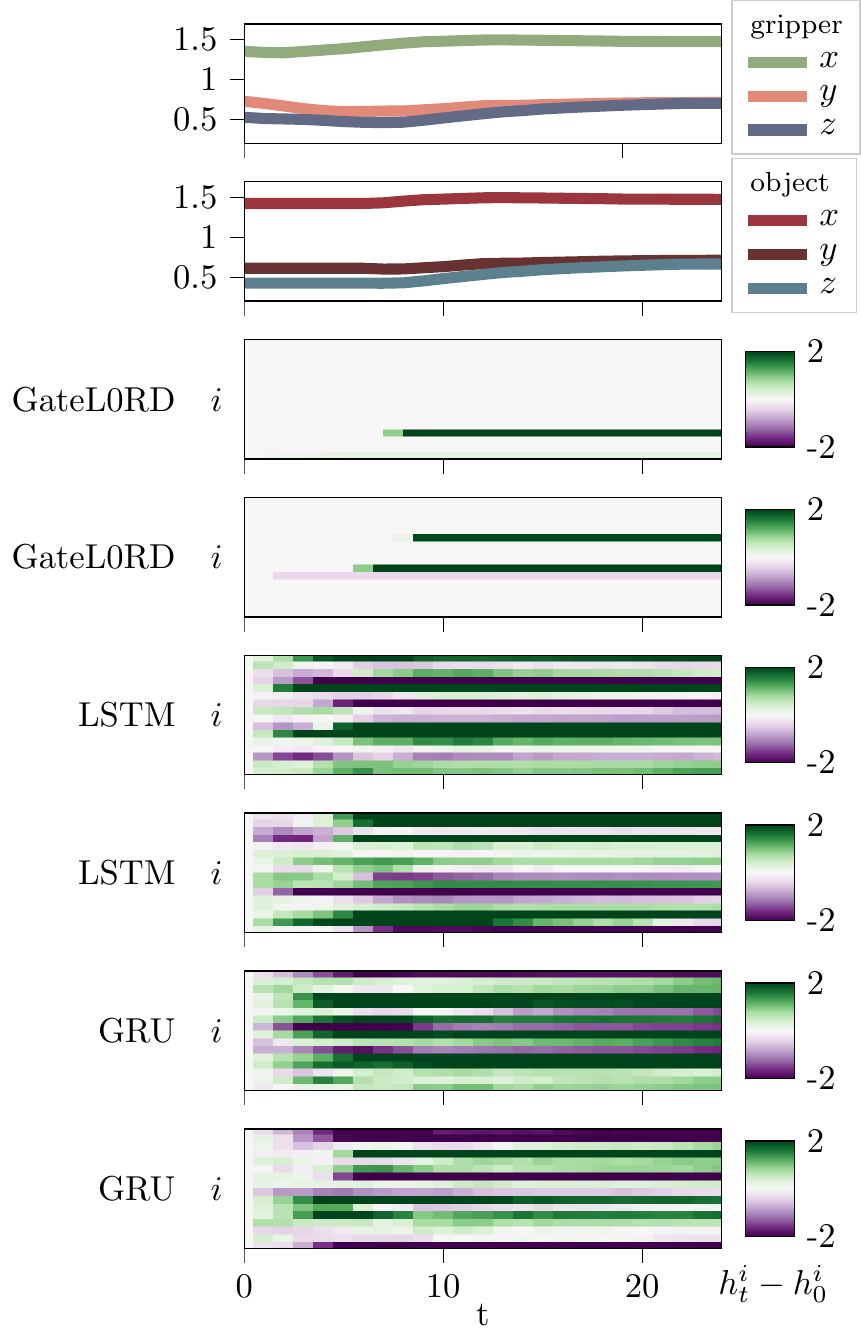}
  \end{minipage}\hfill
  \begin{minipage}[t]{.49\linewidth}\centering
    \subfig{fig:FPPLatentLambdas2} Fetch Pick\&Place example sequence 2\\
    \includegraphics[width=\linewidth]{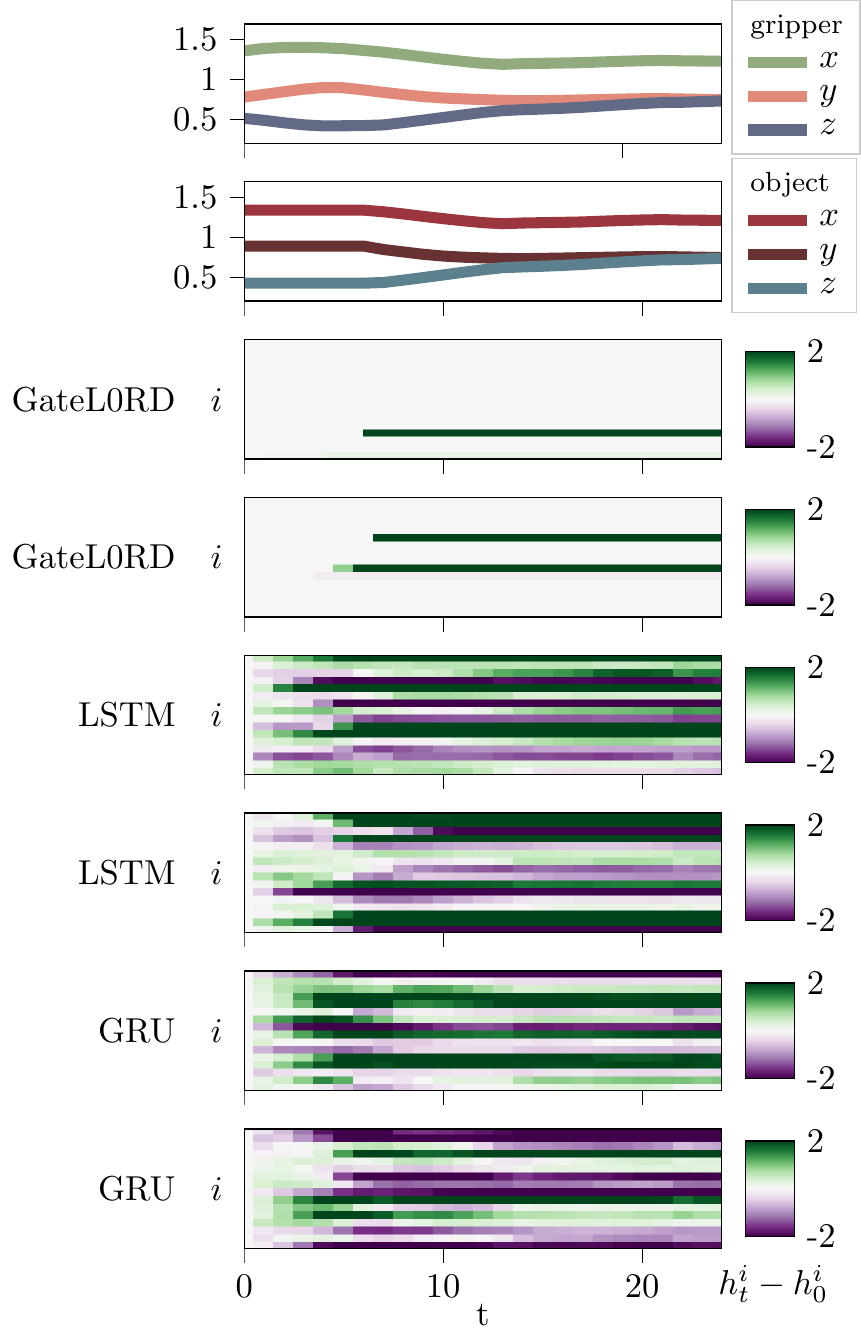}
  \end{minipage}\hfill
  \caption{Latent states of different RNNs for two exemplary sequences in the Fetch Pick\&Place environment. The latent states $\bm h_t$ are shown relative to their initialization $\bm h_0$. We compare the latent states for two random seeds each, where each row shows the same random seed. \label{fig:FPPLatentLambdas}}
\end{figure*}

\Fig{fig:FPPLatentLambdas} shows the latent states of the different RNNs when predicting two exemplary sequences. 
Here, \method ($\lambda=0.0001$) uses either one or three dimension of $\bm h_t$ that changes around the time when \method predicts that the gripper grasps the object.
During the predicted transportation of an object, the latent state does not change anymore. 
This hints at \method encoding the event of ``transporting an object'' in one dimension of its latent state.
For the other RNNs the latent state is not as easily interpretable.

\subsection{Fetch Pick\&Place: Training on diverse sequences}
\label{sec:resFPP}

Previously, we only considered reach-grasp-lift sequences in the Fetch Pick\&Place environment.
However, there are multiple other ways to move the object to a target position, such as pushing, sliding or even flicking.
Thus, in a next experiment we analyze the performance of the RNNs when trained as a model on a diverse set of sequences generated by the policy-guided model-based control method APEX \cite{pinneri2021:strong-policies}.

\begin{wrapfigure}{r}{.3\textwidth}
    \vspace{-0.5\baselineskip}
    \centerline{%
      \setlength{\fboxrule}{0pt}
      \setlength{\fboxsep}{0pt}%
      \fbox{\includegraphics[width=0.9\linewidth]{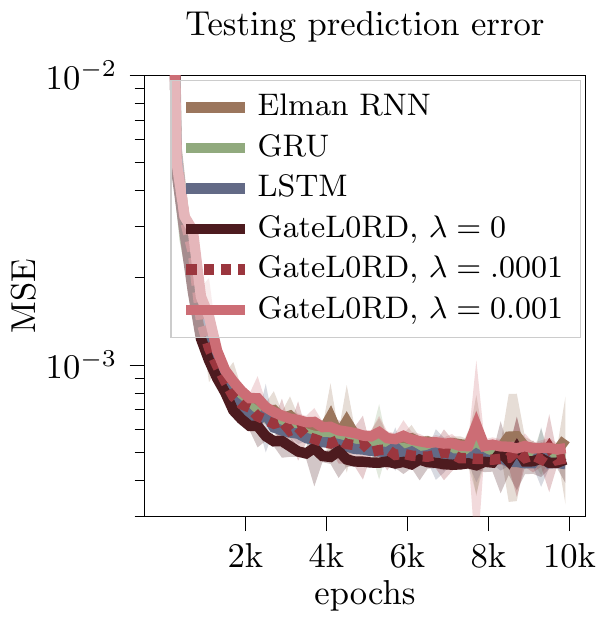}}
    }%
   \caption{Fetch Pick\&Place testing prediction errors.}\label{fig:FPPErrors}
\end{wrapfigure}

\Fig{fig:FPPErrors} shows the prediction errors of the RNNs when predicting testing sequences given the first observations and sequence of actions. 
In this scenario, all RNNs achieve a very similar prediction accuracy. 
\method with $\lambda = 0.001$ produces a slightly higher mean prediction error than the other RNNs, whereas \method with $\lambda = 0$ achieves a slightly lower error. 
We believe that in this scenario the small differences in prediction accuracy are a result of better approximations of the endeffector velocities.
In Fetch Pick\&Place the position control of the endeffector is realized by a PID-controller running at a higher frequency, thus, in this scenario continuous latent state updates are advantageous for predicting the endeffector velocity. 
Hence, in this scenario $\lambda$ regulates the trade-off between prediction accuracy and latent state explainability and needs to be chosen depending on priorities of the application.

\begin{figure*}
    \startsubfig{}
  \begin{tabular}{@{}l@{\ \ }l@{\ \ }l@{\ \ }l@{\ \ }l@{\ \ }l@{}}
    \subfig{} & \subfig{fig:DoorKey8x8Rewards}&\subfig{ }& \subfig{ }&\subfig{ }& \subfig{ }\\[-1em]
    \includegraphics[width=.16\linewidth]{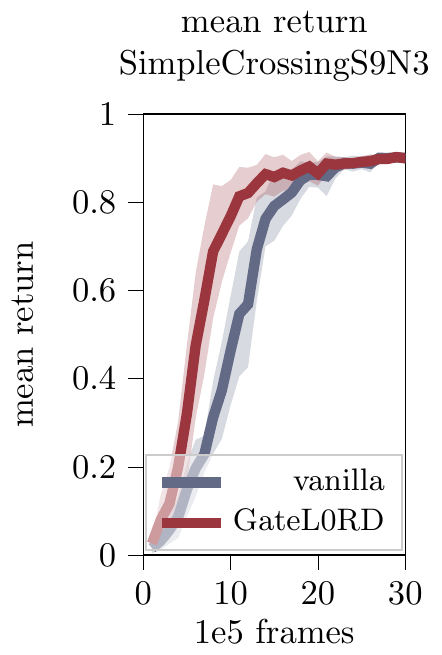}&
    \includegraphics[width=.16\linewidth]{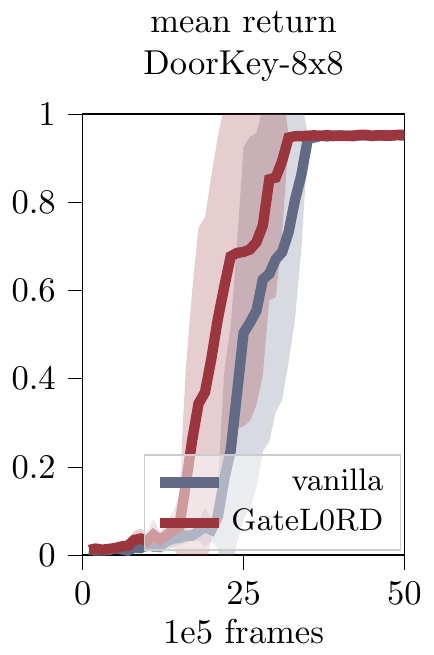}&
    \includegraphics[width=.16\linewidth]{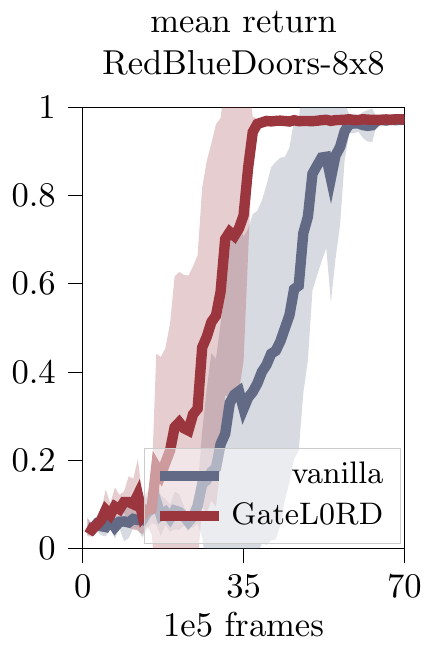}&
    \includegraphics[width=.16\linewidth]{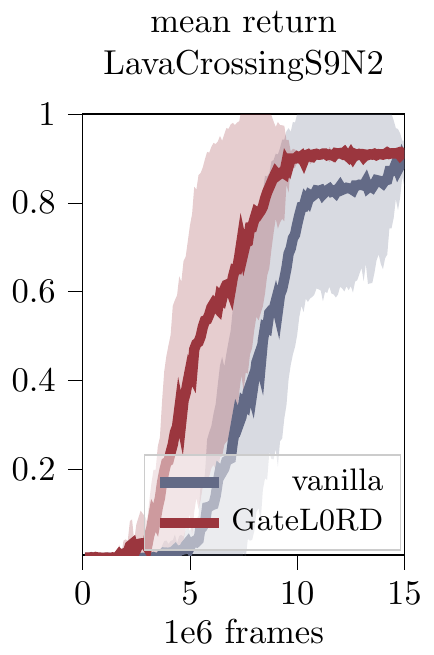}&
    \includegraphics[width=.16\linewidth]{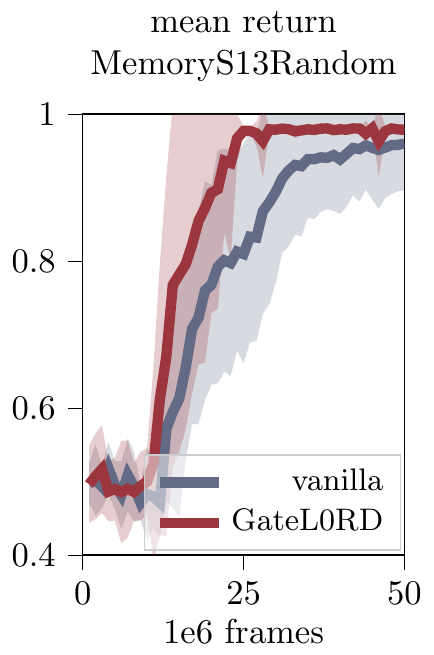}&
    \includegraphics[width=.16\linewidth]{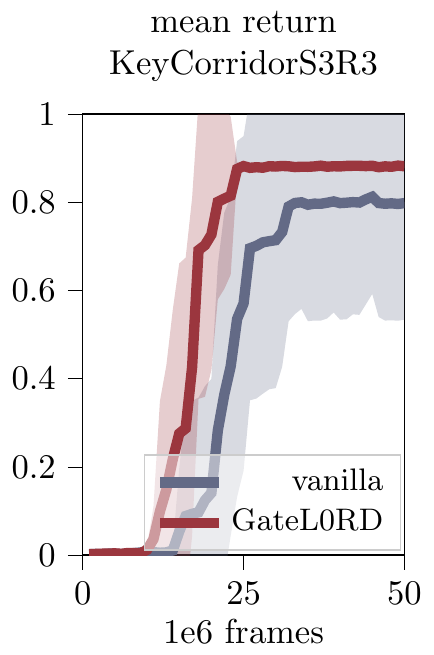}
  \end{tabular}\vspace{-1em}
  \caption{MiniGrid results: Mean rewards in solving various tasks when GateL0RD replaces an LSTM (vanilla) in a PPO architecture. Shaded areas show standard deviation. \label{fig:MiniGridRewards}}
\end{figure*}

\newpage 
\subsection{MiniGrid: Further analysis and experiments}
\label{sec:SuppMiniGridRewards}

\begin{wrapfigure}{r}{.4\textwidth}
    \startsubfig{}
  \begin{minipage}[t]{.49\linewidth}\centering
    \subfig{fig:exp4KeyImprovedSuccess} \\
    \includegraphics[width=\linewidth]{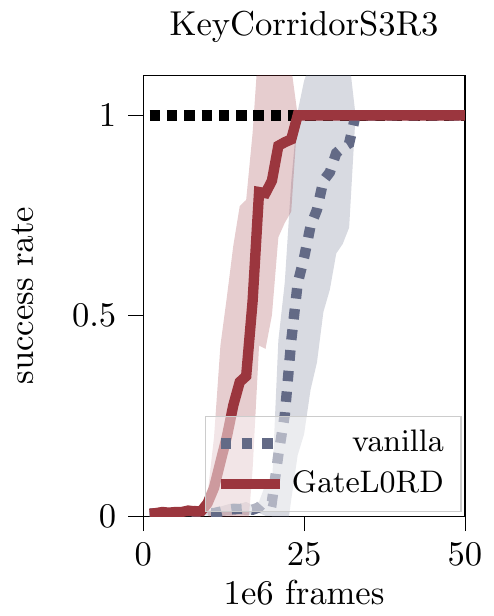}
  \end{minipage}\hfill
  \begin{minipage}[t]{.49\linewidth}\centering
    \subfig{fig:exp4KeyImprovedReward} \\
    \includegraphics[width=\linewidth]{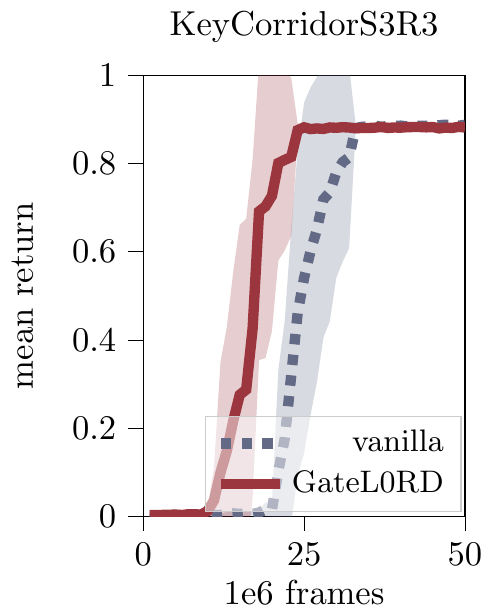}
  \end{minipage}\hfill
  \caption{MiniGrid results: mean success rate and reward for  KeyCorridorS3R3. The learning rate of the vanilla system was optimized for this problem.}
\end{wrapfigure}

In \sec{sec:resMiniGrid} we showed that \method is more sample efficient in achieving a high success rate in various MiniGrid tasks than when it replaces an LSTM in a PPO architecture.
Some problems of MiniGrid discount the overall reward based on the number of actions required to reach the goal.
Thus, another metric to judge success in MiniGrid is the mean reward collected by the systems.
\Fig{fig:MiniGridRewards} shows the mean rewards for the vanilla architecture and architecture containing \method over training experience.
For all problems the architecture containing \method is more sample efficient and achieves high levels of reward faster.

For consistency we used the same hyperparameters in all MiniGrid experiments and only swapped the LSTM cell for \method.
However, as described in \supp{sec:SupplRL} a grid search showed that for the KeyCorridorS3R3 problem a smaller learning rate ($\alpha = 0.0005$) resulted in higher mean rewards for the vanilla architecture.
Thus, to exclude the possibility that \method outperformed the LSTM in this problem based on the choice of learning rate, we ran an additional experiment in the KeyCorridorS3R3 problem with the vanilla architecture using the optimized learning rate.
The resulting mean success rate and mean rewards are shown in \fig{fig:exp4KeyImprovedSuccess} and \fig{fig:exp4KeyImprovedReward}, respectively.
While the vanilla architecture now manages to reach a success rate of 100\% and a mean reward larger than 0.8, \method is still faster in reaching the same level of performance.

\subsection{MiniGrid: Zero-shot policy transfer}

\label{sec:SuppMiniGridZeroShot}

\begin{figure*}[b]
  \startsubfig{}
  \begin{tabular}{@{}l@{\ \ }l@{\ \ }l@{\ \ }l@{}}
    \subfig{} & \subfig{} &\subfig{}& \subfig{}\\[-1em]
    \includegraphics[width=0.24\linewidth]{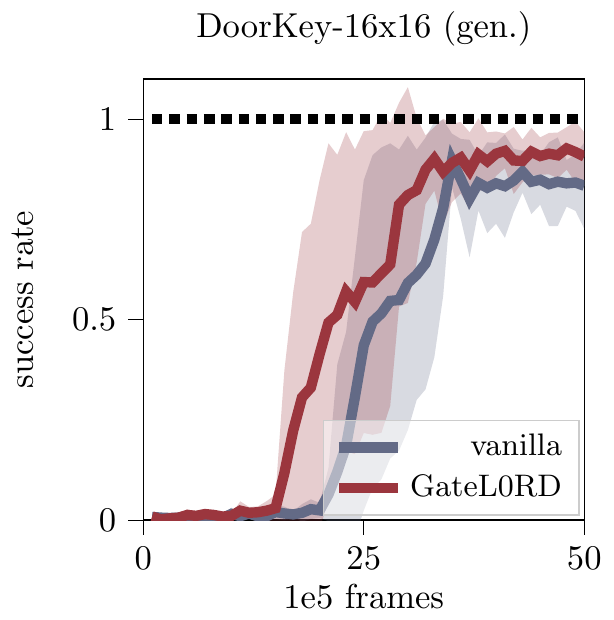} &
    \includegraphics[width=0.24\linewidth]{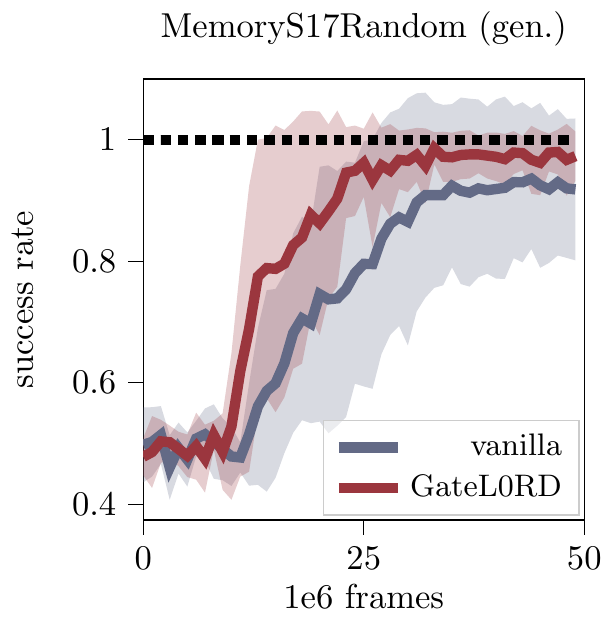} &
    \includegraphics[width=0.24\linewidth]{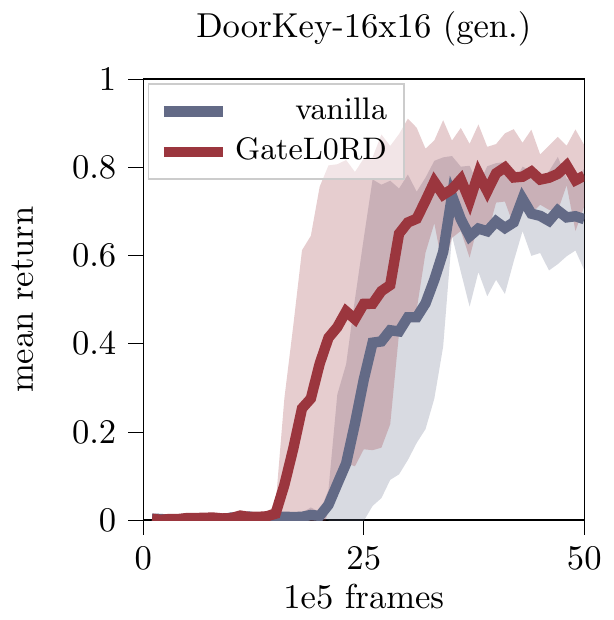} &
    \includegraphics[width=0.24\linewidth]{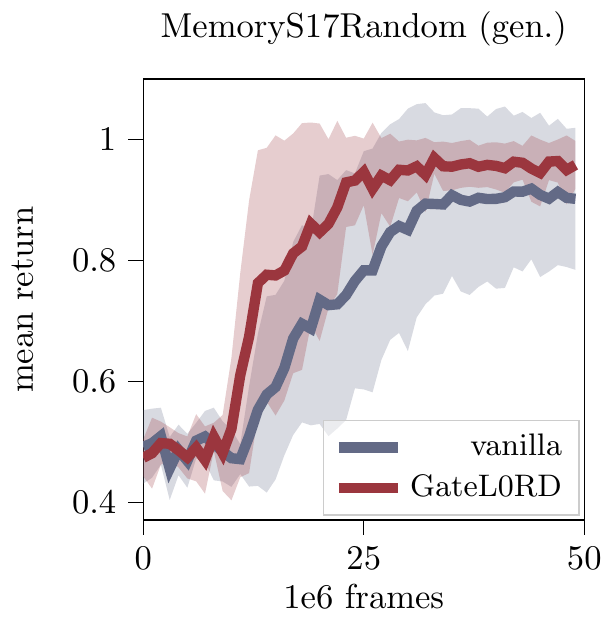}
  \end{tabular}\vspace{-1em}
  \caption{MiniGrid: zero-shot generalization for two environments when trained in a smaller versions of same problem. (a) \& (b) show the success rate and (c) \& (d) show the mean reward. Shaded areas show standard deviation. \label{fig:ZeroShot}}
\end{figure*}

We hypothesize that \method can  memorize information precisely without information loss over time.
Thus, it should be able to generalize well across different memory durations.
We investigate this aspect in in the MiniGrid domain by training a PPO architecture containing an LSTM (vanilla) and the same architecture containing \method on two problems that require memory, \ie DoorKey-8x8 (shown in \fig{fig:DoorKey8x8}) and MemoryS13Random (shown in \fig{fig:MemoryS13}). 
We evaluate the architectures on the same problems in larger environments, \ie DoorKey16x16 (shown in \fig{fig:DoorKey16x16}) and MemoryS17Random (shown in \fig{fig:MemoryS17}).
Thus, one of the main challenges is that during transfer information needs to be memorized for longer periods of time.

\Fig{fig:ZeroShot} shows the zero-shot generalization performance for solving the more complex problems after training only on the simpler variants.
For both problems \method achieves a higher mean success rate and mean reward than the vanilla baseline.
The better performance cannot simply be explained by \method being better at the considered task than the LSTM of the vanilla architecture. When tested in the simple problems both architectures achieve approximately the same performance (c.f., \fig{fig:DoorKey8x8Rewards}).
Instead the better performance is likely due to \method generalizing better from short-term to long-term memorization.

\end{document}